\documentclass[11pt]{article}

\usepackage[preprint]{acl}

\usepackage{times}
\usepackage{latexsym}

\usepackage[T1]{fontenc}

\usepackage[utf8]{inputenc}

\usepackage{microtype}

\usepackage{inconsolata}

\usepackage{graphicx}
\usepackage{amssymb}
\usepackage{pifont}

\newcommand{\cmark}{\ding{51}}
\newcommand{\xmark}{\ding{55}}

\usepackage{microtype}
\usepackage{hyperref}
\usepackage{url}
\usepackage{booktabs}
\definecolor{darkblue}{rgb}{0, 0, 0.5}
\hypersetup{colorlinks=true, citecolor=darkblue, linkcolor=darkblue, urlcolor=darkblue}

\usepackage{multirow}
\usepackage{bbding,twemojis}
\usepackage{graphicx}
\usepackage{adjustbox}
\usepackage{tcolorbox}
\usepackage{colortbl}
\tcbuselibrary{most}
\usepackage{cleveref}
\usepackage{subcaption}
\newtcolorbox{prompt}{
  enhanced,
  colback=gray!10,
  colframe=gray!50,
  arc=0mm,
  boxrule=0.5pt,
  left=5mm,
  right=5mm,
  top=2mm,
  bottom=2mm,
  fonttitle=\bfseries,
  coltitle=black,
  breakable,
  fontupper=\footnotesize\ttfamily,
      overlay={%
        \ifcase\tcbsegmentstate
        \or%
        \else%
        \fi%
    }
}

\newcommand{\verbcell}[1]{%
  \begin{Verbatim}[commandchars=\\\{\},breaklines=true,breakanywhere=true]
 #1
  \end{Verbatim}%
}

\title{\texttt{FLUKE}: A Linguistically-Driven and Task-Agnostic Framework for Robustness Evaluation}

\author{Yulia Otmakhova$^{1*}$ \qquad  Hung Thinh Truong$^{1*}$ \qquad Rahmad Mahendra $^3$  \\ \textbf{Zenan Zhai $^2$ {Rongxin Zhu $^{2}$ \qquad Daniel Beck $^4$ \qquad Jey Han Lau$^1$}}\\
$^1$The University of Melbourne\,\,\, $^2$Oracle\,\,\, $^3$Universitas Indonesia\,\,\,  $^4$RMIT University\\
\texttt{\{y.otmakhova, thinh.truong\}@unimelb.edu.au}\,\,\,  \texttt{rahmad.mahendra@cs.ui.ac.id} \\
\texttt{zenanzhai@gmail.com}\,\,\,
\texttt{rongxin.zhu@unimelb.edu.au}, \\
\texttt{daniel.beck@rmit.edu.au}, \,\,\,
\texttt{jeyhan.lau@unimelb.edu.au}
}

\begin{document}
\maketitle

\begin{abstract}
We present \texttt{FLUKE} (\underline{f}ramework for \underline{l}ing\underline{u}istically-driven and tas\underline{k}-agnostic robustness \underline{e}valuation), a framework for assessing model robustness through systematic minimal variations of test data. 
\texttt{FLUKE} introduces controlled variations across linguistic levels --- from orthography to dialect and style --- and leverages large language models (LLMs) with human validation to generate modifications.
We demonstrate \texttt{FLUKE}'s utility by evaluating both fine-tuned models and LLMs across six diverse NLP tasks (four classification and two generation tasks), and reveal that
(1) the impact of linguistic variations is highly task-dependent, with some tests being critical for certain tasks but irrelevant for others; (2) LLMs still exhibit significant brittleness to certain linguistic variations, with reasoning LLMs surprisingly showing less robustness on some tasks compared to base models, and scaling improving robustness only for surface-level modifications; (3) models are overall more brittle to natural, fluent modifications such as syntax or style changes (and especially to negation), compared to corruption-style tests such as letter flipping; (4) the ability of a model to use a linguistic feature in generation does not correlate to its robustness to this feature on downstream tasks. These findings highlight the importance of systematic robustness testing for understanding model behaviors.

\end{abstract}

\section{Introduction}

The standard evaluation paradigm for machine learning models is to 
assess their performance on test data that shares the same distribution to the training set.
Language evaluation benchmarks (\citet{Nie+:2020,Zellers+:2019,Wang+:2018,Wang+:2019}; inter alia) are typically created based on such an assumption\footnote{An exception here is benchmarks specifically designed for evaluating large language models, which typically have no training data \citep{Zhong+:2024,Hendrycks+:2021,Srivastava+:2022}.} and over the years we have seen a trend of increasing model performance;
 in some instances the state-of-the-art models are on par with human performance \citep{Hupkes+:2023,LeBras+:2020}. Despite this positive trend, many studies found that models still perform poorly when dealing with minor variations \citep{Ribeiro+:2020,McCoy+:2019}, minimally contrastive pairs \citep{Sennrich+:2017,Warstadt+:2020}, out-of-domain data \citep{Singhal+:2023}, and adversarial input \citep{Jia+:2017,Ebrahimi+:2018}. This reveals that the success of NLP models may not be solely attributed to improved linguistic understanding and that they may have learned the data rather than the task \citep{Li+:2023,Hupkes+:2023,Linzen+:2020}.

 In this paper, we introduce \texttt{FLUKE} (\underline{f}ramework for \underline{l}ing\underline{u}istically-driven and tas\underline{k}-agnostic robustness \underline{e}valuation), a task-agnostic framework to generate minimal modifications of existing test data based on linguistic variations to evaluate model robustness.\footnote{Under the generalisation taxonomy of \citet{Hupkes+:2023}, our modifications target specifically covariate shifts.} That is, an original input (\textit{everyone loves dogs}) is modified to examine a model's robustness in handling linguistic variations such as orthography (\textit{\textbf{evryone} loves dogs}), syntax (\textit{\textbf{dogs are loved by everyone}}), semantics (\textit{\textbf{no one} loves dogs}), etc. These modifications are generated by prompting a large language model (LLM), and then validated manually to assess whether they target the intended linguistic property 
and require updating the ground-truth labels.

To demonstrate the generalisability of \texttt{FLUKE}, we apply it to modify instances for four classification tasks of various complexity --- coreference resolution, dialogue contradiction detection, named entity recognition (NER), and sentiment analysis --- as well as two generation tasks --- instruction-following evaluation (IFEval) and mathematical reasoning (GSM).
We then test a number of models --- including fine-tuned pre-trained models (PLMs: BERT \citep{Devlin+:2019}, GPT-2 \citep{Radford+:2019} and T5 \citep{Raffel+:2020}), base LLMs (GPT-4o, Llama 3.1-405b, and Claude 3.5),
as well as reasoning LLMs (GPT-5 and DeepSeek R1) --- and examine their predictions in the original and modified instances. When measuring the overall \textit{robustness} of the model, we account for two cases: where the label of the instance is supposed to stay the same after modification and where the label changes. Accordingly, we consider the model to be robust if its predictions either remain the same or change together with the label. We find that: (1) Robustness to modifications highly depends on the task. 
(2) Even though LLMs are more robust than PLMs, they are still brittle to many modifications, and for some tasks the brittleness is surprisingly more prominent in reasoning LLMs. (3) Scaling helps to improve robustness only to surface-level modifications. (4) Natural, linguistically equivalent modifications are more damaging than traditional adversarial modifications such as random punctuation insertion; and (5) Models can be robust to linguistic features that they struggled to understand when generating, and brittle to features that are easy for them to employ. Thus, using only a handful of well-known tests may fail to reveal the brittleness of a model, highlighting the necessity of a comprehensive, task-agnostic testing framework like \texttt{FLUKE}. To summarise, our contributions are:
\begin{itemize}
    \item We present \texttt{FLUKE}, a task-agnostic framework to generate minimal modifications of test data based on linguistic variations. We design prompts to automate the modifications, and perform extensive quality checks, showing that this approach works well without substantial human effort for most modifications.

    \item We demonstrate the utility of \texttt{FLUKE} by evaluating the performance of multiple NLP models, and show that a comprehensive, linguistically driven approach helps to reveal their task-specific shortcomings. 
    
    \item We release our prompt templates, evaluation scripts, and validated modifications, to enable other researchers to test their models and tasks.\footnote{\url{https://fluke-nlp.github.io/}} We also publish a dashboard that allows to assess models performance by comparing their outputs on original and modified instances.\footnote{\url{https://fluke-viewer.pages.dev/}}
    
\end{itemize}

\section{Related Work}

\paragraph{Task-based Benchmarking}

Much of previous work in evaluating models is centered around collating
datasets that are representative of a range of typical NLP
applications \citep{Wang+:2018, Song2023}. The datasets
in these benchmarks usually follow the typical paradigm
of having an in-domain test set, coming from the same distribution as
the training data. While this is useful in many experimental
settings (such as model comparison), it is restricted to
reporting in-domain performance.
The importance of evaluation in out-of-domain scenarios is well
known. Many shared tasks in NLP include test sets that are unrelated
to the training data \cite[][{\em inter alia}]{Pei2023,Ousidhoum2024}. The development of so-called ``challenge sets'',
initially explored in machine translation \citep{Goyal2022} but now
extended to LLMs \citep{Zhong+:2024} also falls into this space. However, the range of domain differences is
massive, which explains why it is hard to create
``general'', collated benchmarks for out-of-domain scenarios.

\paragraph{Linguistic Competence Probing}

Another way to assess
models is through linguistic competence. A common approach is to use
{\em probing classifiers}, tuned on datasets
tailored to specific competences, such as morphology or syntax
\citep{Blevins2023, Mahowald2024}.
Holmes \citep{Waldis2024} attempt to unify multiple datasets created to test
linguistic competences. While probing for linguistic competence can provide insights about the model's understanding of specific
linguistic phenomena, it is important to note that model competence does not necessarily translate to task performance, as we show below in \Cref{sec:discussion_llm_generated_data}. 

\paragraph{Robustness Evaluation}

One line of work
explores the idea of {\em minimal contrastive
  pairs}
\citep{Kann+:2019,Linzen+:2018,Warstadt+2019,Linzen+:2016},
to evaluate a model's robustness to a specific
phenomena in the context of a specific task (e.g.\ subject-verb agreement in syntax
\citep{Warstadt+:2020} or novel words in machine translation
\citep{Sennrich+:2017}).
Another line focuses on creating {\em adversarial examples} to understand model failures \citep{eger-benz-2020-hero,keller-etal-2021-bert,tan-etal-2020-morphin,formento-etal-2023-using,li-etal-2024-gsm}.
These studies use model loss or gradient to learn how to create perturbations of test data that break the model.

\citet{Ribeiro+:2020} introduce the Checklist framework to test linguistic capabilities, but it is not fully
task-agnostic, as it requires developers to design their own
suite of tests (which, as we show below, may not fully reveal model brittleness), and uses templates to create modifications, which limits their diversity. 
To address these limitations, we propose a comprehensive suite of varied and natural linguistic modifications together with LLMs prompts that are ready to be applied to any task.

\section{\texttt{FLUKE}}
\label{sec:tests}


\texttt{FLUKE} systematically covers phenomena related to different levels of language structure, as well as some model biases. Simplified examples of each test are in Table \ref{tab:fluke}.

\begin{table*}
    \centering
\begin{adjustbox}{max width=\linewidth}

    \begin{footnotesize}
      \begin{tabular}{llp{5cm}p{5cm}}
        \toprule
        Test & Subtest & Original & Modified \\
        \midrule
        \midrule
        {\bf Orthography} &&&\\
        \midrule
        \multirow{3}{*}{Spelling} & Addition & beautiful & beautiful\textbf{l} \\
             & Omission & fantastic & fantstic\\
             & Swapping & not a bid deal & not a big d\textbf{ae}l \\
        \midrule
        \multirow{4}{*}{Capitalization}  & Upper to lower & Battlefield 3 & \textbf{b}attlefield 3 \\
             &  Lower to upper & Hinckley did not hit Reagan & Hinckley did not \textbf{H}it Reagan\\
             &  sPoNgEcAsE & The professor & The pRoFeSsOr\\
             & ALL CAPS & Michael Phelps & MICHAEL Phelps \\
        \midrule
        \multirow{3}{*}{Punctuation}  & Add & not exactly the bees knees & not exactly, the bees knees \\
             & Change & It's worth tracking down. & It's worth tracking down! \\
             & Remove space & so little movie & so littlemovie \\
        \midrule
        \midrule
        {\bf Morphology} &&&\\
        \midrule
        Derivation & Derived & killed & assassinated \\
        Compound & Compound & new & brand-new \\
        \midrule
        \midrule
        {\bf Syntax} &&&\\
        \midrule
        Voice & Active to passive & Billy beat Tommy & Tommy was beaten by Billy \\
        Gramm. role & Entity swap & Bob sued Bill & Bill sued Bob \\
        Conjunctions & Coordinating & excessive hunting & excessive hunting and poaching \\
        \midrule
        \midrule
        {\bf Semantics} &&&\\
        \midrule
        \multirow{4}{*}{Concept} & Synonym & suspect & doubt \\
             & Hyper/hyponym & organization & association \\
             & Nonce word & The bowl had a crack & The bowl had a vibble\\
             & Idiom & they were tasty & they were a real treat\\ 
        \midrule
                \multirow{5}{*}{Negation} & Verbal & They were afraid of the robots & They were \textbf{not} afraid of the robots \\
             & Absolute & They were afraid of the robots & They were afraid of \textbf{none} of the robots\\
             & Approximate & They were afraid of  the robots & They were \textbf{seldom} afraid of the robots\\
             & Lexical & They were afraid of the robots & They were \textbf{fearless} of the robots \\
             & Double & They were afraid of the robots & They were \textbf{not unafraid} of the robots \\
        \midrule
        \midrule
        {\bf Discourse}&&&\\
        \midrule

        \multirow{3}{*}{\shortstack[l]{Discourse\\ markers}} & Addition & Toyota has Lexus: they are built for the rich. & Toyota has Lexus, and they are built for the rich. \\
             & Change & The boss fired the worker when he stopped performing well. & The boss fired the worker after he stopped performing well. \\
             & Remove & Tony helped Jeff as he needed help. & Tony helped Jeff, he needed help.\\
        \midrule
        Appraisal & Addition & She turns her down. & She coldly turns her down.\\
        \midrule
        \midrule
        {\bf Varieties} &&&\\
        \midrule
        Style & Casual & There is no pleasure in watching a child
                         suffer & It's no fun seeing a kid suffer \\
        \midrule
        Dialect & Singlish & He would not say no. & He dun wan say no.\\
        \midrule
        \midrule
        {\bf Biases} &&&\\
        \midrule
        Temporal & Old-fashioned & He treats her badly. & He treats her ill. \\
        \midrule
        \multirow{2}{*}{Geographical} & Names & Anna tried again & Dongxin tried again \\
             & Cultural entities & The bat hit the ball & The lakau hit the polo\\
        \midrule
        \multirow{2}{*}{Length} & Shorten & The lion saw the fish and it was swimming & The lion saw the fish swimming.\\
             & Lengthen & Joseph did not defeat William & Joseph did not manage to defeat William\\
        \bottomrule
         
    \end{tabular}

      \end{footnotesize}
          \end{adjustbox}

    \caption{\texttt{FLUKE}: our proposed linguistically-driven tests.}
    \label{tab:fluke}

\end{table*}

\subsection{Linguistic tests}

We design tests for all levels of language structure, starting from low-level {orthography} tests and ending with dialect and style modifications.

\paragraph{Orthography}
\label{sec:orthography}

Here we test the ability of models to deal with changes in \textbf{spelling} (\textit{adding} or \textit{omitting} a letter, as well as \textit{swapping} two letters), \textbf{capitalization} (modifying the register to \textit{lower case}, \textit{Upper case}, \textit{sPoNgEcAsE} or \textit{ALL CAPS}) and \textbf{punctuation} (where we \textit{add} or \textit{change} a punctuation mark, or \textit{remove the space} between two words).

\paragraph{Morphology}

We test if more complex words are harder for models. In \textbf{derivation} modification, we change a non-derived word (a word without any suffixes or prefixes)
into a derived word. For \textbf{compound} modifications, we replace a word which has one root with a word that has several stems. 

\paragraph{Syntax}

Here we test if a model adjusts to changing the sentence structure. In particular, we change a verb from \textbf{active to passive} (the subject and object are flipped accordingly to preserve the meaning). For \textbf{grammatical role} tests, we leave the verb intact, but swap two entities. 
For \textbf{conjunctions} modification we add an extra word combined with a coordinating conjunction \textit{and} or \textit{or}.

\paragraph{Semantics}

We devise tests that involve different degrees of meaning change. First, in \textbf{concept} modifications, we replace words with \textit{synonyms}, with \textit{hypernyms} or \textit{hyponyms} (words that have a broader or narrower meaning), with artificially created, non-existing \textit{nonce words}\footnote{We use a list of nonce words from \citet{cremers2022interpreting}.}, and with \textit{idioms}. We also use \textbf{negation} to test for the change of meaning that goes beyond individual words. Apart from \textit{verbal} negation (adding \textit{not} to the verb), we consider less straightforward types of negation: \textit{approximate}, where the meaning is only partially negated, \textit{absolute}, which explicitly excludes non-negative interpretation, as well as  \textit{lexical} negation (affixal negation and antonymy) \citep{pullum_huddleston_huddleston_pullum_2002,otmakhova2022m3}. Finally, we look at \textit{double} negation, i.e. grammatically valid structures which contain two negation elements \citep{truong2022not}.

\paragraph{Discourse}
  
We look at sensitivity to manipulating \textbf{discourse markers}: we \textit{add} a discourse marker, explicitly connecting ideas, \textit{change} a discourse marker, which can lead to a different meaning, or \textit{delete} it, making the text more difficult to interpret.
We also test robustness to adding markers of \textbf{appraisal}: sentiment-bearing words or phrases. 

\paragraph{Language Varieties}

In these tests the changes are more extensive and span across all linguistic aspects. Specifically, we change the \textbf{style} to highly \textit{casual} English, and rewrite the sentence in a particular \textbf{dialect:} Singaporean creole English (\textit{Singlish}), which has systematic differences from American or British English varieties used in  benchmarks.

\subsection{Bias tests}
\label{sec:bias-test}

In addition to linguistic tests, we include tests for data artifacts that models are known to exploit. We test if changing words to their old-fashioned variants affects the performance (\textbf{temporal bias}). When testing for \textbf{geographical bias} \citep{faisal-anastasopoulos-2023-geographic,godey-etal-2024-scaling}, we randomly select from a list of regions that are likely to be underrepresented in training corpora.\footnote{Africa, Middle East, Southeast, East and Central Asia, Oceania, Latin America, Eastern Europe} Then we change proper nouns to names and locations specific to such regions, and replace common nouns with relevant cultural entities. Finally, in \textbf{length bias} tests we either \textit{shorten} or \textit{lengthen} the text.

\section{Task Evaluation}
\label{sec:use_cases}

We apply modifications to test model robustness for six tasks, including four classification
tasks (coreference resolution, dialogue contradiction detection, named entity recognition, sentiment analysis) and two generation tasks (instruction-following evaluation, IFEval, and mathematical reasoning, GSM). 
\subsection{Test set creation and validation}
\label{sec:test_creation}

We generate modifications (\Cref{sec:tests}) for each task, randomly selecting samples from a \textit{test partition} of the original dataset and prompting GPT-4o with instructions. We create and evaluate a general prompt template for each modification  (see \Cref{app:data_gen_prompt} for an example), and then use the same prompt template for each task with minimal modifications (such as specifying the text type). 

We follow several steps to ensure the high quality of modifications. To improve their \textbf{diversity}, we stratify modifications across subtests (for example, different types of negation) and, where possible, randomly select a variable from a list (such as region and country for \textit{geographical bias} or non-existing word for \textit{nonce} semantic test) and pass it to the prompt. To ensure that modifications were applied \textbf{minimally}, i.e. the difference in performance can be attributed to a particular linguistic feature change rather than to a large number of changes, we calculate Levenshtein distance between original and modified samples, discarding those which changed more than expected. Finally, we employ thorough human checks to ensure that modifications are \textbf{valid and meaningful}. For \textit{classification tasks} we ran this validation in two stages on Prolific\footnote{\url{https://www.prolific.com/}}, with thorough quality control and four annotators per sample (see \Cref{app:annotation} for annotation details, inter-annotator agreement and quality statistics). During the first stage, we check if resulting texts \textbf{have the required modification} and are still \textbf{fluent}, rejecting samples that do not meet these criteria. We find that most tests have a \textit{high retain rate} of 70-90\%. During the second stage, we ask annotators to select the tasks' label (such as positive or negative for sentiment analysis) for modified samples, without knowing the labels for original samples, to \textbf{check if the ground-truth label changed} (e.g.\ after applying negation the sentiment can change from positive to negative). To make sure the task is \textbf{solvable} after modification, i.e. a label is not ambiguous or hard to choose, we add an option such as ``Neutral'' for sentiment analysis or ``Not sure'' for other tasks, and reject samples where annotators chose it. 

For \textit{generation tasks}, to demonstrate how the framework can be applied with minimal human intervention, we skip the first stage of human checks and only use tests with reliable modifications (over 70\% retain rate). For the second stage, the authors of the paper validate modified samples in terms of whether the modification \textbf{changes the label} (for example, if it changes the premises in the math task, leading to a different answer) and whether the task is still \textbf{solvable} (for example, if there are modifications that prevent correct instruction following). Examples of such valid and rejected modifications, as well as modifications that led to label change, are shown in \Cref{tab:examples} in Appendix.
Such setup reduces the human effort, making the framework \textit{generalizable to new tasks}, while still ensuring the validity of modifications.

In total, we produce around 100 modified instances for each test in each task (1700 modified instances for each classification task and 1300 instances per generation task).

\subsection{Tested models and evaluation}

We experiment with three smaller PLMs (BERT, GPT-2, and T5), three base LLMs (GPT-4o, Llama 3.1 and Claude 3.5), and two reasoning LLMs (GPT-5, DeepSeek R1). Both PLMs and LLMs are tested on classification tasks, while only LLMs are used for generation tasks. The pretrained models are fine-tuned on the respective training set with an additional task-specific classification layer on top (\Cref{app:plms} in Appendix). The LLMs are prompted with task-related instructions in a zero-shot way (\Cref{app:llm_prompt}).

To ensure a fair comparison of performance between the original and modified set, for each test we evaluate only the subset of the original samples that corresponds to the modified instances. We measure instability via an ``unrobustness'' score
$U$ computed on paired original–modified examples. Let $o_i,m_i$ be per-sample correctness indicators for the original and modified items, respectively.\footnote{Correctness is defined as whether the prediction matches the ground truth label. Note that we use this formulation because it handles the situations where there is a label change after modification (e.g.\ sentiment has flipped after negation is introduced). In those cases, the prediction will also need to change for it to be ``correct''.} Then
  \begin{equation}
  \label{eq:unrob-classif}
  U = \frac{1}{N} \sum_{i=1}^{N} \big| m_i - o_i \big| \cdot 100
  \end{equation}
where $N$ is the number of instances. Note that $U$ counts both degradations (1→0) and improvements (0→1) as ``unrobust'': our rationale is that if after modification the model correctly predicts a previously incorrect sample, it is still brittle since the model relies on a particular way the text is written rather than understanding its content.

  We follow the original metrics used for the task to define  correctness. For dialogue contradiction detection, coreference resolution, sentiment analysis, and GSM the correctness metric is accuracy. For NER, we use per-sample entity-level mean F1 \citep{Nakayama+:2018}, while IFEval employs strict success (all constraints satisfied) as correctness.

\section{Results}
\subsection{Use Case: NER and Coreference tasks}

In this section we present two classification tasks: named entity extraction (NER) and coreference resolution (Coreference).\footnote{The results for the other two classification tasks -- Sentiment analysis and Dialogue Contradiction resolution -- are in \Cref{app:dialogue} due to space constraints.} For Coreference task, we use KnowRef dataset \citep{emami-etal-2019-knowref}, which consists of pronoun disambiguation problems 
where each sample has the Winograd Schema Challenge \citep{wsc} format, consisting of a sentence, the pronoun we want to disambiguate, two candidates, and the index of the correct candidate as label. For NER, we chose Few-NERD~\citep{ding-etal-2021-nerd}, a dataset of Wikipedia sentences, using its 8 first-level entity categories, i.e., \textit{Person, Location, Organization, Art, Building, Product, Event, Misc}. As explained in \Cref{sec:test_creation}, we thoroughly check and update the labels to make sure the task is still valid and solvable after modification (for example, check if the pronoun can still be resolved or if the entity has changed).

Interestingly, unrobustness results for NER (see \Cref{tab:ner_unrob}) and Coreference (\Cref{tab:coref_unrob}) show different, almost orthogonal trends in terms of the tests the models were brittle to. For NER the most revealing test is \textbf{Geographical bias}, where all models predict entities from some locales better than from the others. While in some cases the models struggled to identify entities after modification, in the majority of cases it was actually easier for the models to predict ``exotic'' entities from non-English locales. Moreover, almost all models (except for BERT) lacked robustness to \textbf{Punctuation} modifications, heavily relying on \textbf{capitalization} as a cue for named entity presence or type. Interestingly, while base LLMs overall are more robust to modifications on this task, reasoning LLMs (especially DeepSeek) show major fluctuations across all NER tests, demonstrating similar unrobustness to PLMs.

Conversely, for the Coreference task, while models were relatively robust to the tests affecting NER, they demonstrated major instability on tests involving modifications of syntactical structure and meaning, such as change of active \textbf{Voice} to passive, swapping \textbf{Grammatical roles}, or introducing \textbf{Negation}, as well as perturbations involving significant changes of context such as \textbf{Length bias},  change to casual \textbf{Style}, or to \textbf{Singlish} dialect. Unlike NER, on Coreference tasks LLMs (with the exception of GPT-4o) perform better than PLMs; however, they still demonstrate unrobustness to the majority of tests, with even the best model (GPT-5) significantly affected by grammatical roles change, negation and style modifications. 


\begin{table}[t]
\centering
\resizebox{\linewidth}{!}{
\begin{tabular}{llrrrrrrrrr}
\toprule
Category & Modification & \multicolumn{3}{c}{\textbf{PLM}} & \multicolumn{5}{c}{\textbf{LLM}} & \textbf{Avg} \\
 &  & \textbf{BERT} & \textbf{GPT2} & \textbf{T5} & \textbf{GPT4o} & \textbf{Claude} & \textbf{Llama} & \textbf{GPT5} & \textbf{DS} &  \\
\midrule
\textbf{Bias} & \textbf{Temporal} & \cellcolor{blue!10} 3.7 & \cellcolor{blue!8} 3.0 & \cellcolor{blue!3} 1.2 & \cellcolor{blue!16} 6.2 & \cellcolor{blue!11} 4.3 & \cellcolor{blue!5} 1.8 & \cellcolor{blue!20} 7.7 & \cellcolor{blue!27} 10.6 & \cellcolor{blue!12} 4.8 \\
  & \textbf{Geographical} & \cellcolor{blue!65} \textcolor{white}{25.1} & \cellcolor{blue!71} \textcolor{white}{27.6} & \cellcolor{blue!75} \textcolor{white}{29.0} & \cellcolor{blue!58} \textcolor{white}{22.3} & \cellcolor{blue!67} \textcolor{white}{26.0} & \cellcolor{blue!70} \textcolor{white}{27.0} & \cellcolor{blue!57} \textcolor{white}{22.0} & \cellcolor{blue!72} \textcolor{white}{27.8} & \cellcolor{blue!67} \textcolor{white}{25.8} \\
  & \textbf{Length} & \cellcolor{blue!30} 11.8 & \cellcolor{blue!32} 12.2 & \cellcolor{blue!34} 13.1 & \cellcolor{blue!33} 12.7 & \cellcolor{blue!41} 15.7 & \cellcolor{blue!17} 6.4 & \cellcolor{blue!25} 9.7 & \cellcolor{blue!33} 12.6 & \cellcolor{blue!30} 11.8 \\
\midrule
\textbf{Orthogr.} & \textbf{Spelling} & \cellcolor{blue!11} 4.3 & \cellcolor{blue!6} 2.5 & \cellcolor{blue!4} 1.4 & \cellcolor{blue!17} 6.6 & \cellcolor{blue!8} 3.0 & \cellcolor{blue!10} 3.9 & \cellcolor{blue!19} 7.3 & \cellcolor{blue!29} 11.4 & \cellcolor{blue!13} 5.0 \\
  & \textbf{Capitalization} & \cellcolor{blue!2} 0.9 & \cellcolor{blue!50} \textcolor{white}{19.3} & \cellcolor{blue!34} 13.1 & \cellcolor{blue!29} 11.3 & \cellcolor{blue!23} 8.9 & \cellcolor{blue!40} 15.3 & \cellcolor{blue!24} 9.1 & \cellcolor{blue!35} 13.6 & \cellcolor{blue!30} 11.5 \\
  & \textbf{Punctuation} & \cellcolor{blue!17} 6.5 & \cellcolor{blue!10} 3.7 & \cellcolor{blue!13} 4.9 & \cellcolor{blue!18} 7.1 & \cellcolor{blue!24} 9.1 & \cellcolor{blue!20} 7.9 & \cellcolor{blue!20} 7.8 & \cellcolor{blue!39} 15.0 & \cellcolor{blue!20} 7.8 \\
\midrule
\textbf{Morphol.} & \textbf{Derivation} & \cellcolor{blue!5} 1.9 & \cellcolor{blue!11} 4.2 & \cellcolor{blue!15} 5.8 & \cellcolor{blue!9} 3.7 & \cellcolor{blue!5} 2.0 & \cellcolor{blue!3} 1.3 & \cellcolor{blue!25} 9.5 & \cellcolor{blue!21} 8.2 & \cellcolor{blue!12} 4.6 \\
  & \textbf{Compound} & \cellcolor{blue!8} 3.1 & \cellcolor{blue!2} 0.6 & \cellcolor{blue!13} 5.2 & \cellcolor{blue!8} 3.1 & \cellcolor{blue!4} 1.7 & \cellcolor{blue!3} 1.0 & \cellcolor{blue!18} 7.0 & \cellcolor{blue!33} 12.9 & \cellcolor{blue!11} 4.3 \\
\midrule
\textbf{Syntax} & \textbf{Voice} & \cellcolor{blue!20} 7.8 & \cellcolor{blue!28} 10.8 & \cellcolor{blue!15} 5.7 & \cellcolor{blue!19} 7.5 & \cellcolor{blue!14} 5.5 & \cellcolor{blue!11} 4.5 & \cellcolor{blue!21} 8.3 & \cellcolor{blue!29} 11.3 & \cellcolor{blue!20} 7.7 \\
  & \textbf{Grammar} & \cellcolor{blue!21} 8.0 & \cellcolor{blue!41} 15.9 & \cellcolor{blue!27} 10.5 & \cellcolor{blue!21} 8.3 & \cellcolor{blue!9} 3.5 & \cellcolor{blue!14} 5.3 & \cellcolor{blue!28} 10.8 & \cellcolor{blue!32} 12.4 & \cellcolor{blue!24} 9.3 \\
  & \textbf{Conjunction} & \cellcolor{blue!24} 9.1 & \cellcolor{blue!20} 7.6 & \cellcolor{blue!20} 7.7 & \cellcolor{blue!24} 9.4 & \cellcolor{blue!19} 7.5 & \cellcolor{blue!21} 8.0 & \cellcolor{blue!30} 11.7 & \cellcolor{blue!34} 13.2 & \cellcolor{blue!24} 9.3 \\
\midrule
\textbf{Semantics} & \textbf{Concept} & \cellcolor{blue!13} 5.0 & \cellcolor{blue!23} 8.9 & \cellcolor{blue!14} 5.3 & \cellcolor{blue!17} 6.5 & \cellcolor{blue!12} 4.8 & \cellcolor{blue!10} 3.9 & \cellcolor{blue!22} 8.5 & \cellcolor{blue!24} 9.4 & \cellcolor{blue!17} 6.5 \\
  & \textbf{Negation} & \cellcolor{blue!12} 4.8 & \cellcolor{blue!14} 5.2 & \cellcolor{blue!17} 6.5 & \cellcolor{blue!21} 8.2 & \cellcolor{blue!10} 3.9 & \cellcolor{blue!12} 4.6 & \cellcolor{blue!27} 10.5 & \cellcolor{blue!39} 15.2 & \cellcolor{blue!19} 7.4 \\
\midrule
\textbf{Discourse} & \textbf{Disc. markers} & \cellcolor{blue!12} 4.7 & \cellcolor{blue!4} 1.6 & \cellcolor{blue!14} 5.4 & \cellcolor{blue!6} 2.2 & \cellcolor{blue!5} 1.9 & \cellcolor{blue!8} 3.2 & \cellcolor{blue!16} 6.3 & \cellcolor{blue!22} 8.7 & \cellcolor{blue!11} 4.2 \\
  & \textbf{Appraisal} & \cellcolor{blue!14} 5.5 & \cellcolor{blue!7} 2.6 & \cellcolor{blue!12} 4.8 & \cellcolor{blue!8} 3.2 & \cellcolor{blue!8} 3.1 & \cellcolor{blue!8} 2.9 & \cellcolor{blue!18} 7.1 & \cellcolor{blue!36} 14.0 & \cellcolor{blue!14} 5.4 \\
\midrule
\textbf{Varieties} & \textbf{Style} & \cellcolor{blue!30} 11.8 & \cellcolor{blue!32} 12.2 & \cellcolor{blue!34} 13.1 & \cellcolor{blue!9} 3.6 & \cellcolor{blue!12} 4.8 & \cellcolor{blue!17} 6.4 & \cellcolor{blue!19} 7.4 & \cellcolor{blue!28} 10.9 & \cellcolor{blue!23} 8.8 \\
  & \textbf{Dialectal} & \cellcolor{blue!33} 12.6 & \cellcolor{blue!19} 7.4 & \cellcolor{blue!22} 8.4 & \cellcolor{blue!15} 5.9 & \cellcolor{blue!23} 8.9 & \cellcolor{blue!12} 4.7 & \cellcolor{blue!18} 7.0 & \cellcolor{blue!34} 13.2 & \cellcolor{blue!22} 8.5 \\
\midrule
\textbf{Average} & & \cellcolor{blue!19} 7.4 & \cellcolor{blue!22} 8.6 & \cellcolor{blue!21} 8.3 & \cellcolor{blue!19} 7.5 & \cellcolor{blue!17} 6.7 & \cellcolor{blue!16} 6.4 & \cellcolor{blue!24} 9.3 & \cellcolor{blue!33} 13.0 & \cellcolor{blue!22} 8.4 \\
\bottomrule
\end{tabular}}
\caption{NER: Unrobustness (U, \%) by model and modification. \textit{Claude} stands for Claude 3.5, \textit{Llama} for Llama 3.1, and \textit{DS} for DeepSeek R1. Full results with confidence interval (CI) in \Cref{tab:ner_unrob_ci}.}
\label{tab:ner_unrob}
\end{table}

\begin{table}[t]
\centering
\resizebox{\linewidth}{!}{
\begin{tabular}{llrrrrrrrrr}
\toprule
Category & Modification & \multicolumn{3}{c}{\textbf{PLM}} & \multicolumn{5}{c}{\textbf{LLM}} & \textbf{Avg} \\
 &  & \textbf{BERT} & \textbf{GPT2} & \textbf{T5} & \textbf{GPT4o} & \textbf{Claude} & \textbf{Llama} & \textbf{GPT5} & \textbf{DS} & \\
\midrule
\textbf{Bias} & \textbf{Temporal} & \cellcolor{blue!23} 9.0 & \cellcolor{blue!10} 4.0 & \cellcolor{blue!15} 6.0 & \cellcolor{blue!21} 8.0 & \cellcolor{blue!3} 1.0 & \cellcolor{blue!18} 7.0 & \cellcolor{blue!8} 3.0 & \cellcolor{blue!21} 8.0 & \cellcolor{blue!15} 5.8 \\
  & \textbf{Geographical} & \cellcolor{blue!21} 8.0 & \cellcolor{blue!36} 14.0 & \cellcolor{blue!23} 9.0 & \cellcolor{blue!26} 10.0 & \cellcolor{blue!10} 4.0 & \cellcolor{blue!13} 5.0 & \cellcolor{blue!15} 6.0 & \cellcolor{blue!26} 10.0 & \cellcolor{blue!21} 8.2 \\
  & \textbf{Length} & \cellcolor{blue!50} \textcolor{white}{19.2} & \cellcolor{blue!39} 15.2 & \cellcolor{blue!36} 14.1 & \cellcolor{blue!39} 15.2 & \cellcolor{blue!52} \textcolor{white}{20.2} & \cellcolor{blue!44} 17.2 & \cellcolor{blue!21} 8.1 & \cellcolor{blue!34} 13.1 & \cellcolor{blue!39} 15.3 \\
\midrule
\textbf{Orthogr.} & \textbf{Spelling} & \cellcolor{blue!29} 11.2 & \cellcolor{blue!8} 3.1 & \cellcolor{blue!18} 7.1 & \cellcolor{blue!13} 5.1 & \cellcolor{blue!5} 2.0 & \cellcolor{blue!11} 4.1 & \cellcolor{blue!18} 7.1 & \cellcolor{blue!11} 4.1 & \cellcolor{blue!14} 5.5 \\
  & \textbf{Capitalization} & \cellcolor{blue!39} 15.2 & \cellcolor{blue!36} 14.1 & \cellcolor{blue!18} 7.1 & \cellcolor{blue!18} 7.1 & \cellcolor{blue!10} 4.0 & \cellcolor{blue!13} 5.1 & \cellcolor{blue!5} 2.0 & \cellcolor{blue!16} 6.1 & \cellcolor{blue!20} 7.6 \\
  & \textbf{Punctuation} & \cellcolor{blue!3} 1.0 & \cellcolor{blue!18} 7.1 & \cellcolor{blue!3} 1.0 & \cellcolor{blue!3} 1.0 & \cellcolor{blue!8} 3.0 & \cellcolor{blue!3} 1.0 & \cellcolor{blue!5} 2.0 & \cellcolor{blue!10} 4.0 & \cellcolor{blue!7} 2.5 \\
\midrule
\textbf{Morphol.} & \textbf{Derivation} & \cellcolor{blue!11} 4.1 & \cellcolor{blue!8} 3.1 & \cellcolor{blue!13} 5.1 & \cellcolor{blue!11} 4.1 & \cellcolor{blue!3} 1.0 & \cellcolor{blue!3} 1.0 & \cellcolor{blue!13} 5.1 & \cellcolor{blue!5} 2.0 & \cellcolor{blue!8} 3.2 \\
  & \textbf{Compound} & \cellcolor{blue!16} 6.2 & \cellcolor{blue!11} 4.2 & \cellcolor{blue!8} 3.1 & \cellcolor{blue!13} 5.2 & \cellcolor{blue!16} 6.2 & \cellcolor{blue!16} 6.2 & \cellcolor{blue!8} 3.1 & \cellcolor{blue!11} 4.2 & \cellcolor{blue!12} 4.8 \\
\midrule
\textbf{Syntax} & \textbf{Voice} & \cellcolor{blue!80} \textcolor{white}{35.8} & \cellcolor{blue!80} \textcolor{white}{34.7} & \cellcolor{blue!80} \textcolor{white}{41.1} & \cellcolor{blue!68} \textcolor{white}{26.3} & \cellcolor{blue!41} 15.8 & \cellcolor{blue!79} \textcolor{white}{30.5} & \cellcolor{blue!24} 9.5 & \cellcolor{blue!41} 15.8 & \cellcolor{blue!68} \textcolor{white}{26.2} \\
  & \textbf{Grammar} & \cellcolor{blue!79} \textcolor{white}{30.6} & \cellcolor{blue!72} \textcolor{white}{27.8} & \cellcolor{blue!50} \textcolor{white}{19.4} & \cellcolor{blue!57} \textcolor{white}{22.2} & \cellcolor{blue!47} \textcolor{white}{18.1} & \cellcolor{blue!50} \textcolor{white}{19.4} & \cellcolor{blue!43} 16.7 & \cellcolor{blue!54} \textcolor{white}{20.8} & \cellcolor{blue!56} \textcolor{white}{21.9} \\
  & \textbf{Conjunction} & \cellcolor{blue!13} 5.2 & \cellcolor{blue!27} 10.3 & \cellcolor{blue!21} 8.2 & \cellcolor{blue!21} 8.2 & \cellcolor{blue!11} 4.1 & \cellcolor{blue!19} 7.2 & \cellcolor{blue!13} 5.2 & \cellcolor{blue!16} 6.2 & \cellcolor{blue!18} 6.8 \\
\midrule
\textbf{Semantics} & \textbf{Concept} & \cellcolor{blue!21} 8.0 & \cellcolor{blue!8} 3.0 & \cellcolor{blue!13} 5.0 & \cellcolor{blue!28} 11.0 & \cellcolor{blue!46} \textcolor{white}{18.0} & \cellcolor{blue!23} 9.0 & \cellcolor{blue!26} 10.0 & \cellcolor{blue!26} 10.0 & \cellcolor{blue!24} 9.2 \\
  & \textbf{Negation} & \cellcolor{blue!66} \textcolor{white}{25.5} & \cellcolor{blue!61} \textcolor{white}{23.5} & \cellcolor{blue!63} \textcolor{white}{24.5} & \cellcolor{blue!55} \textcolor{white}{21.4} & \cellcolor{blue!63} \textcolor{white}{24.5} & \cellcolor{blue!58} \textcolor{white}{22.4} & \cellcolor{blue!58} \textcolor{white}{22.4} & \cellcolor{blue!79} \textcolor{white}{30.6} & \cellcolor{blue!63} \textcolor{white}{24.4} \\
\midrule
\textbf{Discourse} & \textbf{Disc. markers} & \cellcolor{blue!21} 8.0 & \cellcolor{blue!15} 6.0 & \cellcolor{blue!21} 8.0 & \cellcolor{blue!49} \textcolor{white}{19.0} & \cellcolor{blue!26} 10.0 & \cellcolor{blue!18} 7.0 & \cellcolor{blue!23} 9.0 & \cellcolor{blue!28} 11.0 & \cellcolor{blue!25} 9.8 \\
  & \textbf{Appraisal} & \cellcolor{blue!13} 5.0 & \cellcolor{blue!21} 8.0 & \cellcolor{blue!10} 4.0 & \cellcolor{blue!21} 8.0 & \cellcolor{blue!18} 7.0 & \cellcolor{blue!23} 9.0 & \cellcolor{blue!13} 5.0 & \cellcolor{blue!18} 7.0 & \cellcolor{blue!17} 6.6 \\
\midrule
\textbf{Varieties} & \textbf{Style} & \cellcolor{blue!41} 16.0 & \cellcolor{blue!49} \textcolor{white}{19.0} & \cellcolor{blue!46} \textcolor{white}{18.0} & \cellcolor{blue!49} \textcolor{white}{19.0} & \cellcolor{blue!36} 14.0 & \cellcolor{blue!36} 14.0 & \cellcolor{blue!36} 14.0 & \cellcolor{blue!31} 12.0 & \cellcolor{blue!41} 15.8 \\
  & \textbf{Dialect} & \cellcolor{blue!22} 8.3 & \cellcolor{blue!69} \textcolor{white}{26.9} & \cellcolor{blue!57} \textcolor{white}{22.2} & \cellcolor{blue!63} \textcolor{white}{24.5} & \cellcolor{blue!30} 11.8 & \cellcolor{blue!43} 16.7 & \cellcolor{blue!8} 2.9 & \cellcolor{blue!33} 12.7 & \cellcolor{blue!41} 15.8 \\
\midrule
\textbf{Average} & & \cellcolor{blue!33} 12.7 & \cellcolor{blue!34} 13.2 & \cellcolor{blue!31} 11.9 & \cellcolor{blue!33} 12.7 & \cellcolor{blue!25} 9.7 & \cellcolor{blue!28} 10.7 & \cellcolor{blue!20} 7.7 & \cellcolor{blue!27} 10.5 & \cellcolor{blue!29} 11.1 \\
\bottomrule
\end{tabular}}
\label{tab:coref:unrob}
\caption{Coreference: Unrobustness (U, \%) by model and modification. Notation as in \Cref{tab:ner_unrob}. Full results with CI in \Cref{tab:coref_unrob_ci}.}
\label{tab:coref_unrob}
\end{table}

\subsection{Use case: GSM and IFEval}

We choose two generation tasks --- mathematical reasoning and instruction following --- to demonstrate the applicability of FLUKE to testing generative models. For the mathematical reasoning task we choose Grade School Math (GSM8K) dataset \citep{cobbe2021training}, which contains mathematical problems expressed in natural language. To simplify the task, we use only the final computed answer (rather than the reasoning steps) to evaluate the solution correctness. For the instruction following task, we use the IFEval benchmark \citep{zhou2023instruction} which contains varied generation prompts (``Write me a letter...'') that include 25 types of verifiable instructions such as ``include word XXX at least 3 times" or ``use 2 bullet points". Due to the nature of these tasks, we evaluate only LLMs on them. We ensure that the tasks are solvable after the modification, and that the expected output (solution or generated text) is still the same for all tests except for some types of \textbf{Negation} (\textbf{verbal, absolute, lexical}) where the modification changes the condition of the task and thus the model is expected to generate a different answer (see \Cref{sec:test_creation}).

We find that on the GSM task (\Cref{tab:gsm_unrob}) models demonstrate major brittleness to \textbf{Negation} tasks, where all of them fail to take into account the changed or even reversed logic of the premise, showing memorisation effects. Base LLMs such as Claude-3.5 and especially Llama 3.1 are also affected by context changes, such as failing to accomodate for changed names and currencies in \textbf{Geographical bias} test, or explaining the problem in informal \textbf{style} or \textbf{Singlish} dialect. Interestingly, reasoning LLMs are more robust to such changes.

\begin{table}[t]
\centering
\resizebox{\linewidth}{!}{
\begin{tabular}{llrrrrrr}
\toprule
Category & Modification & \textbf{GPT-4o} & \textbf{Claude-3.5} & \textbf{Llama 3.1} & \textbf{GPT-5} & \textbf{DS R1} & \textbf{Avg} \\
\midrule
\textbf{Bias} & \textbf{Temporal} & \cellcolor{blue!3} 1.0 & \cellcolor{blue!5} 2.0 & \cellcolor{blue!10} 4.0 & \cellcolor{blue!0} 0.0 & \cellcolor{blue!3} 1.0 & \cellcolor{blue!4} 1.6 \\
  & \textbf{Geographical} & \cellcolor{blue!13} 5.0 & \cellcolor{blue!13} 5.0 & \cellcolor{blue!18} 7.0 & \cellcolor{blue!3} 1.0 & \cellcolor{blue!5} 2.0 & \cellcolor{blue!10} 4.0 \\
  & \textbf{Length} & \cellcolor{blue!10} 4.0 & \cellcolor{blue!5} 2.0 & \cellcolor{blue!5} 2.0 & \cellcolor{blue!0} 0.0 & \cellcolor{blue!5} 2.0 & \cellcolor{blue!5} 2.0 \\
\midrule
\textbf{Orthogr.} & \textbf{Spelling} & \cellcolor{blue!3} 1.0 & \cellcolor{blue!0} 0.0 & \cellcolor{blue!5} 2.0 & \cellcolor{blue!3} 1.0 & \cellcolor{blue!0} 0.0 & \cellcolor{blue!2} 0.8 \\
  & \textbf{Capitalization} & \cellcolor{blue!8} 3.0 & \cellcolor{blue!3} 1.0 & \cellcolor{blue!13} 5.0 & \cellcolor{blue!3} 1.0 & \cellcolor{blue!3} 1.0 & \cellcolor{blue!6} 2.2 \\
  & \textbf{Punctuation} & \cellcolor{blue!0} 0.0 & \cellcolor{blue!0} 0.0 & \cellcolor{blue!13} 5.0 & \cellcolor{blue!3} 1.0 & \cellcolor{blue!3} 1.0 & \cellcolor{blue!4} 1.4 \\
\midrule
\textbf{Semantics} & \textbf{Concept} & \cellcolor{blue!3} 1.0 & \cellcolor{blue!0} 0.0 & \cellcolor{blue!13} 5.0 & \cellcolor{blue!5} 2.0 & \cellcolor{blue!3} 1.0 & \cellcolor{blue!5} 1.8 \\
  & \textbf{Negation} & \cellcolor{blue!39} 15.0 & \cellcolor{blue!39} 15.0 & \cellcolor{blue!46} \textcolor{white}{18.0} & \cellcolor{blue!18} 7.0 & \cellcolor{blue!23} 9.0 & \cellcolor{blue!33} 12.8 \\
\midrule
\textbf{Discourse} & \textbf{Appraisal} & \cellcolor{blue!0} 0.0 & \cellcolor{blue!0} 0.0 & \cellcolor{blue!3} 1.0 & \cellcolor{blue!3} 1.0 & \cellcolor{blue!3} 1.0 & \cellcolor{blue!2} 0.6 \\
\midrule
\textbf{Varieties} & \textbf{Style} & \cellcolor{blue!10} 4.0 & \cellcolor{blue!8} 3.0 & \cellcolor{blue!15} 6.0 & \cellcolor{blue!10} 4.0 & \cellcolor{blue!8} 3.0 & \cellcolor{blue!10} 4.0 \\
  & \textbf{Dialect} & \cellcolor{blue!5} 2.0 & \cellcolor{blue!10} 4.0 & \cellcolor{blue!15} 6.0 & \cellcolor{blue!3} 1.0 & \cellcolor{blue!5} 2.0 & \cellcolor{blue!8} 3.0 \\
\midrule
\textbf{Syntactic} & \textbf{Conjunction} & \cellcolor{blue!0} 0.0 & \cellcolor{blue!3} 1.0 & \cellcolor{blue!10} 4.0 & \cellcolor{blue!3} 1.0 & \cellcolor{blue!3} 1.0 & \cellcolor{blue!4} 1.4 \\
  & \textbf{Voice} & \cellcolor{blue!5} 2.0 & \cellcolor{blue!5} 2.0 & \cellcolor{blue!15} 6.0 & \cellcolor{blue!3} 1.0 & \cellcolor{blue!5} 2.0 & \cellcolor{blue!7} 2.6 \\
\midrule
\textbf{Average} &  &  \cellcolor{blue!8} 2.9 & \cellcolor{blue!7} 2.7 & \cellcolor{blue!14} 5.5 & \cellcolor{blue!4} 1.6 & \cellcolor{blue!5} 2.0 & \cellcolor{blue!8} 2.9 \\
\bottomrule
\end{tabular}}
\caption{GSM: Unrobustness (U, \%) by model and modification. Notation as in \cref{tab:ner_unrob}. Full results with CI in \Cref{tab:gsm_unrob_ci}.}
\label{tab:gsm_unrob}
\end{table}

Compared to solving mathematical problems, the LLM's ability to follow instructions are much more affected by the way the instructions are written --- with both base and reasoning LLMs showing significant fluctuations in performance across all tests. This includes not only the ``usual suspects'' such as \textbf{Negation}, \textbf{Style} and \textbf{Dialect}, but also more innocuous changes such as adding an \textbf{appraisal} word or phrase (\textit{please repeat} -> \textit{please be so nice as to repeat}), adding a word with a \textbf{conjunction} (\textit{Rewrite the following text} -> \textit{Rewrite and transform the following text}), changing the case of a single word in instruction, etc. Here, reasoning models do not seem to handle modifications better than the base ones.

\begin{table}[ht]
\centering
\resizebox{\linewidth}{!}{
\begin{tabular}{llrrrrrr}
\toprule
Category & Modification & \textbf{GPT-4o} & \textbf{Claude-3.5} & \textbf{Llama 3.1} & \textbf{GPT-5} & \textbf{DS R1} & \textbf{Avg} \\
\midrule
\textbf{Bias} & \textbf{Temporal} & \cellcolor{blue!29} 11.1 & \cellcolor{blue!10} 4.0 & \cellcolor{blue!31} 12.1 & \cellcolor{blue!13} 5.1 & \cellcolor{blue!31} 12.1 & \cellcolor{blue!23} 8.9 \\
  & \textbf{Geographical} & \cellcolor{blue!23} 9.0 & \cellcolor{blue!31} 12.0 & \cellcolor{blue!36} 14.0 & \cellcolor{blue!23} 9.0 & \cellcolor{blue!26} 10.0 & \cellcolor{blue!28} 10.8 \\
  & \textbf{Length} & \cellcolor{blue!28} 11.0 & \cellcolor{blue!26} 10.0 & \cellcolor{blue!46} \textcolor{white}{18.0} & \cellcolor{blue!15} 6.0 & \cellcolor{blue!41} 16.0 & \cellcolor{blue!31} 12.2 \\
\midrule
\textbf{Orthogr.} & \textbf{Capitalization} & \cellcolor{blue!21} 8.1 & \cellcolor{blue!13} 5.1 & \cellcolor{blue!26} 10.1 & \cellcolor{blue!23} 9.1 & \cellcolor{blue!34} 13.1 & \cellcolor{blue!23} 9.1 \\
  & \textbf{Punctuation} & \cellcolor{blue!18} 7.1 & \cellcolor{blue!8} 3.0 & \cellcolor{blue!34} 13.1 & \cellcolor{blue!21} 8.1 & \cellcolor{blue!31} 12.1 & \cellcolor{blue!22} 8.7 \\
  & \textbf{Spelling} & \cellcolor{blue!8} 3.1 & \cellcolor{blue!11} 4.1 & \cellcolor{blue!19} 7.2 & \cellcolor{blue!16} 6.2 & \cellcolor{blue!21} 8.2 & \cellcolor{blue!15} 5.8 \\
\midrule
\textbf{Syntax} & \textbf{Conjunction} & \cellcolor{blue!28} 11.0 & \cellcolor{blue!18} 7.0 & \cellcolor{blue!15} 6.0 & \cellcolor{blue!21} 8.0 & \cellcolor{blue!28} 11.0 & \cellcolor{blue!22} 8.6 \\
  & \textbf{Voice} & \cellcolor{blue!23} 9.0 & \cellcolor{blue!23} 9.0 & \cellcolor{blue!26} 10.0 & \cellcolor{blue!15} 6.0 & \cellcolor{blue!39} 15.0 & \cellcolor{blue!25} 9.8 \\
\midrule
\textbf{Semantics} & \textbf{Concept} & \cellcolor{blue!23} 9.1 & \cellcolor{blue!18} 7.1 & \cellcolor{blue!29} 11.1 & \cellcolor{blue!18} 7.1 & \cellcolor{blue!18} 7.1 & \cellcolor{blue!21} 8.3 \\
  & \textbf{Negation} & \cellcolor{blue!62} \textcolor{white}{24.0} & \cellcolor{blue!57} \textcolor{white}{22.0} & \cellcolor{blue!62} \textcolor{white}{24.0} & \cellcolor{blue!59} \textcolor{white}{23.0} & \cellcolor{blue!59} \textcolor{white}{23.0} & \cellcolor{blue!60} \textcolor{white}{23.2} \\
\midrule
\textbf{Discourse} & \textbf{Appraisal} & \cellcolor{blue!18} 7.1 & \cellcolor{blue!11} 4.1 & \cellcolor{blue!29} 11.2 & \cellcolor{blue!18} 7.1 & \cellcolor{blue!16} 6.1 & \cellcolor{blue!18} 7.1 \\
\midrule
\textbf{Varieties} & \textbf{Style} & \cellcolor{blue!46} \textcolor{white}{18.0} & \cellcolor{blue!18} 7.0 & \cellcolor{blue!26} 10.0 & \cellcolor{blue!23} 9.0 & \cellcolor{blue!28} 11.0 & \cellcolor{blue!28} 11.0 \\
  & \textbf{Dialect} & \cellcolor{blue!34} 13.0 & \cellcolor{blue!34} 13.0 & \cellcolor{blue!31} 12.0 & \cellcolor{blue!28} 11.0 & \cellcolor{blue!39} 15.0 & \cellcolor{blue!33} 12.8 \\
\midrule
\textbf{Average} &  & \cellcolor{blue!28} 10.8 & \cellcolor{blue!21} 8.3 & \cellcolor{blue!32} 12.2 & \cellcolor{blue!23} 8.8 & \cellcolor{blue!32} 12.3 & \cellcolor{blue!27} 10.5 \\
\bottomrule
\end{tabular}}
\caption{IFEval: Unrobustness (U, \%) by model and modification. Notation in \Cref{tab:ner_unrob}. Full results with CI in \Cref{tab:ifeval_unrob_ci}.}
\label{tab:ifeval_unrob}
\end{table}
\section{Discussion}

\subsection{Model brittleness is task-specific}

We call our framework task-agnostic, since the same set of prompts can be used to generate a battery of tests applicable to all tasks. This, however, does not mean that all tests are meaningful for all the tasks. As the results above (and in \Cref{app:dialogue}) show, the tests that actually ``fired", or showed substantial drops or increases, are different for each task. Even test such as \textbf{negation} which is known to be difficult for models \citep{ravichander-etal-2022-condaqa,truong2022not} has almost no effect on the NER task. On the other hand, seemingly ``simple'' tests (\textbf{capitalization}) that have a negligible effect on most of the tasks led to some dramatic drops on the NER and IFEval tasks. Thus, our framework is task-agnostic because it can be applied to reveal lack of robustness on a new task without having a preconceived (and potentially wrong) notion of what tests are meaningful for it.

\subsection{LLMs are not always more robust than PLMs, and reasoning enhanced training does not always help}

In classification tasks, where PLMs were finetuned on the original dataset, one would expect larger fluctuations in performance compared to LLMs, as modifications are likely to change the data distribution and break the identical distribution assumption that the fine-tuning process is based on. This holds for two out of four classification tasks we tested --- Coreference resolution (\Cref{tab:coref_unrob}) and Sentiment analysis (\Cref{tab:sa_unrob} in \Cref{app:dialogue}) --- where PLMs are overall less robust to perturbations. However, on two other classification tasks --- NER (\Cref{tab:ner_unrob}) and Dialogue contradiction detection (\Cref{tab:dialogue_unrob} in \Cref{app:dialogue}) --- while base LLMs are overall more robust than PLMs, reasoning LLMs are as brittle as the latter. Moreover, while LLMs are on average (across all models and tests) better (see the ``Average'' row in \Cref{tab:all_tasks_unrob}), they are sometimes markedly brittle to a particular test. For example, Claude-3.5 is substantially more affected by \textbf{concept} modifications than any  PLMs on Coreference resolution, or GPT-4o affected by \textbf{grammatical role} modifications on Dialogue contradiction task.  

Comparing between base and reasoning LLMs across tasks does not reveal a clear winner either. Reasoning seems to improve overall robustness on tasks such as GSM (\Cref{tab:gsm_unrob}), Coreference (\Cref{tab:coref_unrob}) or Sentiment Analysys (\Cref{tab:sa_unrob}); however, on other tasks the reasoning models perform similar to base LLMs (IFEval, \Cref{tab:ifeval_unrob}) or even worse than them (NER, \Cref{tab:ner_unrob}, and Dialogue contradiction detection, \Cref{tab:dialogue_unrob}).

We also test if a reasoning model (GPT-5) is able to handle the modifications better if it is \textit{aware} of them. To do that, we include additional {context} into the prompt where we alert the model that the sample was modified, and explicitly specify the type of modification (e.g.\ by telling it that the active voice was changed to passive).  We found no consistent gain when we do this (see Appendix \Cref{tab:gpt5_ctx_unrob}).

\subsection{Scaling helps but with diminishing effects and only for superficial modifications}

To test if increasing the parameter size helps with robustness, we additionally compare Llama-3.1 405B (which was used in all experiments) with two smaller versions -- Llama-3.1-8B and Llama-3.1-70B -- on the GSM task (the results in \Cref{tab:gsm_scaling_unrob}). Overall, we observe a large improvement in robustness when going from 8B $\rightarrow$ 70B but observe diminishing effects when going beyond that. Low-level \textbf{orthographic} and \textbf{syntactic} modifications benefit strongly from increased model size. In contrast, more complex perturbations, especially \textbf{negation}, remain highly challenging even at 405B. This suggests that scaling primarily improves surface-level robustness, while deeper semantic changes remain brittle.

\begin{table}[ht]
\centering
\resizebox{\linewidth}{!}{
\begin{tabular}{llrrrr}
\toprule
Category & Modification & \textbf{Llama-8b} & \textbf{Llama-70b} & \textbf{Llama-405b} & \textbf{Avg} \\
\midrule
\textbf{Bias} & \textbf{Temporal} & \cellcolor{blue!21} 8.0  & \cellcolor{blue!13} 5.0  & \cellcolor{blue!10} 4.0  & \cellcolor{blue!15} 5.7 \\
  & \textbf{Geographical} & \cellcolor{blue!21} 8.0  & \cellcolor{blue!8} 3.0  & \cellcolor{blue!18} 7.0  & \cellcolor{blue!15} 6.0  \\
  & \textbf{Length} & \cellcolor{blue!21} 8.0  & \cellcolor{blue!5} 2.0  & \cellcolor{blue!5} 2.0  & \cellcolor{blue!10} 4.0  \\
  \midrule
\textbf{Orthogr.} & \textbf{Spelling} & \cellcolor{blue!26} 10.0  & \cellcolor{blue!5} 2.0  & \cellcolor{blue!5} 2.0  & \cellcolor{blue!12} 4.7 \\
  & \textbf{Capitalization} & \cellcolor{blue!28} 11.0  & \cellcolor{blue!5} 2.0  & \cellcolor{blue!13} 5.0  & \cellcolor{blue!15} 6.0  \\
  & \textbf{Punctuation} & \cellcolor{blue!13} 5.0  & \cellcolor{blue!3} 1.0  & \cellcolor{blue!13} 5.0  & \cellcolor{blue!9} 3.7 \\
  \midrule
\textbf{Syntax} & \textbf{Conjunction} & \cellcolor{blue!31} 12.0  & \cellcolor{blue!5} 2.0  & \cellcolor{blue!10} 4.0  & \cellcolor{blue!15} 6.0  \\
  & \textbf{Voice} & \cellcolor{blue!28} 11.0  & \cellcolor{blue!5} 2.0 & \cellcolor{blue!15} 6.0  & \cellcolor{blue!16} 6.3  \\
  \midrule
\textbf{Semantic} & \textbf{Concept} & \cellcolor{blue!31} 12.0  & \cellcolor{blue!3} 1.0  & \cellcolor{blue!13} 5.0  & \cellcolor{blue!15} 6.0  \\
  & \textbf{Negation} & \cellcolor{blue!57} \textcolor{white}{22.0} & \cellcolor{blue!44} 17.0 & \cellcolor{blue!46} \textcolor{white}{18.0} & \cellcolor{blue!49} \textcolor{white}{19.0} \\
  \midrule
\textbf{Discourse} & \textbf{Appraisal} & \cellcolor{blue!18} 7.0  & \cellcolor{blue!13} 5.0  & \cellcolor{blue!3} 1.0  & \cellcolor{blue!11} 4.3  \\
\midrule
\textbf{Varieties} & \textbf{Style} & \cellcolor{blue!41} 16.0  & \cellcolor{blue!15} 6.0  & \cellcolor{blue!15} 6.0  & \cellcolor{blue!24} 9.3  \\
  & \textbf{Dialect} & \cellcolor{blue!26} 10.0  & \cellcolor{blue!15} 6.0 & \cellcolor{blue!15} 6.0  & \cellcolor{blue!19} 7.3 \\
\midrule
\textbf{Average} &  & \cellcolor{blue!28} 10.8 & \cellcolor{blue!11} 4.2  & \cellcolor{blue!14} 5.5  & \cellcolor{blue!18} 6.8  \\
\bottomrule
\end{tabular}}
\caption{GSM Scaling: Unrobustness (U, \%) by Llama model size and modification. Notation in \Cref{tab:ner_unrob}. Full results with CI in \Cref{tab:gsm_scaling_unrob_ci}.}
\label{tab:gsm_scaling_unrob}
\end{table}

\subsection{Models are overall less robust to natural, linguistically valid modifications}

Some of the work on robustness focuses on corrupting the text in adversarial way, such as shuffling characters or changing the case \citep{belinkov2017synthetic,li2018textbugger,eger-benz-2020-hero}, while some includes natural, lingustically equivalent perturbations such as synonym replacement \citep{ren2019generating,jin2020bert}. Our framework combines both types of tests and allows to compare their efficiency for a particular task. We find that on average linguistically valid modifications reveal unrobustness more often than corrupting the text (i.e. such tests as \textbf{Orthography}) (see ``Average'' column in \Cref{tab:all_tasks_unrob}). The exception here are the NER task where orthography is directly relevant to performance, and, surprisingly, IFEval, where the model often confused the instructions regarding capitalization with capitalization used in the instruction. Thus, the choice of the test type -- adversarial corruptions or natural modifications -- should also be motivated by the downstream task.

\subsection{The ability to generate a modification does not entail robustness against it}
\label{sec:discussion_llm_generated_data}

\begin{figure}[th]
    \centering
    \includegraphics[width=1\linewidth]{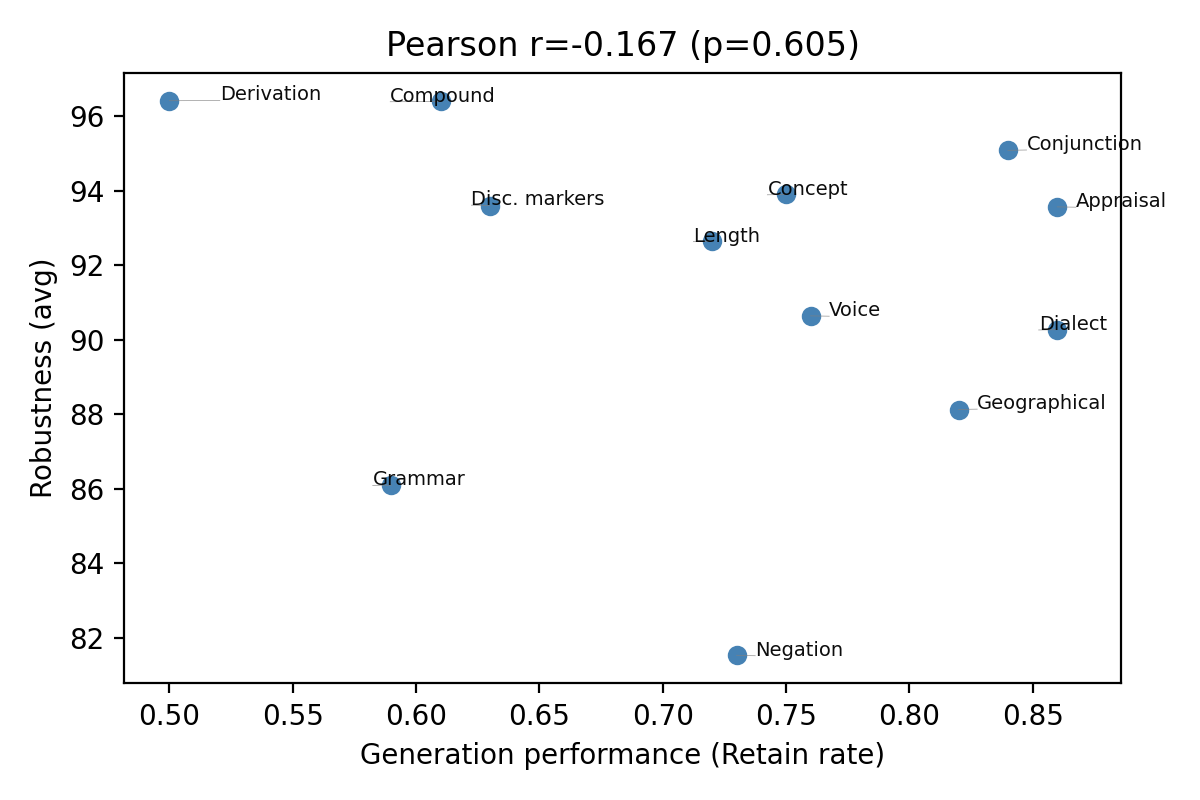}
    \caption{Generation Performance vs Robustness (100-U) of GPT-4o}
    \label{fig:correlation}
\end{figure}

Though it is a common practice to use LLMs to generate test data, this raises a question of circularity if the same model (in our case, GPT-4o) is used to modify instances and then tested against them.
To address this concern, we compare the quality of modifications generated by GPT-4o (as evaluated by annotators in terms of \textbf{retain rate}, i.e. percentage of modifications that were found correct and fluent) with the robustness of the same model to these modifications. As can be seen from \Cref{fig:correlation}, as well as low and insignificant Pearson correlation, there is no clear relation between these two aspects:  while some modifications (\textbf{Geographical bias}, \textbf{Dialect}) are easy for the model to generate, they lead to significant unrobustness; on the other hand, some modifications can be very hard (\textbf{Derivation}, \textbf{Compounds}), but the model is less brittle to them. Thus, the ability of the model to \textit{understand} a linguistic feature (i.e. apply it when generating) does not straightforwardly translate to its ability to handle this feature on downstream tasks, and the samples that were generated by a model so as to include a particular feature can be used to test robustness to that feature. 

%

\section{Conclusion}

In this paper, we present \texttt{FLUKE}, a task-agnostic, linguistically-driven framework that introduces minimal modifications to existing datasets to create a benchmark for assessing robustness. We automate these
modifications by prompting GPT-4o, followed by thorough manual validation to ensure data quality. To showcase the generalizability of this framework, we apply it to create modified instances across six tasks covering both classification and generation, and evaluate the performance of smaller fine-tuned models and large language models. Our findings reveal that: (1) the impact of modifications is highly task-dependent, with some tests being critical for certain tasks but irrelevant for others; (2) while LLMs demonstrate better overall robustness compared to PLMs, they still exhibit significant brittleness, and on some tasks reasoning LLMs are less robust than base LLMs; (3) increasing the model size helps only with more surface-level modifications and has diminishing effects; (4) models are overall less robust to linguistically valid  modifications than to adversarial corruption, but this again depends on the task; (5) the ability of a model to implement a linguistic feature during generation does not correlate with its robustness to that feature on downstream tasks.
We hope that \texttt{FLUKE} provides an alternative approach for evaluating models, and that model developers would consider integrating these results
into model cards  \citep{Mitchell2019} when releasing models. 

\section{Limitations}

Although \texttt{FLUKE} aims to cover a wide range linguistics capabilities, we acknowledge that it is by no means exhaustive: there are still many areas that it can be expanded, and the current suite of capability tests should serve as a foundation for model developers to create more task-specific tests. 

\texttt{FLUKE} is currently designed for English. Although many of its capability tests are applicable to other languages, our LLM prompts, human validation experiments and results are limited to English. To adopt \texttt{FLUKE} to another language, we recommend performing a thorough two-stage human evaluation on a sample of tasks to estimate the retain rate of modifications and reveal the tests that are likely to cause label change. For subsequent task, a simplified evalation with human-in-the-loop is possible. 

Although we generate the modifications with GPT-4o, the process is ultimately not fully automated, as we still require human assessment to validate the modifications. That said, since most of the automatic modifications were shown to be valid and we only need a small number of instances to understand model capability, when \texttt{FLUKE} is used to test a particular model for a particular task, we believe the validation can be done in-house by the model developers.

Previous works have pointed out potential biases in data generated by LLMs, stemming from the lack of its diversity \citep{ding-etal-2024-data}. In our work, such biases are most likely to occur in the geographical bias modification. To ensure a diverse representation, we compile a list of underrepresented regions as discussed in \Cref{sec:bias-test}, and explicitly specify a region from this list in the prompt to generate samples across the variety of underrepresented geographical locations.

\section*{Acknowledgements}

This research is supported by Oracle. Lau was supported by Australian Research Council

\bibliography{custom}

\begin{thebibliography}{59}
\providecommand{\natexlab}[1]{#1}

\bibitem[{Belinkov and Bisk(2017)}]{belinkov2017synthetic}
Yonatan Belinkov and Yonatan Bisk. 2017.
\newblock Synthetic and natural noise both break neural machine translation.
\newblock \emph{arXiv preprint arXiv:1711.02173}.

\bibitem[{Blevins et~al.(2023)Blevins, Gonen, and Zettlemoyer}]{Blevins2023}
Terra Blevins, Hila Gonen, and Luke Zettlemoyer. 2023.
\newblock \href {https://doi.org/10.18653/v1/2023.acl-long.367} {Prompting language models for linguistic structure}.
\newblock In \emph{Proceedings of the 61st Annual Meeting of the Association for Computational Linguistics (Volume 1: Long Papers)}, pages 6649--6663, Toronto, Canada. Association for Computational Linguistics.

\bibitem[{Cobbe et~al.(2021)Cobbe, Kosaraju, Bavarian, Chen, Jun, Kaiser, Plappert, Tworek, Hilton, Nakano et~al.}]{cobbe2021training}
Karl Cobbe, Vineet Kosaraju, Mohammad Bavarian, Mark Chen, Heewoo Jun, Lukasz Kaiser, Matthias Plappert, Jerry Tworek, Jacob Hilton, Reiichiro Nakano, and 1 others. 2021.
\newblock Training verifiers to solve math word problems.
\newblock \emph{arXiv preprint arXiv:2110.14168}.

\bibitem[{Cremers(2022)}]{cremers2022interpreting}
Alexandre Cremers. 2022.
\newblock Interpreting gradable adjectives: rational reasoning or simple heuristics?
\newblock \emph{Empirical Issues in Syntax and Semantics}, 14:31--61.

\bibitem[{Devlin et~al.(2019)Devlin, Chang, Lee, and Toutanova}]{Devlin+:2019}
Jacob Devlin, Ming-Wei Chang, Kenton Lee, and Kristina Toutanova. 2019.
\newblock {BERT}: Pre-training of deep bidirectional transformers for language understanding.
\newblock In \emph{Proceedings of the 2019 Conference of the North {A}merican Chapter of the Association for Computational Linguistics: Human Language Technologies, Volume 1 (Long and Short Papers)}, pages 4171--4186, Minneapolis, Minnesota.

\bibitem[{Ding et~al.(2024)Ding, Qin, Zhao, Luo, Li, Chen, Xia, Hu, Luu, and Joty}]{ding-etal-2024-data}
Bosheng Ding, Chengwei Qin, Ruochen Zhao, Tianze Luo, Xinze Li, Guizhen Chen, Wenhan Xia, Junjie Hu, Anh~Tuan Luu, and Shafiq Joty. 2024.
\newblock \href {https://doi.org/10.18653/v1/2024.findings-acl.97} {Data augmentation using {LLM}s: Data perspectives, learning paradigms and challenges}.
\newblock In \emph{Findings of the Association for Computational Linguistics: ACL 2024}, pages 1679--1705, Bangkok, Thailand. Association for Computational Linguistics.

\bibitem[{Ding et~al.(2021)Ding, Xu, Chen, Wang, Han, Xie, Zheng, and Liu}]{ding-etal-2021-nerd}
Ning Ding, Guangwei Xu, Yulin Chen, Xiaobin Wang, Xu~Han, Pengjun Xie, Haitao Zheng, and Zhiyuan Liu. 2021.
\newblock \href {https://doi.org/10.18653/v1/2021.acl-long.248} {Few-{NERD}: A few-shot named entity recognition dataset}.
\newblock In \emph{Proceedings of the 59th Annual Meeting of the Association for Computational Linguistics and the 11th International Joint Conference on Natural Language Processing (Volume 1: Long Papers)}, pages 3198--3213, Online. Association for Computational Linguistics.

\bibitem[{Ebrahimi et~al.(2018)Ebrahimi, Rao, Lowd, and Dou}]{Ebrahimi+:2018}
Javid Ebrahimi, Anyi Rao, Daniel Lowd, and Dejing Dou. 2018.
\newblock {H}ot{F}lip: White-box adversarial examples for text classification.
\newblock In \emph{Proceedings of the 56th Annual Meeting of the Association for Computational Linguistics (Volume 2: Short Papers)}, pages 31--36, Melbourne, Australia.

\bibitem[{Eger and Benz(2020)}]{eger-benz-2020-hero}
Steffen Eger and Yannik Benz. 2020.
\newblock From hero to z{\'e}roe: A benchmark of low-level adversarial attacks.
\newblock In \emph{Proceedings of the 1st Conference of the Asia-Pacific Chapter of the Association for Computational Linguistics and the 10th International Joint Conference on Natural Language Processing}, pages 786--803, Suzhou, China.

\bibitem[{Emami et~al.(2019)Emami, Trichelair, Trischler, Suleman, Schulz, and Cheung}]{emami-etal-2019-knowref}
Ali Emami, Paul Trichelair, Adam Trischler, Kaheer Suleman, Hannes Schulz, and Jackie Chi~Kit Cheung. 2019.
\newblock \href {https://doi.org/10.18653/v1/P19-1386} {The {K}now{R}ef coreference corpus: Removing gender and number cues for difficult pronominal anaphora resolution}.
\newblock In \emph{Proceedings of the 57th Annual Meeting of the Association for Computational Linguistics}, pages 3952--3961, Florence, Italy. Association for Computational Linguistics.

\bibitem[{Faisal and Anastasopoulos(2023)}]{faisal-anastasopoulos-2023-geographic}
Fahim Faisal and Antonios Anastasopoulos. 2023.
\newblock \href {https://doi.org/10.18653/v1/2023.mrl-1.12} {Geographic and geopolitical biases of language models}.
\newblock In \emph{Proceedings of the 3rd Workshop on Multi-lingual Representation Learning (MRL)}, pages 139--163, Singapore. Association for Computational Linguistics.

\bibitem[{Formento et~al.(2023)Formento, Foo, Tuan, and Ng}]{formento-etal-2023-using}
Brian Formento, Chuan~Sheng Foo, Luu~Anh Tuan, and See~Kiong Ng. 2023.
\newblock Using punctuation as an adversarial attack on deep learning-based {NLP} systems: An empirical study.
\newblock In \emph{Findings of the Association for Computational Linguistics: EACL 2023}, pages 1--34, Dubrovnik, Croatia.

\bibitem[{Godey et~al.(2024)Godey, de~la Clergerie, and Sagot}]{godey-etal-2024-scaling}
Nathan Godey, {\'E}ric de~la Clergerie, and Beno{\^\i}t Sagot. 2024.
\newblock \href {https://aclanthology.org/2024.lrec-main.1087} {On the scaling laws of geographical representation in language models}.
\newblock In \emph{Proceedings of the 2024 Joint International Conference on Computational Linguistics, Language Resources and Evaluation (LREC-COLING 2024)}, pages 12416--12422, Torino, Italia. ELRA and ICCL.

\bibitem[{Goyal et~al.(2022)Goyal, Gao, Chaudhary, Chen, Wenzek, Ju, Krishnan, Ranzato, Guzm{\'a}n, and Fan}]{Goyal2022}
Naman Goyal, Cynthia Gao, Vishrav Chaudhary, Peng-Jen Chen, Guillaume Wenzek, Da~Ju, Sanjana Krishnan, Marc{'}Aurelio Ranzato, Francisco Guzm{\'a}n, and Angela Fan. 2022.
\newblock \href {https://doi.org/10.1162/tacl_a_00474} {The {F}lores-101 evaluation benchmark for low-resource and multilingual machine translation}.
\newblock \emph{Transactions of the Association for Computational Linguistics}, 10:522--538.

\bibitem[{Hendrycks et~al.(2021)Hendrycks, Burns, Basart, Zou, Mazeika, Song, and Steinhardt}]{Hendrycks+:2021}
Dan Hendrycks, Collin Burns, Steven Basart, Andy Zou, Mantas Mazeika, Dawn Song, and Jacob Steinhardt. 2021.
\newblock Measuring massive multitask language understanding.
\newblock \emph{Proceedings of the International Conference on Learning Representations (ICLR)}.

\bibitem[{Hupkes et~al.(2023)Hupkes, Giulianelli, Dankers, Artetxe, Elazar, Pimentel, Christodoulopoulos, Lasri, Saphra, Sinclair et~al.}]{Hupkes+:2023}
Dieuwke Hupkes, Mario Giulianelli, Verna Dankers, Mikel Artetxe, Yanai Elazar, Tiago Pimentel, Christos Christodoulopoulos, Karim Lasri, Naomi Saphra, Arabella Sinclair, and 1 others. 2023.
\newblock A taxonomy and review of generalization research in {NLP}.
\newblock \emph{Nature Machine Intelligence}, 5(10):1161--1174.

\bibitem[{Jia and Liang(2017)}]{Jia+:2017}
Robin Jia and Percy Liang. 2017.
\newblock Adversarial examples for evaluating reading comprehension systems.
\newblock In \emph{Proceedings of the 2017 Conference on Empirical Methods in Natural Language Processing}, pages 2021--2031, Copenhagen, Denmark.

\bibitem[{Jin et~al.(2020)Jin, Jin, Zhou, and Szolovits}]{jin2020bert}
Di~Jin, Zhijing Jin, Joey~Tianyi Zhou, and Peter Szolovits. 2020.
\newblock Is {BERR} really robust? a strong baseline for natural language attack on text classification and entailment.
\newblock In \emph{Proceedings of the AAAI conference on artificial intelligence}, volume~34, pages 8018--8025.

\bibitem[{Kann et~al.(2019)Kann, Warstadt, Williams, and Bowman}]{Kann+:2019}
Katharina Kann, Alex Warstadt, Adina Williams, and Samuel~R. Bowman. 2019.
\newblock Verb argument structure alternations in word and sentence embeddings.
\newblock In \emph{Proceedings of the Society for Computation in Linguistics ({SC}i{L}) 2019}, pages 287--297.

\bibitem[{Keller et~al.(2021)Keller, Mackensen, and Eger}]{keller-etal-2021-bert}
Yannik Keller, Jan Mackensen, and Steffen Eger. 2021.
\newblock {BERT}-defense: A probabilistic model based on {BERT} to combat cognitively inspired orthographic adversarial attacks.
\newblock In \emph{Findings of the Association for Computational Linguistics: ACL-IJCNLP 2021}, pages 1616--1629, Online.

\bibitem[{Krippendorff()}]{krippendorff1computing}
Klaus Krippendorff.
\newblock Computing {K}rippendorff's {A}lpha-reliability.
\newblock \emph{Computing}, 1:25--2011.

\bibitem[{Le~Bras et~al.(2020)Le~Bras, Swayamdipta, Bhagavatula, Zellers, Peters, Sabharwal, and Choi}]{LeBras+:2020}
Ronan Le~Bras, Swabha Swayamdipta, Chandra Bhagavatula, Rowan Zellers, Matthew Peters, Ashish Sabharwal, and Yejin Choi. 2020.
\newblock Adversarial filters of dataset biases.
\newblock In \emph{International conference on machine learning}, pages 1078--1088.

\bibitem[{Levesque et~al.(2012)Levesque, Davis, and Morgenstern}]{wsc}
Hector~J. Levesque, Ernest Davis, and Leora Morgenstern. 2012.
\newblock The {Winograd} schema challenge.
\newblock In \emph{Proceedings of the Thirteenth International Conference on Principles of Knowledge Representation and Reasoning}, KR'12, page 552–561. AAAI Press.

\bibitem[{Li et~al.(2018)Li, Ji, Du, Li, and Wang}]{li2018textbugger}
Jinfeng Li, Shouling Ji, Tianyu Du, Bo~Li, and Ting Wang. 2018.
\newblock Textbugger: Generating adversarial text against real-world applications.
\newblock \emph{arXiv preprint arXiv:1812.05271}.

\bibitem[{Li et~al.(2024)Li, Cui, Zhao, Kong, and Bi}]{li-etal-2024-gsm}
Qintong Li, Leyang Cui, Xueliang Zhao, Lingpeng Kong, and Wei Bi. 2024.
\newblock \href {https://doi.org/10.18653/v1/2024.acl-long.163} {{GSM}-plus: A comprehensive benchmark for evaluating the robustness of {LLM}s as mathematical problem solvers}.
\newblock In \emph{Proceedings of the 62nd Annual Meeting of the Association for Computational Linguistics (Volume 1: Long Papers)}, pages 2961--2984, Bangkok, Thailand. Association for Computational Linguistics.

\bibitem[{Li et~al.(2023)Li, Liu, Gao, and Buntine}]{Li+:2023}
Xinzhe Li, Ming Liu, Shang Gao, and Wray Buntine. 2023.
\newblock A survey on out-of-distribution evaluation of neural {NLP} models.
\newblock In \emph{Proceedings of the Thirty-Second International Joint Conference on Artificial Intelligence}, pages 6683--6691.

\bibitem[{Linzen(2020)}]{Linzen+:2020}
Tal Linzen. 2020.
\newblock How can we accelerate progress towards human-like linguistic generalization?
\newblock In \emph{Proceedings of the 58th Annual Meeting of the Association for Computational Linguistics}, pages 5210--5217, Online.

\bibitem[{Linzen et~al.(2016)Linzen, Dupoux, and Goldberg}]{Linzen+:2016}
Tal Linzen, Emmanuel Dupoux, and Yoav Goldberg. 2016.
\newblock Assessing the ability of {LSTM}s to learn syntax-sensitive dependencies.
\newblock \emph{Transactions of the Association for Computational Linguistics}, 4:521--535.

\bibitem[{Mahowald et~al.(2024)Mahowald, Ivanova, Blank, Kanwisher, Tenenbaum, and Fedorenko}]{Mahowald2024}
Kyle Mahowald, Anna~A. Ivanova, Idan~A. Blank, Nancy Kanwisher, Joshua~B. Tenenbaum, and Evenlina Fedorenko. 2024.
\newblock Dissociating language and thought in large language models.
\newblock \emph{Trends in Cognitive Sciences}, 28:517--540.

\bibitem[{Marvin and Linzen(2018)}]{Linzen+:2018}
Rebecca Marvin and Tal Linzen. 2018.
\newblock Targeted syntactic evaluation of language models.
\newblock In \emph{Proceedings of the 2018 Conference on Empirical Methods in Natural Language Processing}, pages 1192--1202, Brussels, Belgium.

\bibitem[{McCoy et~al.(2019)McCoy, Pavlick, and Linzen}]{McCoy+:2019}
Tom McCoy, Ellie Pavlick, and Tal Linzen. 2019.
\newblock Right for the wrong reasons: Diagnosing syntactic heuristics in natural language inference.
\newblock In \emph{Proceedings of ACL}, pages 3428--3448, Florence, Italy.

\bibitem[{Mitchell et~al.(2019)Mitchell, Wu, Zaldivar, Barnes, Vasserman, Hutchinson, Spitzer, Raji, and Gebru}]{Mitchell2019}
Margaret Mitchell, Simone Wu, Andrew Zaldivar, Parker Barnes, Lucy Vasserman, Ben Hutchinson, Elena Spitzer, Inioluwa~Deborah Raji, and Timnit Gebru. 2019.
\newblock \href {https://doi.org/10.1145/3287560.3287596} {Model cards for model reporting}.
\newblock In \emph{Proceedings of the Conference on Fairness, Accountability, and Transparency}, FAT* '19, page 220–229, New York, NY, USA. Association for Computing Machinery.

\bibitem[{Nakayama(2018)}]{Nakayama+:2018}
Hiroki Nakayama. 2018.
\newblock \href {https://github.com/chakki-works/seqeval} {{seqeval}: A {Python} framework for sequence labeling evaluation}.
\newblock Software available from https://github.com/chakki-works/seqeval.

\bibitem[{Nie et~al.(2020)Nie, Williams, Dinan, Bansal, Weston, and Kiela}]{Nie+:2020}
Yixin Nie, Adina Williams, Emily Dinan, Mohit Bansal, Jason Weston, and Douwe Kiela. 2020.
\newblock Adversarial {NLI}: A new benchmark for natural language understanding.
\newblock In \emph{Proceedings of the 58th Annual Meeting of the Association for Computational Linguistics}. Association for Computational Linguistics.

\bibitem[{Nie et~al.(2021)Nie, Williamson, Bansal, Kiela, and Weston}]{nie-etal-2021-like}
Yixin Nie, Mary Williamson, Mohit Bansal, Douwe Kiela, and Jason Weston. 2021.
\newblock \href {https://doi.org/10.18653/v1/2021.acl-long.134} {{I} like fish, especially dolphins: Addressing contradictions in dialogue modeling}.
\newblock In \emph{Proceedings of the 59th Annual Meeting of the Association for Computational Linguistics and the 11th International Joint Conference on Natural Language Processing (Volume 1: Long Papers)}, pages 1699--1713, Online. Association for Computational Linguistics.

\bibitem[{Otmakhova et~al.(2022)Otmakhova, Verspoor, Baldwin, Yepes, and Lau}]{otmakhova2022m3}
Julia Otmakhova, Karin Verspoor, Timothy Baldwin, Antonio~Jimeno Yepes, and Jey~Han Lau. 2022.
\newblock {M3}: Multi-level dataset for multi-document summarisation of medical studies.
\newblock In \emph{Findings of the Association for Computational Linguistics: EMNLP 2022}, pages 3887--3901.

\bibitem[{Ousidhoum et~al.(2024)Ousidhoum, Muhammad, Abdalla, Abdulmumin, Ahmad, Ahuja, Aji, Araujo, Beloucif, De~Kock, Hourrane, Shrivastava, Solorio, Surange, Vishnubhotla, Yimam, and Mohammad}]{Ousidhoum2024}
Nedjma Ousidhoum, Shamsuddeen~Hassan Muhammad, Mohamed Abdalla, Idris Abdulmumin, Ibrahim~Said Ahmad, Sanchit Ahuja, Alham~Fikri Aji, Vladimir Araujo, Meriem Beloucif, Christine De~Kock, Oumaima Hourrane, Manish Shrivastava, Thamar Solorio, Nirmal Surange, Krishnapriya Vishnubhotla, Seid~Muhie Yimam, and Saif~M. Mohammad. 2024.
\newblock \href {https://doi.org/10.18653/v1/2024.semeval-1.272} {{S}em{E}val task 1: Semantic textual relatedness for {A}frican and {A}sian languages}.
\newblock In \emph{Proceedings of the 18th International Workshop on Semantic Evaluation (SemEval-2024)}, pages 1963--1978, Mexico City, Mexico. Association for Computational Linguistics.

\bibitem[{Pei et~al.(2023)Pei, Silva, Bos, Liu, Neves, Jurgens, and Barbieri}]{Pei2023}
Jiaxin Pei, V{\'\i}tor Silva, Maarten Bos, Yozen Liu, Leonardo Neves, David Jurgens, and Francesco Barbieri. 2023.
\newblock \href {https://doi.org/10.18653/v1/2023.semeval-1.309} {{S}em{E}val-2023 task 9: Multilingual tweet intimacy analysis}.
\newblock In \emph{Proceedings of the 17th International Workshop on Semantic Evaluation (SemEval-2023)}, pages 2235--2246, Toronto, Canada. Association for Computational Linguistics.

\bibitem[{Pullum and Huddleston(2002)}]{pullum_huddleston_huddleston_pullum_2002}
Geoffrey~K. Pullum and Rodney Huddleston. 2002.
\newblock \href {https://doi.org/10.1017/9781316423530.010} {\emph{Negation}}, chapter~9.
\newblock Cambridge University Press.

\bibitem[{Radford et~al.(2019)Radford, Wu, Child, Luan, Amodei, and Sutskever}]{Radford+:2019}
Alec Radford, Jeff Wu, Rewon Child, David Luan, Dario Amodei, and Ilya Sutskever. 2019.
\newblock Language models are unsupervised multitask learners.

\bibitem[{Raffel et~al.(2020)Raffel, Shazeer, Roberts, Lee, Narang, Matena, Zhou, Li, and Liu}]{Raffel+:2020}
Colin Raffel, Noam Shazeer, Adam Roberts, Katherine Lee, Sharan Narang, Michael Matena, Yanqi Zhou, Wei Li, and Peter~J. Liu. 2020.
\newblock Exploring the limits of transfer learning with a unified text-to-text transformer.
\newblock \emph{Journal of Machine Learning Research}, 21(140):1--67.

\bibitem[{Ravichander et~al.(2022)Ravichander, Gardner, and Marasovic}]{ravichander-etal-2022-condaqa}
Abhilasha Ravichander, Matt Gardner, and Ana Marasovic. 2022.
\newblock \href {https://doi.org/10.18653/v1/2022.emnlp-main.598} {{CONDAQA}: A contrastive reading comprehension dataset for reasoning about negation}.
\newblock In \emph{Proceedings of the 2022 Conference on Empirical Methods in Natural Language Processing}, pages 8729--8755, Abu Dhabi, United Arab Emirates. Association for Computational Linguistics.

\bibitem[{Ren et~al.(2019)Ren, Deng, He, and Che}]{ren2019generating}
Shuhuai Ren, Yihe Deng, Kun He, and Wanxiang Che. 2019.
\newblock Generating natural language adversarial examples through probability weighted word saliency.
\newblock In \emph{Proceedings of the 57th annual meeting of the association for computational linguistics}, pages 1085--1097.

\bibitem[{Ribeiro et~al.(2020)Ribeiro, Wu, Guestrin, and Singh}]{Ribeiro+:2020}
Marco~Tulio Ribeiro, Tongshuang Wu, Carlos Guestrin, and Sameer Singh. 2020.
\newblock Beyond accuracy: Behavioral testing of {NLP} models with {CheckList}.
\newblock In \emph{Proceedings of the 58th Annual Meeting of the Association for Computational Linguistics}, pages 4902--4912.

\bibitem[{Sennrich(2017)}]{Sennrich+:2017}
Rico Sennrich. 2017.
\newblock How grammatical is character-level neural machine translation? assessing {MT} quality with contrastive translation pairs.
\newblock In \emph{Proceedings of the 15th Conference of the European Chapter of the Association for Computational Linguistics: Volume 2, Short Papers}, pages 376--382.

\bibitem[{Singhal et~al.(2023)Singhal, Forristal, Ye, and Durrett}]{Singhal+:2023}
Prasann Singhal, Jarad Forristal, Xi~Ye, and Greg Durrett. 2023.
\newblock Assessing out-of-domain language model performance from few examples.
\newblock In \emph{Proceedings of the 17th Conference of the European Chapter of the Association for Computational Linguistics}, pages 2385--2397, Dubrovnik, Croatia.

\bibitem[{Socher et~al.(2013)Socher, Perelygin, Wu, Chuang, Manning, Ng, and Potts}]{socher-etal-2013-recursive}
Richard Socher, Alex Perelygin, Jean Wu, Jason Chuang, Christopher~D. Manning, Andrew Ng, and Christopher Potts. 2013.
\newblock \href {https://aclanthology.org/D13-1170} {Recursive deep models for semantic compositionality over a sentiment treebank}.
\newblock In \emph{Proceedings of the 2013 Conference on Empirical Methods in Natural Language Processing}, pages 1631--1642, Seattle, Washington, USA. Association for Computational Linguistics.

\bibitem[{Song et~al.(2023)Song, Khanuja, Liu, Faisal, Ostapenko, Winata, Aji, Cahyawijaya, Tsvetkov, Anastasopoulos, and Neubig}]{Song2023}
Yueqi Song, Simran Khanuja, Pengfei Liu, Fahim Faisal, Alissa Ostapenko, Genta Winata, Alham~Fikri Aji, Samuel Cahyawijaya, Yulia Tsvetkov, Antonios Anastasopoulos, and Graham Neubig. 2023.
\newblock \href {https://doi.org/10.18653/v1/2023.emnlp-main.875} {{G}lobal{B}ench: A benchmark for global progress in natural language processing}.
\newblock In \emph{Proceedings of the 2023 Conference on Empirical Methods in Natural Language Processing}, pages 14157--14171, Singapore. Association for Computational Linguistics.

\bibitem[{Srivastava et~al.(2022)Srivastava, Rastogi, Rao, Shoeb, Abid, Fisch, Brown, Santoro, Gupta, Garriga-Alonso et~al.}]{Srivastava+:2022}
Aarohi Srivastava, Abhinav Rastogi, Abhishek Rao, Abu Awal~Md Shoeb, Abubakar Abid, Adam Fisch, Adam~R Brown, Adam Santoro, Aditya Gupta, Adri{\`a} Garriga-Alonso, and 1 others. 2022.
\newblock Beyond the imitation game: Quantifying and extrapolating the capabilities of language models.
\newblock \emph{arXiv preprint arXiv:2206.04615}.

\bibitem[{Tan et~al.(2020)Tan, Joty, Kan, and Socher}]{tan-etal-2020-morphin}
Samson Tan, Shafiq Joty, Min-Yen Kan, and Richard Socher. 2020.
\newblock It`s morphin' time! {C}ombating linguistic discrimination with inflectional perturbations.
\newblock In \emph{Proceedings of the 58th Annual Meeting of the Association for Computational Linguistics}, pages 2920--2935, Online.

\bibitem[{Truong et~al.(2022)Truong, Otmakhova, Baldwin, Cohn, Lau, and Verspoor}]{truong2022not}
Thinh~Hung Truong, Julia Otmakhova, Timothy Baldwin, Trevor Cohn, Jey~Han Lau, and Karin Verspoor. 2022.
\newblock {Not another Negation Benchmark: The NaN-NLI} test suite for sub-clausal negation.
\newblock In \emph{Proceedings of the 2nd Conference of the Asia-Pacific Chapter of the Association for Computational Linguistics and the 12th International Joint Conference on Natural Language Processing (Volume 1: Long Papers)}, pages 883--894.

\bibitem[{Waldis et~al.(2024)Waldis, Perlitz, Choshen, Hou, and Gurevych}]{Waldis2024}
Andreas Waldis, Yotam Perlitz, Leshem Choshen, Yufang Hou, and Iryna Gurevych. 2024.
\newblock \href {https://arxiv.org/abs/2404.18923} {{Holmes}: A benchmark to assess the linguistic competence of language models}.
\newblock \emph{Preprint}, arXiv:2404.18923.

\bibitem[{Wang et~al.(2019)Wang, Pruksachatkun, Nangia, Singh, Michael, Hill, Levy, and Bowman}]{Wang+:2019}
Alex Wang, Yada Pruksachatkun, Nikita Nangia, Amanpreet Singh, Julian Michael, Felix Hill, Omer Levy, and Samuel Bowman. 2019.
\newblock Superglue: A stickier benchmark for general-purpose language understanding systems.
\newblock \emph{Advances in neural information processing systems}, 32.

\bibitem[{Wang et~al.(2018)Wang, Singh, Michael, Hill, Levy, and Bowman}]{Wang+:2018}
Alex Wang, Amanpreet Singh, Julian Michael, Felix Hill, Omer Levy, and Samuel Bowman. 2018.
\newblock {GLUE}: A multi-task benchmark and analysis platform for natural language understanding.
\newblock In \emph{Proceedings of the 2018 {EMNLP} Workshop {B}lackbox{NLP}: Analyzing and Interpreting Neural Networks for {NLP}}, pages 353--355, Brussels, Belgium.

\bibitem[{Warstadt et~al.(2019)Warstadt, Cao, Grosu, Peng, Blix, Nie, Alsop, Bordia, Liu, Parrish, Wang, Phang, Mohananey, Htut, Jeretic, and Bowman}]{Warstadt+2019}
Alex Warstadt, Yu~Cao, Ioana Grosu, Wei Peng, Hagen Blix, Yining Nie, Anna Alsop, Shikha Bordia, Haokun Liu, Alicia Parrish, Sheng-Fu Wang, Jason Phang, Anhad Mohananey, Phu~Mon Htut, Paloma Jeretic, and Samuel~R. Bowman. 2019.
\newblock Investigating {BERT}{'}s knowledge of language: Five analysis methods with {NPI}s.
\newblock In \emph{Proceedings of the 2019 Conference on Empirical Methods in Natural Language Processing and the 9th International Joint Conference on Natural Language Processing (EMNLP-IJCNLP)}, pages 2877--2887, Hong Kong, China.

\bibitem[{Warstadt et~al.(2020)Warstadt, Parrish, Liu, Mohananey, Peng, Wang, and Bowman}]{Warstadt+:2020}
Alex Warstadt, Alicia Parrish, Haokun Liu, Anhad Mohananey, Wei Peng, Sheng-Fu Wang, and Samuel~R Bowman. 2020.
\newblock {BLiMP}: The benchmark of linguistic minimal pairs for english.
\newblock \emph{Transactions of the Association for Computational Linguistics}, 8:377--392.

\bibitem[{Zellers et~al.(2019)Zellers, Holtzman, Bisk, Farhadi, and Choi}]{Zellers+:2019}
Rowan Zellers, Ari Holtzman, Yonatan Bisk, Ali Farhadi, and Yejin Choi. 2019.
\newblock {HellaSwag}: Can a machine really finish your sentence?
\newblock In \emph{Proceedings of the 57th Annual Meeting of the Association for Computational Linguistics}, pages 4791--4800.

\bibitem[{Zhong et~al.(2024)Zhong, Cui, Guo, Liang, Lu, Wang, Saied, Chen, and Duan}]{Zhong+:2024}
Wanjun Zhong, Ruixiang Cui, Yiduo Guo, Yaobo Liang, Shuai Lu, Yanlin Wang, Amin Saied, Weizhu Chen, and Nan Duan. 2024.
\newblock {AGIEval}: A human-centric benchmark for evaluating foundation models.
\newblock In \emph{Findings of the Association for Computational Linguistics: NAACL 2024}, pages 2299--2314.

\bibitem[{Zhou et~al.(2023)Zhou, Lu, Mishra, Brahma, Basu, Luan, Zhou, and Hou}]{zhou2023instruction}
Jeffrey Zhou, Tianjian Lu, Swaroop Mishra, Siddhartha Brahma, Sujoy Basu, Yi~Luan, Denny Zhou, and Le~Hou. 2023.
\newblock Instruction-following evaluation for large language models.
\newblock \emph{arXiv preprint arXiv:2311.07911}.

\end{thebibliography}

\appendix

\appendix
\section{Appendix}
\subsection{Data generation prompts}
\label{app:data_gen_prompt}
An example of generating compound word modification for coreference resolution task.

\begin{prompt}

Find any non-compound (single-root) word in the text below and change it into a compound word (word with several roots). Don't make change to the pronoun. \\

Example: "a sequence of ridiculous shooting scenes" -> "a sequence of ridiculous shoot-'em-up scenes" \\
Example: dull acting ->  lacklustre acting \\

Text: \textbf{Joe raced against Steven because he thought it would be easy.}\\
Pronoun: \textbf{he}\\

Modified text: \color{brown}{Joe raced against Steven because he thought it would be a no-brainer.}

\end{prompt}

\subsection{Annotation process and instructions}
\label{app:annotation}

We perform a two-staged validation of modified samples using a crowdsource annotation platform Prolific \footnote{\url{https://www.prolific.com/}}. To ensure high quality of annotations, we use a high per hour rate of 12 euro, which is above the minimal payment rate in the annotators' country of residence and well above the usual Prolific rate (overall cost: 7233 euro), choose only annotators from English-speaking countries (except for the \textbf{Dialect} test where we only used annotators from Singapore), and with a minimum success rate of 90\%, as well as perform quality checks as described below. The human validation experiments has been approved by the ethics
board (application ID: anonymised), and the annotators were informed of how their data will be used and had to express their consent.

\begin{figure*}[!tbp]
\centering
\includegraphics[width=\textwidth,height=\textheight,keepaspectratio]{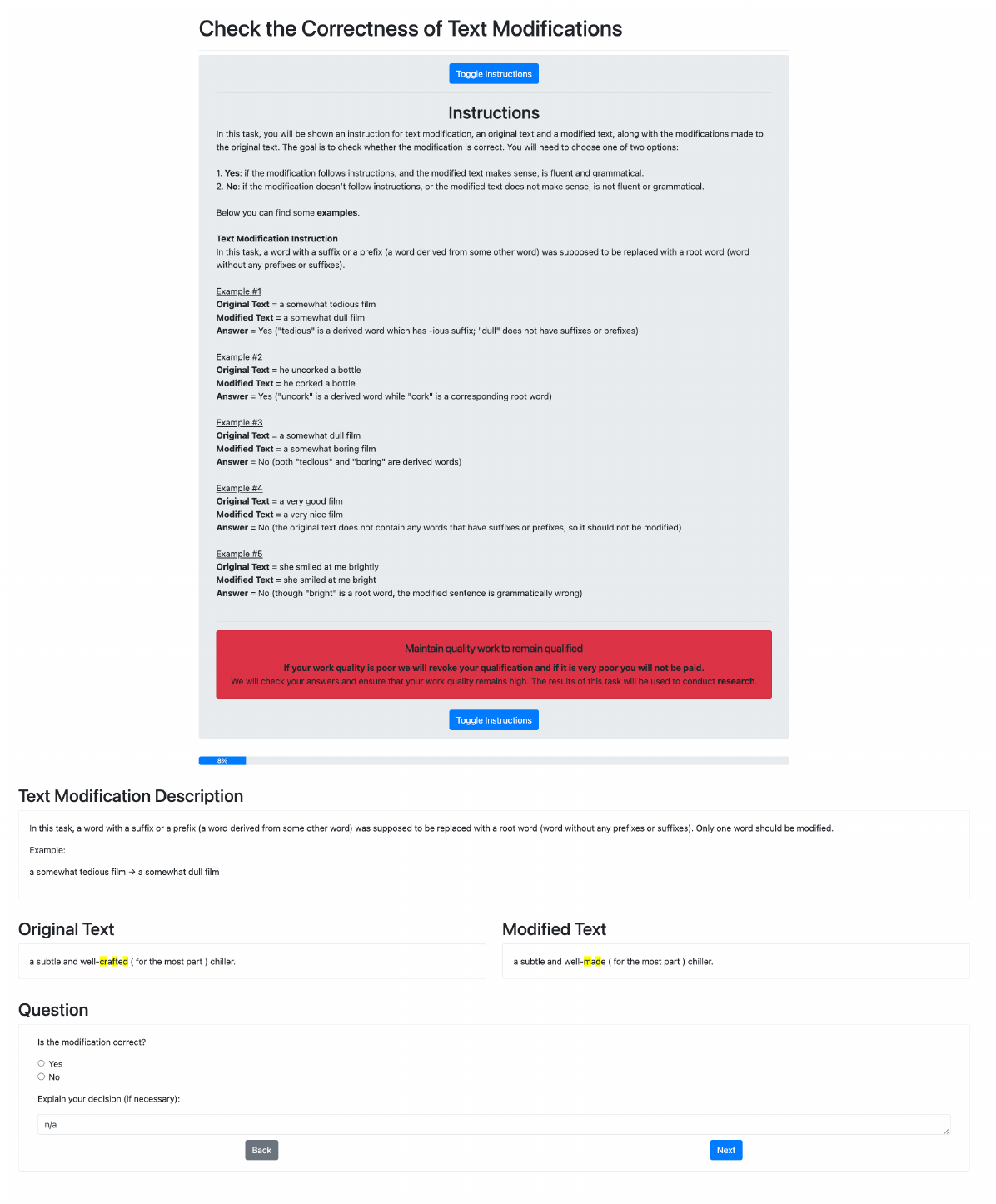}
\caption{The example annotation page for the use case \textbf{Sentiment Analysis (SA)} for the linguistic capability test \textbf{derivation} in Stage 1.}
\label{fig:annotation_interface_stage1}
\end{figure*}

During Stage 1, the annotators received instructions that described an intended modification (\textbf{Voice, Negation, Style} etc., see \Cref{sec:tests}) and were asked if the sample they see is modified correctly according to these instructions (see Figure \ref{fig:annotation_interface_stage1} for an example of annotation interface). 
For quality control, we include four control questions, which were randomly sampled from a different modification.
We remove annotators who fail to answer more than one of such questions (i.e. selecting ``Correct'' twice), or select one option too frequently (i.e. selecting ``Incorrect'' for all questions). Each of the samples is annotated by four annotators, and receives at least two annotations after the quality checks. 
We filter our samples which were not deemed to be ``Correct'' by the majority (over 50\%) of annotators.
In case of tie, we select the answer of the annotator with better performance on the control questions.  
The inter-annotator agreement in terms of Krippendorf's $\alpha$ \citep{krippendorff1computing}, averaged majority-class agreement (percentage of samples where the majority of annotators chose the same label), and the retain rate (percentage of samples that pass phase 1 annotation) are shown in Table \ref{tab:iaa_phase1}. We choose Krippendorf's $\alpha$ as an agreement metrics to accommodate for the difference in number of annotations per sample. We note, however, that it is not directly applicable here as it assumes that all annotations are provided by a fixed, ordered set of observers and factors in the consistency of their decisions, i.e. measures the annotators behavior agreement rather than agreement of labels. Thus, in our case the Majority rate agreement reveals more about the patterns of annotation agreement, and achieves substantially high levels. 

\begin{figure*}[!tbp]
\centering
\includegraphics[width=\textwidth,height=\textheight,keepaspectratio]{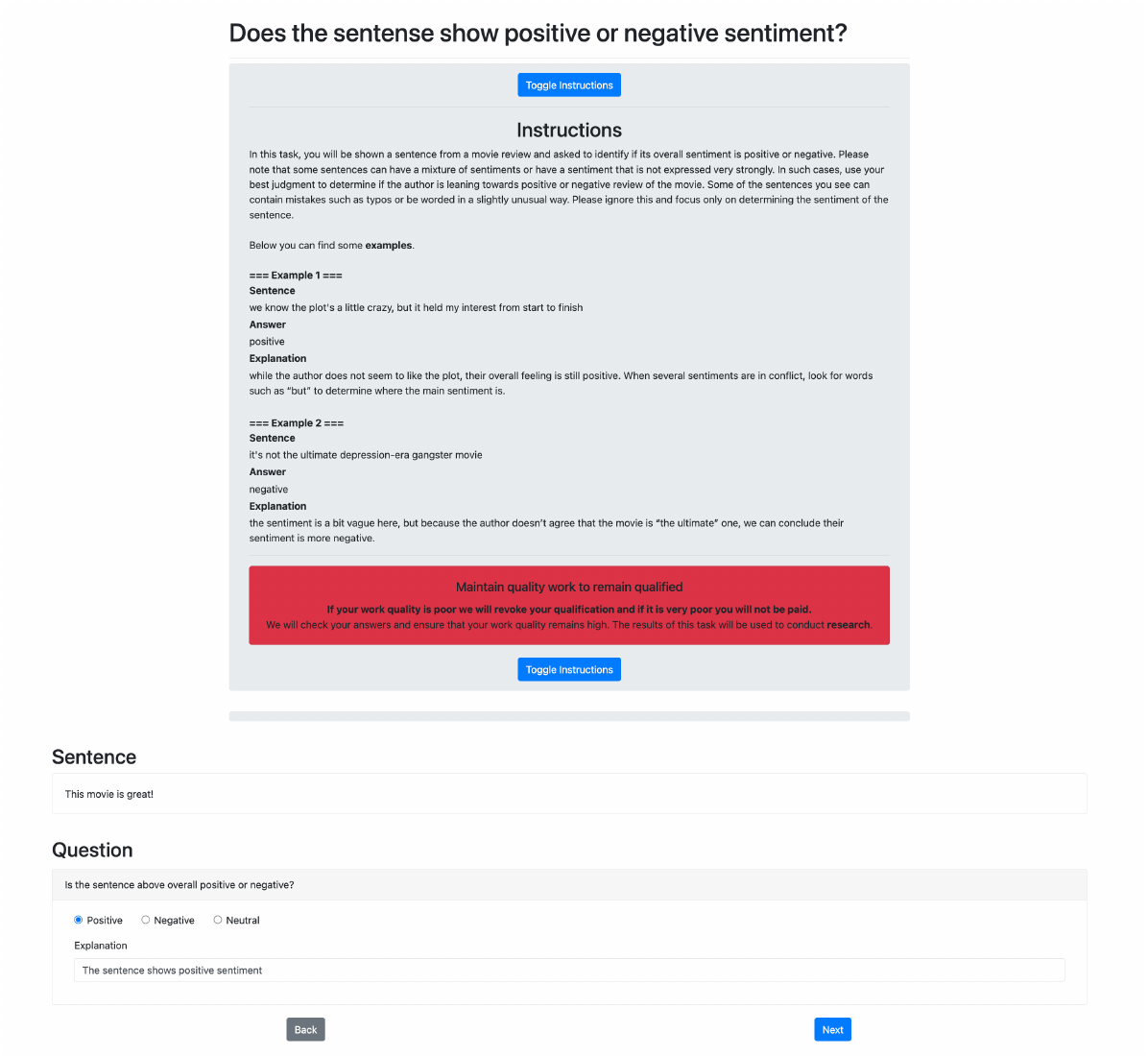}
\caption{The example annotation page for the use case \textbf{Sentiment Analysis (SA)} in Stage 2.}
\label{fig:annotation_interface_stage2}
\end{figure*}

During Stage 2, the annotators were given instructions for one of the four tasks (\textbf{Sentiment Analysis, Dialogue Contradiction Detection, Coreference Resolution, Named Entity Recognition}) and asked to assign a label specific for that task (for example, judge a modified sample as having \textit{positive} or \textit{negative} sentiment for the \textbf{Sentiment Analysis} task, see Figure \ref{fig:annotation_interface_stage2}). 
For quality checks, we used unmodified samples from the training data of the task (i.e. samples for which we knew the label); and apply the same filtering criteria in Stage 1. The final label was also selected based on the majority vote. We compared the resulting labels for modified samples with the labels for the original data to check if any inconsistencies were due to the genuine label change resulting from the modification, rather than to annotators disagreement. The inter-annotator agreement results (measured similarly to Stage 1) are presented in Table \ref{tab:iaa_phase2}.

\begin{table}[!tbp]
\footnotesize
\begin{center}
\begin{adjustbox}{max width=\linewidth}
\begin{tabular}{ll cccc}

\toprule
& Modification &\textbf{Krippendorff's $\alpha$} & \textbf{Majority rate} & \textbf{Retain rate} \\
\midrule
\textbf{Bias} &  \textbf{Temporal} & 0.33  & 0.85 & 0.80 \\
& \textbf{Geographical} & 0.35  & 0.84 & 0.82 \\
& \textbf{Length} & 0.30  & 0.83 & 0.72 \\
\midrule
\textbf{Morphology} & \textbf{Derivation} & 0.14 & 0.79 & 0.50 \\
& \textbf{Compound} & 0.39 & 0.86 & 0.61 \\
\midrule
\textbf{Syntax} & \textbf{Voice} & 0.30  & 0.83 & 0.76 \\
& \textbf{Grammar} & 0.17  & 0.80 & 0.59 \\
& \textbf{Conjunction} & 0.39  & 0.88 & 0.84 \\
\midrule
\textbf{Semantics} & \textbf{Concept} & 0.37  & 0.84 & 0.75 \\
& \textbf{Negation} & 0.42  & 0.90 & 0.73 \\
\textbf{Pragmatic}  & \textbf{Disc. markers} & 0.22 & 0.78 & 0.63 \\
& \textbf{Appraisal} & 0.46  & 0.89 & 0.86 \\
\midrule
\textbf{Varieties} &  \textbf{Style} & 0.53  & 0.91 & 0.93 \\
& \textbf{Dialect} & 0.41 & 0.87 & 0.86 \\

\bottomrule
\end{tabular}
\end{adjustbox}
\end{center}
\caption{Phase 1 annotation quality}
\label{tab:iaa_phase1}
\end{table}

\begin{table}[!tbp]
\footnotesize
\begin{center}
\begin{adjustbox}{max width=\linewidth}

    \begin{tabular}{r ccc}
    &\textbf{Krippendorff's $\alpha$} & \textbf{Majority rate} & \textbf{Label change rate} \\
    \hline
        \multicolumn{1}{l}{\textit{Sentiment Analysis}} & & &    \\
        \textbf{Semantics: Negation} & 0.54 & 0.84 & 0.53 \\
        \hline
        \multicolumn{1}{l}{\textit{Dialogue Contradiction Detection}} &  \\
        \textbf{Syntax: Grammar} & 0.34 & 0.81 & 0.33 \\
        \textbf{Semantics: Negation} & 0.39 & 0.80  & 0.41 \\
        \hline
        \multicolumn{1}{l}{\textit{Coreference Resolution}} &  \\
        \textbf{Syntax: Grammar} & 0.66 & 0.91 & 0.82  \\
        \textbf{Semantics: Negation} & 0.57 & 0.79 & 0.49\\
        \hline
        \multicolumn{1}{l}{\textit{NER}} &  \\
        \textbf{Bias: Geography} & 0.36 & 0.59 &  1.00 \\
        \textbf{Syntax: Grammar} & 0.43 & 0.67 & 0.94 \\
        \textbf{Syntax: Conjunction} & 0.36 & 0.63 & 1.00 \\
        
        \hline

    \end{tabular}
    \end{adjustbox}
\end{center}

    \caption{Phase 2 annotation quality}
    \label{tab:iaa_phase2}
\end{table}

\subsection{Examples of annotation decisions for generative tasks (simplified quality assurance scenario)}

We provide examples of decisions where we kept the modification together with the original label for the task, changed the label, or rejected the modification, in \Cref{tab:examples}.

\begin{table*}
\footnotesize
    \centering
    \begin{tabular}{l|c|p{3.5cm}|p{3.5cm}|c|p{3.5cm}}
    \toprule
        Task & Test & Original & Modified & Keep? & Reason \\
        \midrule
        GSM & Dialect & Reggie, Lynn, and Paisley ran together. Paisley ran 4 miles. Reggie ran 5 times what Paisley ran and 3 miles farther than Lynn. How many miles did Lynn run? & Reggie, Lynn, and Paisley \textit{go running kaki}. Paisley ran 4 miles. Reggie ran 5 times what Paisley ran and 3 miles \textit{more} than Lynn. \textit{Lynn run how many miles}? &  \cmark & Despite substantial language change, nothing has changed in the premises of the mathematical problem, so we keep the sample. \\
              GSM & Negation & Miss Albert's class is composed of 12 boys and 12 girls. One-third of the girls and one-fourth of the boys are on varsity. How many students are not on varsity? & Miss Albert's class is composed of 12 boys and 12 girls. One-third of the girls and one-fourth of the boys are \textit{not} on varsity. How many students are not on varsity?  & $\rightarrow$ & The modification is valid (the problem can be solved), but leads to a different answer (label).\\
        GSM & Concept & 3 trees each had 7 blue birds in them. 2 different trees each had 4 blue birds. 1 final tree had 3 blue birds. How many blue birds were in the trees in total? & 3 trees each had 7 blue birds in them. 2 different trees each had 4 blue birds. 1 final tree had 3 \textit{sprat} birds. How many blue birds were in the trees in total? &  \xmark & ``Blue'' was replaced by a nonce (non-existing) word ``sprat''. It is impossible to determine if ``sprat birds'' are blue or not, so the problem cannot be solved.\\
   
         IFEval & Temp. bias & Write exactly 4 paragraphs about tips for installing a car seat for moms. Use 2 new lines to separate paragraphs. Start the 4th paragraph with the word elm. & Write exactly 4 paragraphs about tips for installing a car seat for moms. Use 2 new lines to separate paragraphs. \textit{Commence} the 4th paragraph with the word 'elm'. & \cmark & The word ``begin'' was replaced by a slightly outdated word ``commence'', which is still well understandable in the context. The modification is valid and does not change the label. \\
         IFEval & Negation & Write me a resume for Matthias Algiers. Use words with all capital letters to highlight key abilities, but make sure that words with all capital letters appear less than 10 times. & Write me a resume for Matthias Algiers. Use words with all capital letters to highlight key abilities, but make sure that words with all capital letters appear \textit{more} than 10 times. & $\rightarrow$ & ``Less'' was changed to ``more", so if the model generates less than 10 words in all caps, it should be considered wrong. \\
         IFEval & Conjunct. & ...Your entire response should be in English, and should not contain any capital letters.  &  ...Your entire response should be in English, and should not contain any capital letters \textit{or punctuation}. & \xmark & The modification is fluent and applied correctly, but it makes the task more difficult (it is hard to avoid any punctuation), so we reject it.\\
    \end{tabular}
    \caption{Decision process for keeping (\cmark), rejecting (\xmark) modification, or changing the label ($\rightarrow$)}
    \label{tab:examples}
\end{table*}

\subsection{LLM settings}
We report the exact endpoint and sampling parameters for LLMs in \Cref{tab:llm_setting}.

\begin{table*}[!htbp]
\footnotesize
\begin{center}
\begin{adjustbox}{max width=\linewidth}
    \begin{tabular}{c  c c c c c }
    \toprule
     \textbf{Model}    & \textbf{GPT4o} & \textbf{Claude-3.5} & \textbf{Llama-3.1}  & \textbf{GPT-5} & \textbf{Deepseek R1}\\
     \midrule
     Endpoint & gpt-4o-2024-05-13 & claude-3.5-sonnet & llama-3.1-405b-instruct & gpt-5 & deepseek-r1\\
     \# Params & Undisclosed & Undisclosed & 405B & Undisclosed & 671B\\
     Temperature & 0 & 0 & 0 & 1 (default by reasoning model) & 1 (default by reasoning model)\\
     Max token & 1024 & 1024 & 1024  & 4096  & 4096 \\
     \bottomrule
     \end{tabular}
\end{adjustbox}
\end{center}
\caption{LLMs settings}
\label{tab:llm_setting}
\end{table*}

\label{app:llm_prompt}
\subsection*{Sentiment analysis prompt}

\begin{prompt}
Classify the sentiment of the given text. Answer with 1 for positive, 0 for negative. \\
Text: \textbf{it's a charming and often affecting journey.} \\
Answer: \color{brown} 1
\end{prompt}

\subsection*{Dialogue contradiction prompt}

\begin{prompt}
Does the last utterance contradict the dialogue context? Answer with 1 if contradict, 0 if not contradict. \\
Dialogue context: \\
\textbf{...} \\
\textbf{agent 0: well, i'm a big aerosmith fan, but i also like country.} \\
\textbf{agent 1: aerosmith is not bad but i love country so much} \\
\textbf{...} \\
Last utterance: \textbf{agent 1: i am a nascar fan too!} \\
Answer: \color{brown} 1

\end{prompt}

\subsection*{Coreference resolution prompt}

\begin{prompt}
Which candidate does the pronoun refer to? Answer with either 0 or 1.\\
Text: \textbf{The sniper shot the terrorist because he was a bad guy.} \\
Pronoun: \textbf{he} \\
Candidates: \textbf{0: The sniper, 1: the terrorist} \\
Answer: \color{brown}{1}

\end{prompt}

\subsection*{Named entity recognition prompt}
\begin{prompt}
Extract named entities from the text. Possible entity types: ART, BUILDING, EVENT, LOCATION, ORGANIZATION, OTHER, PERSON, PRODUCT. Reply with the list of entities in the format [{"text": entity\_span , "label": entity\_label},]. \\
Text: \textbf{Ronald will travel to Iceland.}\\
Answer: \color{brown}{[\{"text": "Ronald", "value": "PERSON"\}, \{"text": "Iceland", "value": "LOCATION"\}]}

\end{prompt}

\subsection{PLM settings}
\label{app:plms}

We report the finetuning details for PLMs in \Cref{tab:plm_setting}.

\textbf{SA:} We add a classification head on top of PLMs and fine-tune  as a sequence classification task. Input: text, Output: label (0|1)\\
 \textbf{Dialogue:} We add a classification head on top of PLMs and fine-tune  as a sequence classification task. Input: dialog\_context[SEP]last\_utterance, Output: label (0|1)\\
\textbf{Coref:} We add a classification head on top of PLMs and fine-tune  as a sequence classification task. Input: text[SEP]pronoun[SEP]candidates. Ouput: label (0|1)\\
\textbf{NER:} We add a classification head on top of PLMs and fine-tune as a sequence labeling task. Input: Text, Output: Label (CoNLL format)\\

Fine-tuning details are in \Cref{tab:plm_setting}.

\begin{table}[!htbp]
\footnotesize
\begin{center}
\begin{adjustbox}{max width=\linewidth}
    
    \begin{tabular}{r  c c c}
    \toprule
     \textbf{Model}    & \textbf{BERT} & \textbf{GPT-2} & \textbf{T5}  \\
     \midrule
     Checkpoint & bert-base-cased  & gpt2 & t5-base  \\
     Param & 110M & 124M & 220M    \\
     Optimizer & AdamW & AdamW & AdamW \\
     Loss & CrossEntropyLoss & CrossEntropyLoss & CrossEntropyLoss \\
    \midrule
        \multicolumn{1}{l}{\textit{SA}} & & &  \\ 
    Batch size & 16 & 32  & 32 \\
    Epoch  & 3 & 5 & 10\\
    Learning rate & 2e-5 & 2e-5 & 5e-5 \\ 
    Max length  & 128 & 128  & 128 \\
\midrule
    \multicolumn{1}{l}{\textit{Dialog}} & & &  \\ 
    Batch size & 8 & 8  & 8 \\
    Training step  & 10000 & 10000 & 10000 \\
    Learning rate & 2e-5 & 2e-5 & 2e-5 \\ 
    Max length  & 512 & 512  & 512 \\
    \midrule
    \multicolumn{1}{l}{\textit{Coref}} & & &  \\ 
    Batch size & 16 & 16  & 16 \\
    Epoch  & 3 & 3 & 3\\
    Learning rate & 2e-5 & 2e-5 & 2e-5 \\ 
    Max length  & 128 & 128  & 128 \\
    \midrule
    \multicolumn{1}{l}{\textit{NER}} & & &  \\ 
    Batch size & 32 & 32  & 32 \\
    Epoch  & 5 & 5 & 5\\
    Learning rate & 2e-5 & 2e-5 & 2e-5 \\ 
    Max length  & 128 & 128  & 128 \\
    \midrule

    \end{tabular}
\end{adjustbox}
\end{center}

    \caption{PLM fine-tuning details}
    \label{tab:plm_setting}
\end{table}

\subsection{Explicitly making model aware of the modification}
\label{app:context}

We analyze whether making the model aware of the intended modification would help them become more robust. For this experiment, we select GPT-5 model and add a single-line instruction at the beginning of the prompt (e.g. ``The following text has been transformed from active to passive voice''). Then, we observe the performance change between GPT-5 vs. GPT-5 (w. context): $\Delta$U = U(ctx) - U(std). Overall, there are no consistent gains across all tasks. We even observe degradation in IFEval, as the extra context add noise to instruction following capabilities (e.g. model confuse the extra context of ``modified to contain negation'' with the actual constraint that they must follow).

\begin{table}[ht]
\centering
\footnotesize
\begin{adjustbox}{max width=\linewidth}
\begin{tabular}{llrrrrrrr}
\toprule
\textbf{Category} & \textbf{Modification} & \textbf{SA} & \textbf{COREF} & \textbf{DIALOGUE} & \textbf{NER} & \textbf{GSM} & \textbf{IFEVAL} & \textbf{AVG} \\
\midrule
\textbf{Bias} & \textbf{Geographical} & \cellcolor{red!2} 1.0 & \cellcolor{red!8} 4.0 & \cellcolor{green!2} -1.1 & \cellcolor{red!1} 0.3 & \cellcolor{red!4} 2.0 & 0.0 & \cellcolor{red!2} 1.0 \\
 & \textbf{Length} & \cellcolor{red!2} 1.0 & 0.0 & \cellcolor{green!2} -1.0 & \cellcolor{green!4} -1.9 & \cellcolor{red!6} 3.0 & \cellcolor{red!4} 2.0 & \cellcolor{red!1} 0.5 \\
 & \textbf{Temporal} & \cellcolor{green!2} -1.0 & \cellcolor{green!4} -2.0 & 0.0 & \cellcolor{green!5} -2.3 & \cellcolor{red!2} 1.0 & \cellcolor{red!4} 2.0 & \cellcolor{green!1} -0.4 \\
\textbf{Orthography} & \textbf{Capitalization} & 0.0 & \cellcolor{green!2} -1.0 & \cellcolor{green!8} -4.2 & \cellcolor{green!7} -3.7 & 0.0 & \cellcolor{green!4} -2.0 & \cellcolor{green!4} -1.8 \\
 & \textbf{Punctuation} & 0.0 & \cellcolor{green!2} -1.0 & \cellcolor{green!4} -2.0 & \cellcolor{green!0} -0.1 & \cellcolor{red!4} 2.0 & \cellcolor{red!4} 2.0 & \cellcolor{red!0} 0.1 \\
 & \textbf{Spelling} & \cellcolor{red!2} 1.0 & \cellcolor{green!8} -4.1 & \cellcolor{red!6} 3.0 & \cellcolor{green!5} -2.5 & \cellcolor{red!2} 1.0 & \cellcolor{red!4} 2.0 & \cellcolor{red!0} 0.1 \\
\textbf{Morphology} & \textbf{Compound} & 0.0 & 0.0 & \cellcolor{green!4} -2.0 & \cellcolor{green!5} -2.7 & NA & NA & \cellcolor{green!2} -1.2 \\
 & \textbf{Derivation} & \cellcolor{red!2} 1.1 & \cellcolor{green!4} -2.0 & \cellcolor{red!6} 3.2 & \cellcolor{green!10} -5.2 & NA & NA & \cellcolor{green!1} -0.7 \\
\textbf{Syntax} & \textbf{Conjunction} & 0.0 & \cellcolor{green!2} -1.0 & \cellcolor{red!2} 1.0 & \cellcolor{green!8} -3.9 & 0.0 & \cellcolor{red!4} 2.0 & \cellcolor{green!1} -0.3 \\
 & \textbf{Grammar} & 0.0 & \cellcolor{green!6} -2.8 & \cellcolor{red!3} 1.5 & \cellcolor{green!7} -3.7 & NA & NA & \cellcolor{green!2} -1.2 \\
 & \textbf{Voice} & \cellcolor{green!2} -1.0 & 0.0 & 0.0 & \cellcolor{red!1} 0.6 & 0.0 & \cellcolor{red!12} 6.0 & \cellcolor{red!2} 0.9 \\
\textbf{Semantics} & \textbf{Concept} & 0.0 & \cellcolor{red!6} 3.0 & \cellcolor{green!10} -5.0 & \cellcolor{green!7} -3.4 & \cellcolor{green!2} -1.0 & \cellcolor{red!12} 6.0 & \cellcolor{green!0} -0.1 \\
 & \textbf{Negation} & 0.0 & \cellcolor{green!4} -2.0 & \cellcolor{red!6} 3.0 & \cellcolor{green!6} -3.0 & \cellcolor{red!8} 4.0 & \cellcolor{green!10} -5.0 & \cellcolor{green!1} -0.5 \\
\textbf{Discourse} & \textbf{Appraisal} & 0.0 & \cellcolor{green!6} -3.0 & 0.0 & \cellcolor{green!5} -2.7 & 0.0 & \cellcolor{red!2} 1.0 & \cellcolor{green!2} -0.8 \\
 & \textbf{Disc. markers} & \cellcolor{red!4} 2.0 & \cellcolor{red!6} 3.0 & \cellcolor{green!7} -3.4 & \cellcolor{red!2} 1.0 & NA & NA & \cellcolor{red!1} 0.6 \\
\textbf{Varieties} & \textbf{Dialect} & \cellcolor{green!2} -1.0 & \cellcolor{red!4} 2.0 & \cellcolor{green!4} -1.9 & \cellcolor{red!0} 0.2 & \cellcolor{red!2} 1.0 & \cellcolor{green!4} -2.0 & \cellcolor{green!1} -0.3 \\
 & \textbf{Style} & 0.0 & \cellcolor{green!6} -3.0 & \cellcolor{red!2} 1.0 & \cellcolor{red!0} 0.1 & 0.0 & \cellcolor{red!12} 6.0 & \cellcolor{red!1} 0.7 \\
 \midrule
\textbf{Average} & \textbf{Average} & \cellcolor{red!0} 0.2 & \cellcolor{green!1} -0.6 & \cellcolor{green!1} -0.5 & \cellcolor{green!4} -1.9 & \cellcolor{red!2} 1.0 & \cellcolor{red!3} 1.5 & \cellcolor{green!0} -0.2 \\
\bottomrule
\end{tabular}
\end{adjustbox}
\caption{GPT-5 vs GPT-5 (w. context): $\Delta$U = U(ctx) - U(std) (flip \%) by task and modification. Positive value (red) indicates a degradation in robustness while negative value (green) indicates an increase.}
\label{tab:gpt5_ctx_unrob}
\end{table}

\subsection{Use Case: NER and Coreference tasks (extended results)}

We present the full results of NER and Coreference tasks with confidence interval in \Cref{tab:ner_unrob_ci} and \Cref{tab:coref_unrob_ci}.

\begin{table*}[ht]
\centering
\resizebox{\linewidth}{!}{
\begin{tabular}{llrrrrrrrrr}
\toprule
Category & Modification & \multicolumn{3}{c}{\textbf{PLM}} & \multicolumn{5}{c}{\textbf{LLM}} \\
 &  & \textbf{BERT} & \textbf{GPT-2} & \textbf{T5} & \textbf{GPT-4o} & \textbf{Claude-3.5} & \textbf{Llama 3.1} & \textbf{GPT-5} & \textbf{DS R1} & \textbf{Avg} \\
\midrule
\textbf{Bias} & \textbf{Temporal} & \cellcolor{blue!10} 3.7 [1.4, 6.9] & \cellcolor{blue!8} 3.0 [0.4, 6.4] & \cellcolor{blue!3} 1.2 [0.3, 2.6] & \cellcolor{blue!16} 6.2 [2.5, 10.8] & \cellcolor{blue!11} 4.3 [1.1, 8.2] & \cellcolor{blue!5} 1.8 [0.3, 4.0] & \cellcolor{blue!20} 7.7 [4.4, 11.5] & \cellcolor{blue!27} 10.6 [6.4, 15.4] & \cellcolor{blue!12} 4.8 [2.1, 8.2] \\
  & \textbf{Geographical} & \cellcolor{blue!65} \textcolor{white}{25.1 [19.6, 30.9]} & \cellcolor{blue!71} \textcolor{white}{27.6 [21.7, 33.9]} & \cellcolor{blue!75} \textcolor{white}{29.0 [23.6, 34.5]} & \cellcolor{blue!58} \textcolor{white}{22.3 [17.5, 27.5]} & \cellcolor{blue!67} \textcolor{white}{26.0 [19.5, 32.8]} & \cellcolor{blue!70} \textcolor{white}{27.0 [21.5, 32.6]} & \cellcolor{blue!57} \textcolor{white}{22.0 [16.7, 27.7]} & \cellcolor{blue!72} \textcolor{white}{27.8 [22.7, 33.0]} & \cellcolor{blue!67} \textcolor{white}{25.8 [20.4, 31.6]} \\
  & \textbf{Length} & \cellcolor{blue!30} 11.8 [6.6, 17.7] & \cellcolor{blue!32} 12.2 [7.4, 17.7] & \cellcolor{blue!34} 13.1 [8.5, 18.3] & \cellcolor{blue!33} 12.7 [7.6, 18.5] & \cellcolor{blue!41} 15.7 [9.3, 23.1] & \cellcolor{blue!17} 6.4 [3.0, 10.5] & \cellcolor{blue!25} 9.7 [5.2, 15.1] & \cellcolor{blue!33} 12.6 [7.8, 18.0] & \cellcolor{blue!30} 11.8 [6.9, 17.4] \\
\midrule
\textbf{Orthography} & \textbf{Spelling} & \cellcolor{blue!11} 4.3 [1.6, 7.7] & \cellcolor{blue!6} 2.5 [0.4, 5.2] & \cellcolor{blue!4} 1.4 [0.5, 2.5] & \cellcolor{blue!17} 6.6 [3.8, 9.8] & \cellcolor{blue!8} 3.0 [0.5, 6.2] & \cellcolor{blue!10} 3.9 [1.6, 6.6] & \cellcolor{blue!19} 7.3 [4.3, 10.8] & \cellcolor{blue!29} 11.4 [7.3, 15.9] & \cellcolor{blue!13} 5.0 [2.5, 8.1] \\
  & \textbf{Capitalization} & \cellcolor{blue!2} 0.9 [0.1, 1.8] & \cellcolor{blue!50} \textcolor{white}{19.3 [13.1, 26.1]} & \cellcolor{blue!34} 13.1 [8.5, 18.2] & \cellcolor{blue!29} 11.3 [6.6, 16.6] & \cellcolor{blue!23} 8.9 [4.5, 14.1] & \cellcolor{blue!40} 15.3 [9.7, 21.6] & \cellcolor{blue!24} 9.1 [5.6, 13.1] & \cellcolor{blue!35} 13.6 [9.1, 18.5] & \cellcolor{blue!30} 11.5 [7.2, 16.3] \\
  & \textbf{Punctuation} & \cellcolor{blue!17} 6.5 [3.0, 10.7] & \cellcolor{blue!10} 3.7 [1.3, 6.8] & \cellcolor{blue!13} 4.9 [2.4, 7.8] & \cellcolor{blue!18} 7.1 [3.6, 11.2] & \cellcolor{blue!24} 9.1 [5.0, 13.7] & \cellcolor{blue!20} 7.9 [4.1, 12.4] & \cellcolor{blue!20} 7.8 [4.5, 11.7] & \cellcolor{blue!39} 15.0 [10.3, 20.2] & \cellcolor{blue!20} 7.8 [4.3, 11.8] \\
\midrule
\textbf{Morphology} & \textbf{Derivation} & \cellcolor{blue!5} 1.9 [0.5, 3.7] & \cellcolor{blue!11} 4.2 [1.4, 7.5] & \cellcolor{blue!15} 5.8 [2.1, 10.5] & \cellcolor{blue!9} 3.7 [1.1, 7.2] & \cellcolor{blue!5} 2.0 [0.1, 4.5] & \cellcolor{blue!3} 1.3 [0.1, 3.2] & \cellcolor{blue!25} 9.5 [4.4, 15.7] & \cellcolor{blue!21} 8.2 [4.2, 12.9] & \cellcolor{blue!12} 4.6 [1.7, 8.1] \\
  & \textbf{Compound} & \cellcolor{blue!8} 3.1 [0.8, 6.1] & \cellcolor{blue!2} 0.6 [0.0, 1.7] & \cellcolor{blue!13} 5.2 [2.0, 9.1] & \cellcolor{blue!8} 3.1 [0.8, 6.5] & \cellcolor{blue!4} 1.7 [0.0, 4.4] & \cellcolor{blue!3} 1.0 [0.0, 2.7] & \cellcolor{blue!18} 7.0 [3.4, 11.2] & \cellcolor{blue!33} 12.9 [8.2, 18.1] & \cellcolor{blue!11} 4.3 [1.9, 7.5] \\
\midrule
\textbf{Syntax} & \textbf{Voice} & \cellcolor{blue!20} 7.8 [3.9, 12.4] & \cellcolor{blue!28} 10.8 [6.0, 16.2] & \cellcolor{blue!15} 5.7 [3.1, 8.7] & \cellcolor{blue!19} 7.5 [3.6, 12.2] & \cellcolor{blue!14} 5.5 [2.1, 9.7] & \cellcolor{blue!11} 4.5 [1.5, 8.2] & \cellcolor{blue!21} 8.3 [4.3, 13.1] & \cellcolor{blue!29} 11.3 [7.2, 16.0] & \cellcolor{blue!20} 7.7 [4.0, 12.1] \\
  & \textbf{Grammar} & \cellcolor{blue!21} 8.0 [4.0, 12.9] & \cellcolor{blue!41} 15.9 [10.4, 21.8] & \cellcolor{blue!27} 10.5 [6.3, 15.5] & \cellcolor{blue!21} 8.3 [4.7, 12.5] & \cellcolor{blue!9} 3.5 [1.2, 6.6] & \cellcolor{blue!14} 5.3 [1.8, 9.6] & \cellcolor{blue!28} 10.8 [6.1, 16.1] & \cellcolor{blue!32} 12.4 [8.1, 17.0] & \cellcolor{blue!24} 9.3 [5.3, 14.0] \\
  & \textbf{Conjunction} & \cellcolor{blue!24} 9.1 [4.1, 15.4] & \cellcolor{blue!20} 7.6 [4.3, 11.3] & \cellcolor{blue!20} 7.7 [4.4, 11.4] & \cellcolor{blue!24} 9.4 [5.6, 14.0] & \cellcolor{blue!19} 7.5 [3.3, 12.9] & \cellcolor{blue!21} 8.0 [3.4, 13.7] & \cellcolor{blue!30} 11.7 [7.1, 17.1] & \cellcolor{blue!34} 13.2 [8.4, 18.7] & \cellcolor{blue!24} 9.3 [5.1, 14.3] \\
\midrule
\textbf{Semantics} & \textbf{Concept} & \cellcolor{blue!13} 5.0 [2.3, 8.4] & \cellcolor{blue!23} 8.9 [4.3, 14.2] & \cellcolor{blue!14} 5.3 [2.6, 8.7] & \cellcolor{blue!17} 6.5 [3.2, 10.5] & \cellcolor{blue!12} 4.8 [1.2, 9.3] & \cellcolor{blue!10} 3.9 [1.2, 7.5] & \cellcolor{blue!22} 8.5 [4.1, 13.7] & \cellcolor{blue!24} 9.4 [5.7, 13.7] & \cellcolor{blue!17} 6.5 [3.1, 10.7] \\
  & \textbf{Negation} & \cellcolor{blue!12} 4.8 [2.2, 8.0] & \cellcolor{blue!14} 5.2 [2.9, 7.9] & \cellcolor{blue!17} 6.5 [3.7, 9.9] & \cellcolor{blue!21} 8.2 [4.8, 12.2] & \cellcolor{blue!10} 3.9 [1.3, 7.0] & \cellcolor{blue!12} 4.6 [2.1, 7.6] & \cellcolor{blue!27} 10.5 [6.9, 14.7] & \cellcolor{blue!39} 15.2 [10.3, 20.6] & \cellcolor{blue!19} 7.4 [4.3, 11.0] \\
\midrule
\textbf{Discourse} & \textbf{Disc. markers} & \cellcolor{blue!12} 4.7 [2.0, 8.0] & \cellcolor{blue!4} 1.6 [0.1, 3.6] & \cellcolor{blue!14} 5.4 [2.8, 8.3] & \cellcolor{blue!6} 2.2 [0.8, 3.9] & \cellcolor{blue!5} 1.9 [0.4, 4.0] & \cellcolor{blue!8} 3.2 [0.8, 6.1] & \cellcolor{blue!16} 6.3 [3.4, 9.5] & \cellcolor{blue!22} 8.7 [5.2, 12.9] & \cellcolor{blue!11} 4.2 [1.9, 7.0] \\
  & \textbf{Appraisal} & \cellcolor{blue!14} 5.5 [2.9, 8.6] & \cellcolor{blue!7} 2.6 [0.8, 5.0] & \cellcolor{blue!12} 4.8 [2.4, 7.7] & \cellcolor{blue!8} 3.2 [1.4, 5.3] & \cellcolor{blue!8} 3.1 [1.0, 5.8] & \cellcolor{blue!8} 2.9 [0.9, 5.5] & \cellcolor{blue!18} 7.1 [4.2, 10.4] & \cellcolor{blue!36} 14.0 [9.2, 19.2] & \cellcolor{blue!14} 5.4 [2.9, 8.4] \\
\midrule
\textbf{Varieties} & \textbf{Style} & \cellcolor{blue!30} 11.8 [6.6, 17.7] & \cellcolor{blue!32} 12.2 [7.4, 17.7] & \cellcolor{blue!34} 13.1 [8.5, 18.3] & \cellcolor{blue!9} 3.6 [1.8, 5.7] & \cellcolor{blue!12} 4.8 [2.1, 8.0] & \cellcolor{blue!17} 6.4 [2.9, 10.6] & \cellcolor{blue!19} 7.4 [4.1, 11.2] & \cellcolor{blue!28} 10.9 [6.9, 15.3] & \cellcolor{blue!23} 8.8 [5.0, 13.1] \\
  & \textbf{Dialectal} & \cellcolor{blue!33} 12.6 [8.5, 17.3] & \cellcolor{blue!19} 7.4 [4.1, 11.4] & \cellcolor{blue!22} 8.4 [5.8, 11.4] & \cellcolor{blue!15} 5.9 [3.2, 9.3] & \cellcolor{blue!23} 8.9 [4.5, 14.0] & \cellcolor{blue!12} 4.7 [2.2, 7.9] & \cellcolor{blue!18} 7.0 [4.0, 10.6] & \cellcolor{blue!34} 13.2 [8.4, 18.3] & \cellcolor{blue!22} 8.5 [5.1, 12.5] \\
\midrule
\textbf{Average} & &  \cellcolor{blue!19} 7.4 [4.1, 11.4] & \cellcolor{blue!22} 8.6 [5.1, 12.6] & \cellcolor{blue!21} 8.3 [5.1, 12.0] & \cellcolor{blue!19} 7.5 [4.3, 11.4] & \cellcolor{blue!17} 6.7 [3.4, 10.8] & \cellcolor{blue!16} 6.4 [3.4, 10.0] & \cellcolor{blue!24} 9.3 [5.5, 13.7] & \cellcolor{blue!33} 13.0 [8.6, 17.9] & \cellcolor{blue!22} 8.4 [4.9, 12.5] \\
\bottomrule
\end{tabular}}
\caption{NER: Unrobustness (U, \%) by model and modification}\label{tab:ner_unrob_ci}
\end{table*}

\begin{table*}[ht]
\centering
\resizebox{\linewidth}{!}{
\begin{tabular}{llrrrrrrrrr}
\toprule
Category & Modification & \multicolumn{3}{c}{\textbf{PLM}} & \multicolumn{5}{c}{\textbf{LLM}} & \textbf{Avg} \\
 &  & \textbf{BERT} & \textbf{GPT-2} & \textbf{T5} & \textbf{GPT-4o} & \textbf{Claude-3.5} & \textbf{Llama 3.1} & \textbf{GPT-5} & \textbf{DS R1} & \textbf{Avg} \\
\midrule
\textbf{Bias} & \textbf{Temporal} & \cellcolor{blue!23} 9.0 [4.8, 16.2] & \cellcolor{blue!10} 4.0 [1.6, 9.8] & \cellcolor{blue!15} 6.0 [2.8, 12.5] & \cellcolor{blue!21} 8.0 [4.1, 15.0] & \cellcolor{blue!3} 1.0 [0.2, 5.4] & \cellcolor{blue!18} 7.0 [3.4, 13.7] & \cellcolor{blue!8} 3.0 [1.0, 8.5] & \cellcolor{blue!21} 8.0 [4.1, 15.0] & \cellcolor{blue!15} 5.8 [2.8, 12.0] \\
  & \textbf{Geographical} & \cellcolor{blue!21} 8.0 [4.1, 15.0] & \cellcolor{blue!36} 14.0 [8.5, 22.1] & \cellcolor{blue!23} 9.0 [4.8, 16.2] & \cellcolor{blue!26} 10.0 [5.5, 17.4] & \cellcolor{blue!10} 4.0 [1.6, 9.8] & \cellcolor{blue!13} 5.0 [2.2, 11.2] & \cellcolor{blue!15} 6.0 [2.8, 12.5] & \cellcolor{blue!26} 10.0 [5.5, 17.4] & \cellcolor{blue!21} 8.2 [4.4, 15.2] \\
  & \textbf{Length} & \cellcolor{blue!50} \textcolor{white}{19.2 [12.6, 28.0]} & \cellcolor{blue!39} 15.2 [9.4, 23.5] & \cellcolor{blue!36} 14.1 [8.6, 22.3] & \cellcolor{blue!39} 15.2 [9.4, 23.5] & \cellcolor{blue!52} \textcolor{white}{20.2 [13.5, 29.2]} & \cellcolor{blue!44} 17.2 [11.0, 25.8] & \cellcolor{blue!21} 8.1 [4.2, 15.1] & \cellcolor{blue!34} 13.1 [7.8, 21.2] & \cellcolor{blue!39} 15.3 [9.6, 23.6] \\
\midrule
\textbf{Orthography} & \textbf{Spelling} & \cellcolor{blue!29} 11.2 [6.4, 19.0] & \cellcolor{blue!8} 3.1 [1.0, 8.6] & \cellcolor{blue!18} 7.1 [3.5, 14.0] & \cellcolor{blue!13} 5.1 [2.2, 11.4] & \cellcolor{blue!5} 2.0 [0.6, 7.1] & \cellcolor{blue!11} 4.1 [1.6, 10.0] & \cellcolor{blue!18} 7.1 [3.5, 14.0] & \cellcolor{blue!11} 4.1 [1.6, 10.0] & \cellcolor{blue!14} 5.5 [2.5, 11.8] \\
  & \textbf{Capitalization} & \cellcolor{blue!39} 15.2 [9.4, 23.5] & \cellcolor{blue!36} 14.1 [8.6, 22.3] & \cellcolor{blue!18} 7.1 [3.5, 13.9] & \cellcolor{blue!18} 7.1 [3.5, 13.9] & \cellcolor{blue!10} 4.0 [1.6, 9.9] & \cellcolor{blue!13} 5.1 [2.2, 11.3] & \cellcolor{blue!5} 2.0 [0.6, 7.1] & \cellcolor{blue!16} 6.1 [2.8, 12.6] & \cellcolor{blue!20} 7.6 [4.0, 14.3] \\
  & \textbf{Punctuation} & \cellcolor{blue!3} 1.0 [0.2, 5.5] & \cellcolor{blue!18} 7.1 [3.5, 13.9] & \cellcolor{blue!3} 1.0 [0.2, 5.5] & \cellcolor{blue!3} 1.0 [0.2, 5.5] & \cellcolor{blue!8} 3.0 [1.0, 8.5] & \cellcolor{blue!3} 1.0 [0.2, 5.5] & \cellcolor{blue!5} 2.0 [0.6, 7.1] & \cellcolor{blue!10} 4.0 [1.6, 9.9] & \cellcolor{blue!7} 2.5 [0.9, 7.7] \\
\midrule
\textbf{Morphology} & \textbf{Derivation} & \cellcolor{blue!11} 4.1 [1.6, 10.0] & \cellcolor{blue!8} 3.1 [1.0, 8.6] & \cellcolor{blue!13} 5.1 [2.2, 11.4] & \cellcolor{blue!11} 4.1 [1.6, 10.0] & \cellcolor{blue!3} 1.0 [0.2, 5.6] & \cellcolor{blue!3} 1.0 [0.2, 5.6] & \cellcolor{blue!13} 5.1 [2.2, 11.4] & \cellcolor{blue!5} 2.0 [0.6, 7.1] & \cellcolor{blue!8} 3.2 [1.2, 8.7] \\
  & \textbf{Compound} & \cellcolor{blue!16} 6.2 [2.9, 13.0] & \cellcolor{blue!11} 4.2 [1.6, 10.2] & \cellcolor{blue!8} 3.1 [1.1, 8.8] & \cellcolor{blue!13} 5.2 [2.2, 11.6] & \cellcolor{blue!16} 6.2 [2.9, 13.0] & \cellcolor{blue!16} 6.2 [2.9, 13.0] & \cellcolor{blue!8} 3.1 [1.1, 8.8] & \cellcolor{blue!11} 4.2 [1.6, 10.2] & \cellcolor{blue!12} 4.8 [2.0, 11.1] \\
\midrule
\textbf{Syntax} & \textbf{Voice} & \cellcolor{blue!80} \textcolor{white}{35.8 [26.9, 45.8]} & \cellcolor{blue!80} \textcolor{white}{34.7 [25.9, 44.7]} & \cellcolor{blue!80} \textcolor{white}{41.1 [31.7, 51.1]} & \cellcolor{blue!68} \textcolor{white}{26.3 [18.5, 36.0]} & \cellcolor{blue!41} 15.8 [9.8, 24.4] & \cellcolor{blue!79} \textcolor{white}{30.5 [22.2, 40.4]} & \cellcolor{blue!24} 9.5 [5.1, 17.0] & \cellcolor{blue!41} 15.8 [9.8, 24.4] & \cellcolor{blue!68} \textcolor{white}{26.2 [18.7, 35.5]} \\
  & \textbf{Grammar} & \cellcolor{blue!79} \textcolor{white}{30.6 [21.1, 42.0]} & \cellcolor{blue!72} \textcolor{white}{27.8 [18.8, 39.0]} & \cellcolor{blue!50} \textcolor{white}{19.4 [12.0, 30.0]} & \cellcolor{blue!57} \textcolor{white}{22.2 [14.2, 33.1]} & \cellcolor{blue!47} \textcolor{white}{18.1 [10.9, 28.5]} & \cellcolor{blue!50} \textcolor{white}{19.4 [12.0, 30.0]} & \cellcolor{blue!43} 16.7 [9.8, 26.9] & \cellcolor{blue!54} \textcolor{white}{20.8 [13.1, 31.6]} & \cellcolor{blue!56} \textcolor{white}{21.9 [14.0, 32.6]} \\
  & \textbf{Conjunction} & \cellcolor{blue!13} 5.2 [2.2, 11.5] & \cellcolor{blue!27} 10.3 [5.7, 17.9] & \cellcolor{blue!21} 8.2 [4.2, 15.4] & \cellcolor{blue!21} 8.2 [4.2, 15.4] & \cellcolor{blue!11} 4.1 [1.6, 10.1] & \cellcolor{blue!19} 7.2 [3.5, 14.2] & \cellcolor{blue!13} 5.2 [2.2, 11.5] & \cellcolor{blue!16} 6.2 [2.9, 12.8] & \cellcolor{blue!18} 6.8 [3.3, 13.6] \\
\midrule
\textbf{Semantics} & \textbf{Concept} & \cellcolor{blue!21} 8.0 [4.1, 15.0] & \cellcolor{blue!8} 3.0 [1.0, 8.5] & \cellcolor{blue!13} 5.0 [2.2, 11.2] & \cellcolor{blue!28} 11.0 [6.3, 18.6] & \cellcolor{blue!46} \textcolor{white}{18.0 [11.7, 26.7]} & \cellcolor{blue!23} 9.0 [4.8, 16.2] & \cellcolor{blue!26} 10.0 [5.5, 17.4] & \cellcolor{blue!26} 10.0 [5.5, 17.4] & \cellcolor{blue!24} 9.2 [5.1, 16.4] \\
  & \textbf{Negation} & \cellcolor{blue!66} \textcolor{white}{25.5 [17.9, 35.0]} & \cellcolor{blue!61} \textcolor{white}{23.5 [16.2, 32.8]} & \cellcolor{blue!63} \textcolor{white}{24.5 [17.0, 33.9]} & \cellcolor{blue!55} \textcolor{white}{21.4 [14.5, 30.5]} & \cellcolor{blue!63} \textcolor{white}{24.5 [17.0, 33.9]} & \cellcolor{blue!58} \textcolor{white}{22.4 [15.3, 31.7]} & \cellcolor{blue!58} \textcolor{white}{22.4 [15.3, 31.7]} & \cellcolor{blue!79} \textcolor{white}{30.6 [22.4, 40.3]} & \cellcolor{blue!63} \textcolor{white}{24.4 [17.0, 33.7]} \\
\midrule
\textbf{Discourse} & \textbf{Disc. markers} & \cellcolor{blue!21} 8.0 [4.1, 15.0] & \cellcolor{blue!15} 6.0 [2.8, 12.5] & \cellcolor{blue!21} 8.0 [4.1, 15.0] & \cellcolor{blue!49} \textcolor{white}{19.0 [12.5, 27.8]} & \cellcolor{blue!26} 10.0 [5.5, 17.4] & \cellcolor{blue!18} 7.0 [3.4, 13.7] & \cellcolor{blue!23} 9.0 [4.8, 16.2] & \cellcolor{blue!28} 11.0 [6.3, 18.6] & \cellcolor{blue!25} 9.8 [5.4, 17.0] \\
  & \textbf{Appraisal} & \cellcolor{blue!13} 5.0 [2.2, 11.2] & \cellcolor{blue!21} 8.0 [4.1, 15.0] & \cellcolor{blue!10} 4.0 [1.6, 9.8] & \cellcolor{blue!21} 8.0 [4.1, 15.0] & \cellcolor{blue!18} 7.0 [3.4, 13.7] & \cellcolor{blue!23} 9.0 [4.8, 16.2] & \cellcolor{blue!13} 5.0 [2.2, 11.2] & \cellcolor{blue!18} 7.0 [3.4, 13.7] & \cellcolor{blue!17} 6.6 [3.2, 13.2] \\
\midrule
\textbf{Varieties} & \textbf{Style} & \cellcolor{blue!41} 16.0 [10.1, 24.4] & \cellcolor{blue!49} \textcolor{white}{19.0 [12.5, 27.8]} & \cellcolor{blue!46} \textcolor{white}{18.0 [11.7, 26.7]} & \cellcolor{blue!49} \textcolor{white}{19.0 [12.5, 27.8]} & \cellcolor{blue!36} 14.0 [8.5, 22.1] & \cellcolor{blue!36} 14.0 [8.5, 22.1] & \cellcolor{blue!36} 14.0 [8.5, 22.1] & \cellcolor{blue!31} 12.0 [7.0, 19.8] & \cellcolor{blue!41} 15.8 [9.9, 24.1] \\
  & \textbf{Dialect} & \cellcolor{blue!22} 8.3 [4.4, 15.1] & \cellcolor{blue!69} \textcolor{white}{26.9 [19.4, 35.9]} & \cellcolor{blue!57} \textcolor{white}{22.2 [15.4, 30.9]} & \cellcolor{blue!63} \textcolor{white}{24.5 [17.2, 33.7]} & \cellcolor{blue!30} 11.8 [6.9, 19.4] & \cellcolor{blue!43} 16.7 [10.7, 25.1] & \cellcolor{blue!8} 2.9 [1.0, 8.3] & \cellcolor{blue!33} 12.7 [7.6, 20.6] & \cellcolor{blue!41} 15.8 [10.3, 23.6] \\
\midrule
\textbf{Average} & &  \cellcolor{blue!33} 12.7 [7.9, 20.3] & \cellcolor{blue!34} 13.2 [8.3, 20.8] & \cellcolor{blue!31} 11.9 [7.4, 19.3] & \cellcolor{blue!33} 12.7 [7.8, 20.4] & \cellcolor{blue!25} 9.7 [5.7, 16.8] & \cellcolor{blue!28} 10.7 [6.4, 18.0] & \cellcolor{blue!20} 7.7 [4.1, 14.5] & \cellcolor{blue!27} 10.5 [6.1, 17.8] & \cellcolor{blue!29} 11.1 [6.7, 18.5] \\
\bottomrule
\end{tabular}}
\caption{Coreference: Unrobustness (U, \%) by model and modification}
\label{tab:coref_unrob_ci}
\end{table*}

\subsection{Use Case: Dialogue Contradiction Detection and Sentiment Analysis (extended results)}
\label{app:dialogue}

In this section we introduce two additional classification tasks -- Dialogue Contradiction Detection and Sentiment Analysis.

We use the DECODE dataset \citep{nie-etal-2021-like} to evaluate dialogue understanding capabilities. 
Each sample is a multi-turn dialogue, where the final turn may be consistent with (0) or contradict (1) the preceding context. 
For Sentiment Analysis, We chose the SST-2 dataset \citep{socher-etal-2013-recursive}, containing around 11K sentences extracted from movies reviews with either 0 (negative) or 1 (positive) sentiment. 

As \Cref{tab:dialogue_unrob} shows, on Dialogue contradiction detection task base LLMs are overall more robust to modifications than PLMs; however, reasoning LLMs on average perform similar to LLMs. All models show major brittleness to negation, and the majority of them also lack robustness to change of \textbf{grammatical roles}, which shows that they tend to rely on surface-level, token consistency between the context and the final turn, rather than on logical coherence. Moreover, LLMs (but to a lesser extent PLMs) demonstrate \textbf{geographical bias}, whuile PLMs are particularly brittle to \textbf{casual style}. 

\begin{table*}[ht]
\centering
\resizebox{\linewidth}{!}{
\begin{tabular}{llrrrrrrrrr}
\toprule
Category & Modification & \textbf{BERT} & \textbf{GPT-2} & \textbf{T5} & \textbf{GPT-4o} & \textbf{Claude-3.5} & \textbf{Llama 3.1} & \textbf{GPT-5} & \textbf{DS R1} & \textbf{Avg} \\
\midrule
\textbf{Bias} & \textbf{Temporal} & \cellcolor{blue!21} 8.0 [4.1, 15.0] & \cellcolor{blue!15} 6.0 [2.8, 12.5] & \cellcolor{blue!18} 7.0 [3.4, 13.7] & \cellcolor{blue!5} 2.0 [0.6, 7.0] & \cellcolor{blue!13} 5.0 [2.2, 11.2] & \cellcolor{blue!3} 1.0 [0.2, 5.4] & \cellcolor{blue!8} 3.0 [1.0, 8.5] & \cellcolor{blue!5} 2.0 [0.6, 7.0] & \cellcolor{blue!11} 4.2 [1.8, 10.0] \\
  & \textbf{Geographical} & \cellcolor{blue!17} 6.5 [3.0, 13.5] & \cellcolor{blue!22} 8.7 [4.5, 16.2] & \cellcolor{blue!31} 12.0 [6.8, 20.2] & \cellcolor{blue!34} 13.0 [7.6, 21.4] & \cellcolor{blue!39} 15.2 [9.3, 23.9] & \cellcolor{blue!31} 12.0 [6.8, 20.2] & \cellcolor{blue!48} \textcolor{white}{18.5 [11.9, 27.6]} & \cellcolor{blue!34} 13.0 [7.6, 21.4] & \cellcolor{blue!32} 12.4 [7.2, 20.6] \\
  & \textbf{Length} & \cellcolor{blue!26} 10.0 [5.5, 17.4] & \cellcolor{blue!15} 6.0 [2.8, 12.5] & \cellcolor{blue!13} 5.0 [2.2, 11.2] & \cellcolor{blue!10} 4.0 [1.6, 9.8] & \cellcolor{blue!15} 6.0 [2.8, 12.5] & \cellcolor{blue!3} 1.0 [0.2, 5.4] & \cellcolor{blue!15} 6.0 [2.8, 12.5] & \cellcolor{blue!18} 7.0 [3.4, 13.7] & \cellcolor{blue!15} 5.6 [2.6, 11.9] \\
\midrule
\textbf{Orthographic} & \textbf{Spelling} & \cellcolor{blue!15} 6.0 [2.8, 12.5] & \cellcolor{blue!15} 6.0 [2.8, 12.5] & \cellcolor{blue!10} 4.0 [1.6, 9.8] & \cellcolor{blue!21} 8.0 [4.1, 15.0] & \cellcolor{blue!8} 3.0 [1.0, 8.5] & \cellcolor{blue!5} 2.0 [0.6, 7.0] & \cellcolor{blue!3} 1.0 [0.2, 5.4] & \cellcolor{blue!15} 6.0 [2.8, 12.5] & \cellcolor{blue!12} 4.5 [2.0, 10.4] \\
  & \textbf{Capitalization} & \cellcolor{blue!0} 0.0 [0.0, 0.0] & \cellcolor{blue!16} 6.2 [2.9, 13.0] & \cellcolor{blue!16} 6.2 [2.9, 13.0] & \cellcolor{blue!8} 3.1 [1.1, 8.8] & \cellcolor{blue!5} 2.1 [0.6, 7.3] & \cellcolor{blue!5} 2.1 [0.6, 7.3] & \cellcolor{blue!11} 4.2 [1.6, 10.2] & \cellcolor{blue!3} 1.0 [0.2, 5.7] & \cellcolor{blue!8} 3.1 [1.2, 8.1] \\
  & \textbf{Punctuation} & \cellcolor{blue!10} 4.0 [1.6, 9.8] & \cellcolor{blue!5} 2.0 [0.6, 7.0] & \cellcolor{blue!21} 8.0 [4.1, 15.0] & \cellcolor{blue!8} 3.0 [1.0, 8.5] & \cellcolor{blue!8} 3.0 [1.0, 8.5] & \cellcolor{blue!3} 1.0 [0.2, 5.4] & \cellcolor{blue!13} 5.0 [2.2, 11.2] & \cellcolor{blue!5} 2.0 [0.6, 7.0] & \cellcolor{blue!9} 3.5 [1.4, 9.0] \\
\midrule
\textbf{Semantic} & \textbf{Concept} & \cellcolor{blue!18} 7.0 [3.4, 13.7] & \cellcolor{blue!13} 5.0 [2.2, 11.2] & \cellcolor{blue!10} 4.0 [1.6, 9.8] & \cellcolor{blue!21} 8.0 [4.1, 15.0] & \cellcolor{blue!18} 7.0 [3.4, 13.7] & \cellcolor{blue!5} 2.0 [0.6, 7.0] & \cellcolor{blue!26} 10.0 [5.5, 17.4] & \cellcolor{blue!15} 6.0 [2.8, 12.5] & \cellcolor{blue!16} 6.1 [2.9, 12.6] \\
  & \textbf{Negation} & \cellcolor{blue!75} \textcolor{white}{29.0 [21.0, 38.5]} & \cellcolor{blue!80} \textcolor{white}{39.0 [30.0, 48.8]} & \cellcolor{blue!80} \textcolor{white}{31.0 [22.8, 40.6]} & \cellcolor{blue!80} \textcolor{white}{31.0 [22.8, 40.6]} & \cellcolor{blue!77} \textcolor{white}{30.0 [21.9, 39.6]} & \cellcolor{blue!77} \textcolor{white}{30.0 [21.9, 39.6]} & \cellcolor{blue!65} \textcolor{white}{25.0 [17.5, 34.3]} & \cellcolor{blue!65} \textcolor{white}{25.0 [17.5, 34.3]} & \cellcolor{blue!77} \textcolor{white}{30.0 [21.9, 39.5]} \\
\midrule
\textbf{Pragmatic} & \textbf{Appraisal} & \cellcolor{blue!15} 6.0 [2.8, 12.5] & \cellcolor{blue!26} 10.0 [5.5, 17.4] & \cellcolor{blue!13} 5.0 [2.2, 11.2] & \cellcolor{blue!18} 7.0 [3.4, 13.7] & \cellcolor{blue!15} 6.0 [2.8, 12.5] & \cellcolor{blue!10} 4.0 [1.6, 9.8] & \cellcolor{blue!26} 10.0 [5.5, 17.4] & \cellcolor{blue!28} 11.0 [6.3, 18.6] & \cellcolor{blue!19} 7.4 [3.8, 14.2] \\
\midrule
\textbf{Genre} & \textbf{Style} & \cellcolor{blue!31} 12.0 [7.0, 19.8] & \cellcolor{blue!26} 10.0 [5.5, 17.4] & \cellcolor{blue!31} 12.0 [7.0, 19.8] & \cellcolor{blue!15} 6.0 [2.8, 12.5] & \cellcolor{blue!10} 4.0 [1.6, 9.8] & \cellcolor{blue!5} 2.0 [0.6, 7.0] & \cellcolor{blue!15} 6.0 [2.8, 12.5] & \cellcolor{blue!15} 6.0 [2.8, 12.5] & \cellcolor{blue!19} 7.2 [3.7, 13.9] \\
  & \textbf{Dialect} & \cellcolor{blue!54} \textcolor{white}{20.8 [14.1, 29.4]} & \cellcolor{blue!15} 5.7 [2.6, 11.8] & \cellcolor{blue!24} 9.4 [5.2, 16.5] & \cellcolor{blue!29} 11.3 [6.6, 18.8] & \cellcolor{blue!39} 15.1 [9.5, 23.1] & \cellcolor{blue!19} 7.5 [3.9, 14.2] & \cellcolor{blue!15} 5.7 [2.6, 11.8] & \cellcolor{blue!34} 13.2 [8.0, 21.0] & \cellcolor{blue!29} 11.1 [6.6, 18.3] \\
\midrule
\textbf{Syntactic} & \textbf{Conjunction} & \cellcolor{blue!13} 5.0 [2.2, 11.2] & \cellcolor{blue!10} 4.0 [1.6, 9.8] & \cellcolor{blue!18} 7.0 [3.4, 13.7] & \cellcolor{blue!15} 6.0 [2.8, 12.5] & \cellcolor{blue!3} 1.0 [0.2, 5.4] & \cellcolor{blue!5} 2.0 [0.6, 7.0] & \cellcolor{blue!10} 4.0 [1.6, 9.8] & \cellcolor{blue!10} 4.0 [1.6, 9.8] & \cellcolor{blue!11} 4.1 [1.7, 9.9] \\
  & \textbf{Voice} & \cellcolor{blue!23} 9.0 [4.8, 16.2] & \cellcolor{blue!28} 11.0 [6.3, 18.6] & \cellcolor{blue!18} 7.0 [3.4, 13.7] & \cellcolor{blue!18} 7.0 [3.4, 13.7] & \cellcolor{blue!5} 2.0 [0.6, 7.0] & \cellcolor{blue!8} 3.0 [1.0, 8.5] & \cellcolor{blue!23} 9.0 [4.8, 16.2] & \cellcolor{blue!13} 5.0 [2.2, 11.2] & \cellcolor{blue!17} 6.6 [3.3, 13.2] \\
  & \textbf{Role} & \cellcolor{blue!37} 14.3 [7.9, 24.3] & \cellcolor{blue!33} 12.9 [6.9, 22.7] & \cellcolor{blue!18} 7.1 [3.1, 15.7] & \cellcolor{blue!53} \textcolor{white}{20.6 [12.7, 31.6]} & \cellcolor{blue!19} 7.4 [3.2, 16.1] & \cellcolor{blue!30} 11.8 [6.1, 21.5] & \cellcolor{blue!42} 16.2 [9.3, 26.7] & \cellcolor{blue!49} \textcolor{white}{19.1 [11.5, 30.0]} & \cellcolor{blue!35} 13.7 [7.6, 23.6] \\
\midrule
\textbf{Morphological} & \textbf{Derivation} & \cellcolor{blue!0} 0.0 [0.0, 0.0] & \cellcolor{blue!6} 2.2 [0.6, 7.5] & \cellcolor{blue!3} 1.1 [0.2, 5.8] & \cellcolor{blue!11} 4.3 [1.7, 10.5] & \cellcolor{blue!14} 5.4 [2.3, 12.0] & \cellcolor{blue!3} 1.1 [0.2, 5.8] & \cellcolor{blue!14} 5.4 [2.3, 12.0] & \cellcolor{blue!17} 6.5 [3.0, 13.4] & \cellcolor{blue!8} 3.2 [1.3, 8.4] \\
  & \textbf{Compound} & \cellcolor{blue!15} 6.0 [2.8, 12.5] & \cellcolor{blue!10} 4.0 [1.6, 9.8] & \cellcolor{blue!8} 3.0 [1.0, 8.5] & \cellcolor{blue!10} 4.0 [1.6, 9.8] & \cellcolor{blue!18} 7.0 [3.4, 13.7] & \cellcolor{blue!3} 1.0 [0.2, 5.4] & \cellcolor{blue!21} 8.0 [4.1, 15.0] & \cellcolor{blue!15} 6.0 [2.8, 12.5] & \cellcolor{blue!13} 4.9 [2.2, 10.9] \\
  & \textbf{Discourse} & \cellcolor{blue!15} 5.7 [2.5, 12.8] & \cellcolor{blue!24} 9.2 [4.7, 17.1] & \cellcolor{blue!9} 3.4 [1.2, 9.7] & \cellcolor{blue!9} 3.4 [1.2, 9.7] & \cellcolor{blue!6} 2.3 [0.6, 8.0] & \cellcolor{blue!9} 3.4 [1.2, 9.7] & \cellcolor{blue!15} 5.7 [2.5, 12.8] & \cellcolor{blue!15} 5.7 [2.5, 12.8] & \cellcolor{blue!13} 4.9 [2.0, 11.5] \\
\midrule
\textbf{Average} &  & \cellcolor{blue!23} 8.8 [5.0, 15.2] & \cellcolor{blue!22} 8.7 [4.9, 15.6] & \cellcolor{blue!20} 7.8 [4.2, 14.6] & \cellcolor{blue!22} 8.3 [4.6, 15.2] & \cellcolor{blue!18} 7.1 [3.9, 13.7] & \cellcolor{blue!13} 5.1 [2.7, 11.0] & \cellcolor{blue!22} 8.4 [4.6, 15.4] & \cellcolor{blue!21} 8.2 [4.5, 15.0] & \cellcolor{blue!20} 7.8 [4.3, 14.5] \\
\bottomrule
\end{tabular}}
\caption{Dialogue: Unrobustness (U, \%) by model and modification}
\label{tab:dialogue_unrob}
\end{table*}


On Sentiment Analysis task, overall, PLMs were less robust to modifications compared to LLMs (see \Cref{tab:sa_unrob}), with the exception of Claude-3.5 which lacks robusteness to many tests that the other LLMs are stable on. Most prominently, all LLMs and PLMs are significantly unrobust to \textbf{negation}, and can be easily misguided by adding another \textbf{appraisal} marker to a sentence already bearing some sentiment. 

\begin{table*}[ht]
\centering
\resizebox{\linewidth}{!}{
\begin{tabular}{llrrrrrrrrr}
\toprule
Category & Modification & \multicolumn{3}{c}{\textbf{PLM}} & \multicolumn{5}{c}{\textbf{LLM}} \\
 &  & \textbf{BERT} & \textbf{GPT-2} & \textbf{T5} & \textbf{GPT-4o} & \textbf{Claude-3.5} & \textbf{Llama 3.1} & \textbf{GPT-5} & \textbf{DS R1} & \textbf{Avg} \\
\midrule
\textbf{Bias} & \textbf{Temporal} & \cellcolor{blue!13} 5.0 [2.2, 11.2] & \cellcolor{blue!13} 5.0 [2.2, 11.2] & \cellcolor{blue!8} 3.0 [1.0, 8.5] & \cellcolor{blue!0} 0.0 [0.0, 0.0] & \cellcolor{blue!21} 8.0 [4.1, 15.0] & \cellcolor{blue!3} 1.0 [0.2, 5.4] & \cellcolor{blue!8} 3.0 [1.0, 8.5] & \cellcolor{blue!8} 3.0 [1.0, 8.5] & \cellcolor{blue!9} 3.5 [1.5, 8.5] \\
  & \textbf{Geographical} & \cellcolor{blue!13} 5.0 [2.2, 11.2] & \cellcolor{blue!10} 4.0 [1.6, 9.8] & \cellcolor{blue!8} 3.0 [1.0, 8.5] & \cellcolor{blue!23} 9.0 [4.8, 16.2] & \cellcolor{blue!21} 8.0 [4.1, 15.0] & \cellcolor{blue!18} 7.0 [3.4, 13.7] & \cellcolor{blue!10} 4.0 [1.6, 9.8] & \cellcolor{blue!21} 8.0 [4.1, 15.0] & \cellcolor{blue!15} 6.0 [2.8, 12.4] \\
  & \textbf{Length} & \cellcolor{blue!8} 3.0 [1.0, 8.5] & \cellcolor{blue!5} 2.0 [0.6, 7.0] & \cellcolor{blue!8} 3.0 [1.0, 8.5] & \cellcolor{blue!8} 3.0 [1.0, 8.5] & \cellcolor{blue!23} 9.0 [4.8, 16.2] & \cellcolor{blue!5} 2.0 [0.6, 7.0] & \cellcolor{blue!3} 1.0 [0.2, 5.4] & \cellcolor{blue!3} 1.0 [0.2, 5.4] & \cellcolor{blue!8} 3.0 [1.2, 8.3] \\
\midrule
\textbf{Orthographic} & \textbf{Spelling} & \cellcolor{blue!10} 4.0 [1.6, 9.8] & \cellcolor{blue!10} 4.0 [1.6, 9.8] & \cellcolor{blue!0} 0.0 [0.0, 0.0] & \cellcolor{blue!0} 0.0 [0.0, 0.0] & \cellcolor{blue!15} 6.0 [2.8, 12.5] & \cellcolor{blue!8} 3.0 [1.0, 8.5] & \cellcolor{blue!0} 0.0 [0.0, 0.0] & \cellcolor{blue!0} 0.0 [0.0, 0.0] & \cellcolor{blue!5} 2.1 [0.9, 5.1] \\
  & \textbf{Capitalization} & \cellcolor{blue!0} 0.0 [0.0, 0.0] & \cellcolor{blue!8} 3.0 [1.0, 8.5] & \cellcolor{blue!8} 3.0 [1.0, 8.5] & \cellcolor{blue!3} 1.0 [0.2, 5.5] & \cellcolor{blue!18} 7.1 [3.5, 13.9] & \cellcolor{blue!8} 3.0 [1.0, 8.5] & \cellcolor{blue!0} 0.0 [0.0, 0.0] & \cellcolor{blue!8} 3.0 [1.0, 8.5] & \cellcolor{blue!7} 2.5 [1.0, 6.7] \\
  & \textbf{Punctuation} & \cellcolor{blue!3} 1.0 [0.2, 5.4] & \cellcolor{blue!5} 2.0 [0.6, 7.0] & \cellcolor{blue!0} 0.0 [0.0, 0.0] & \cellcolor{blue!3} 1.0 [0.2, 5.4] & \cellcolor{blue!8} 3.0 [1.0, 8.5] & \cellcolor{blue!8} 3.0 [1.0, 8.5] & \cellcolor{blue!3} 1.0 [0.2, 5.4] & \cellcolor{blue!3} 1.0 [0.2, 5.4] & \cellcolor{blue!4} 1.5 [0.4, 5.7] \\
\midrule
\textbf{Morphological} & \textbf{Derivation} & \cellcolor{blue!9} 3.4 [1.2, 9.7] & \cellcolor{blue!15} 5.7 [2.5, 12.8] & \cellcolor{blue!6} 2.3 [0.6, 8.0] & \cellcolor{blue!6} 2.3 [0.6, 8.0] & \cellcolor{blue!18} 6.9 [3.2, 14.2] & \cellcolor{blue!6} 2.3 [0.6, 8.0] & \cellcolor{blue!9} 3.4 [1.2, 9.7] & \cellcolor{blue!12} 4.6 [1.8, 11.2] & \cellcolor{blue!10} 3.9 [1.5, 10.2] \\
  & \textbf{Compound} & \cellcolor{blue!8} 3.2 [1.1, 8.9] & \cellcolor{blue!22} 8.4 [4.3, 15.7] & \cellcolor{blue!8} 3.2 [1.1, 8.9] & \cellcolor{blue!5} 2.1 [0.6, 7.4] & \cellcolor{blue!14} 5.3 [2.3, 11.7] & \cellcolor{blue!3} 1.1 [0.2, 5.7] & \cellcolor{blue!8} 3.2 [1.1, 8.9] & \cellcolor{blue!11} 4.2 [1.6, 10.3] & \cellcolor{blue!10} 3.8 [1.5, 9.7] \\
\midrule
\textbf{Syntactic} & \textbf{Voice} & \cellcolor{blue!23} 9.0 [4.8, 16.2] & \cellcolor{blue!26} 10.0 [5.5, 17.4] & \cellcolor{blue!23} 9.0 [4.8, 16.2] & \cellcolor{blue!10} 4.0 [1.6, 9.8] & \cellcolor{blue!26} 10.0 [5.5, 17.4] & \cellcolor{blue!8} 3.0 [1.0, 8.5] & \cellcolor{blue!13} 5.0 [2.2, 11.2] & \cellcolor{blue!10} 4.0 [1.6, 9.8] & \cellcolor{blue!17} 6.8 [3.4, 13.3] \\
  & \textbf{Grammar} & \cellcolor{blue!12} 4.5 [1.6, 12.5] & \cellcolor{blue!31} 12.1 [6.3, 22.1] & \cellcolor{blue!8} 3.0 [0.8, 10.4] & \cellcolor{blue!12} 4.5 [1.6, 12.5] & \cellcolor{blue!20} 7.6 [3.3, 16.5] & \cellcolor{blue!12} 4.5 [1.6, 12.5] & \cellcolor{blue!12} 4.5 [1.6, 12.5] & \cellcolor{blue!12} 4.5 [1.6, 12.5] & \cellcolor{blue!15} 5.7 [2.3, 14.0] \\
  & \textbf{Conjunction} & \cellcolor{blue!8} 3.0 [1.0, 8.5] & \cellcolor{blue!15} 6.0 [2.8, 12.5] & \cellcolor{blue!3} 1.0 [0.2, 5.4] & \cellcolor{blue!3} 1.0 [0.2, 5.4] & \cellcolor{blue!8} 3.0 [1.0, 8.5] & \cellcolor{blue!8} 3.0 [1.0, 8.5] & \cellcolor{blue!0} 0.0 [0.0, 0.0] & \cellcolor{blue!0} 0.0 [0.0, 0.0] & \cellcolor{blue!5} 2.1 [0.8, 6.1] \\
\midrule
\textbf{Semantic} & \textbf{Concept} & \cellcolor{blue!15} 6.0 [2.8, 12.5] & \cellcolor{blue!15} 6.0 [2.8, 12.5] & \cellcolor{blue!13} 5.0 [2.2, 11.2] & \cellcolor{blue!10} 4.0 [1.6, 9.8] & \cellcolor{blue!10} 4.0 [1.6, 9.8] & \cellcolor{blue!5} 2.0 [0.6, 7.0] & \cellcolor{blue!3} 1.0 [0.2, 5.4] & \cellcolor{blue!13} 5.0 [2.2, 11.2] & \cellcolor{blue!11} 4.1 [1.7, 9.9] \\
  & \textbf{Negation} & \cellcolor{blue!59} \textcolor{white}{22.9 [15.6, 32.3]} & \cellcolor{blue!54} \textcolor{white}{20.8 [13.9, 30.0]} & \cellcolor{blue!65} \textcolor{white}{25.0 [17.4, 34.5]} & \cellcolor{blue!43} 16.7 [10.5, 25.4] & \cellcolor{blue!46} \textcolor{white}{17.7 [11.4, 26.5]} & \cellcolor{blue!40} 15.6 [9.7, 24.2] & \cellcolor{blue!43} 16.7 [10.5, 25.4] & \cellcolor{blue!46} \textcolor{white}{17.7 [11.4, 26.5]} & \cellcolor{blue!49} \textcolor{white}{19.1 [12.6, 28.1]} \\
\midrule
\textbf{Pragmatic} & \textbf{Discourse} & \cellcolor{blue!8} 3.0 [1.0, 8.5] & \cellcolor{blue!16} 6.1 [2.8, 12.6] & \cellcolor{blue!10} 4.0 [1.6, 9.9] & \cellcolor{blue!3} 1.0 [0.2, 5.5] & \cellcolor{blue!31} 12.1 [7.1, 20.0] & \cellcolor{blue!8} 3.0 [1.0, 8.5] & \cellcolor{blue!5} 2.0 [0.6, 7.1] & \cellcolor{blue!8} 3.0 [1.0, 8.5] & \cellcolor{blue!11} 4.3 [1.9, 10.1] \\
  & \textbf{Sentiment} & \cellcolor{blue!49} \textcolor{white}{19.0 [12.5, 27.8]} & \cellcolor{blue!41} 16.0 [10.1, 24.4] & \cellcolor{blue!31} 12.0 [7.0, 19.8] & \cellcolor{blue!36} 14.0 [8.5, 22.1] & \cellcolor{blue!39} 15.0 [9.3, 23.3] & \cellcolor{blue!41} 16.0 [10.1, 24.4] & \cellcolor{blue!28} 11.0 [6.3, 18.6] & \cellcolor{blue!26} 10.0 [5.5, 17.4] & \cellcolor{blue!36} 14.1 [8.7, 22.2] \\
\midrule
\textbf{Genre} & \textbf{Casual} & \cellcolor{blue!23} 9.0 [4.8, 16.2] & \cellcolor{blue!23} 9.0 [4.8, 16.2] & \cellcolor{blue!13} 5.0 [2.2, 11.2] & \cellcolor{blue!8} 3.0 [1.0, 8.5] & \cellcolor{blue!18} 7.0 [3.4, 13.7] & \cellcolor{blue!15} 6.0 [2.8, 12.5] & \cellcolor{blue!8} 3.0 [1.0, 8.5] & \cellcolor{blue!5} 2.0 [0.6, 7.0] & \cellcolor{blue!14} 5.5 [2.6, 11.7] \\
  & \textbf{Dialectal} & \cellcolor{blue!25} 9.8 [5.4, 17.1] & \cellcolor{blue!25} 9.8 [5.4, 17.1] & \cellcolor{blue!20} 7.8 [4.0, 14.7] & \cellcolor{blue!13} 4.9 [2.1, 11.0] & \cellcolor{blue!18} 6.9 [3.4, 13.5] & \cellcolor{blue!15} 5.9 [2.7, 12.2] & \cellcolor{blue!8} 2.9 [1.0, 8.3] & \cellcolor{blue!10} 3.9 [1.5, 9.7] & \cellcolor{blue!17} 6.5 [3.2, 12.9] \\
\midrule
\textbf{Average} &  & \cellcolor{blue!17} 6.5 [3.5, 12.7] & \cellcolor{blue!20} 7.6 [4.0, 14.5] & \cellcolor{blue!14} 5.3 [2.7, 10.8] & \cellcolor{blue!11} 4.2 [2.0, 9.5] & \cellcolor{blue!21} 8.0 [4.2, 15.1] & \cellcolor{blue!12} 4.8 [2.3, 10.8] & \cellcolor{blue!9} 3.6 [1.7, 8.5] & \cellcolor{blue!11} 4.4 [2.1, 9.8] & \cellcolor{blue!14} 5.6 [2.8, 11.5] \\
\bottomrule
\end{tabular}}
\caption{Sentiment Analysis: Unrobustness (U, \%) by model and modification}
\label{tab:sa_unrob}
\end{table*}

\subsection{Use Case: GSM and IFEval (extended results)}

We present the full results of GSM and IFEval tasks with CI intervals in \Cref{tab:gsm_unrob_ci} and \Cref{tab:ifeval_unrob_ci}.

\begin{table*}[ht]
\centering
\resizebox{\linewidth}{!}{
\begin{tabular}{llrrrrrr}
\toprule
Category & Modification & \textbf{GPT-4o} & \textbf{Claude-3.5} & \textbf{Llama 3.1} & \textbf{GPT-5} & \textbf{DS R1} & \textbf{Avg} \\
\midrule
\textbf{Bias} & \textbf{Temporal} & \cellcolor{blue!3} 1.0 [0.2, 5.4] & \cellcolor{blue!5} 2.0 [0.6, 7.0] & \cellcolor{blue!10} 4.0 [1.6, 9.8] & \cellcolor{blue!0} 0.0 [0.0, 0.0] & \cellcolor{blue!3} 1.0 [0.2, 5.4] & \cellcolor{blue!4} 1.6 [0.5, 5.5] \\
  & \textbf{Geographical} & \cellcolor{blue!13} 5.0 [2.2, 11.2] & \cellcolor{blue!13} 5.0 [2.2, 11.2] & \cellcolor{blue!18} 7.0 [3.4, 13.7] & \cellcolor{blue!3} 1.0 [0.2, 5.4] & \cellcolor{blue!5} 2.0 [0.6, 7.0] & \cellcolor{blue!10} 4.0 [1.7, 9.7] \\
  & \textbf{Length} & \cellcolor{blue!10} 4.0 [1.6, 9.8] & \cellcolor{blue!5} 2.0 [0.6, 7.0] & \cellcolor{blue!5} 2.0 [0.6, 7.0] & \cellcolor{blue!0} 0.0 [0.0, 0.0] & \cellcolor{blue!5} 2.0 [0.6, 7.0] & \cellcolor{blue!5} 2.0 [0.6, 6.2] \\
\midrule
\textbf{Orthographic} & \textbf{Spelling} & \cellcolor{blue!3} 1.0 [0.2, 5.4] & \cellcolor{blue!0} 0.0 [0.0, 0.0] & \cellcolor{blue!5} 2.0 [0.6, 7.0] & \cellcolor{blue!3} 1.0 [0.2, 5.4] & \cellcolor{blue!0} 0.0 [0.0, 0.0] & \cellcolor{blue!2} 0.8 [0.2, 3.6] \\
  & \textbf{Capitalization} & \cellcolor{blue!8} 3.0 [1.0, 8.5] & \cellcolor{blue!3} 1.0 [0.2, 5.4] & \cellcolor{blue!13} 5.0 [2.2, 11.2] & \cellcolor{blue!3} 1.0 [0.2, 5.4] & \cellcolor{blue!3} 1.0 [0.2, 5.4] & \cellcolor{blue!6} 2.2 [0.7, 7.2] \\
  & \textbf{Punctuation} & \cellcolor{blue!0} 0.0 [0.0, 0.0] & \cellcolor{blue!0} 0.0 [0.0, 0.0] & \cellcolor{blue!13} 5.0 [2.2, 11.2] & \cellcolor{blue!3} 1.0 [0.2, 5.4] & \cellcolor{blue!3} 1.0 [0.2, 5.4] & \cellcolor{blue!4} 1.4 [0.5, 4.4] \\
\midrule
\textbf{Semantic} & \textbf{Concept} & \cellcolor{blue!3} 1.0 [0.2, 5.4] & \cellcolor{blue!0} 0.0 [0.0, 0.0] & \cellcolor{blue!13} 5.0 [2.2, 11.2] & \cellcolor{blue!5} 2.0 [0.6, 7.0] & \cellcolor{blue!3} 1.0 [0.2, 5.4] & \cellcolor{blue!5} 1.8 [0.6, 5.8] \\
  & \textbf{Negation} & \cellcolor{blue!39} 15.0 [9.3, 23.3] & \cellcolor{blue!39} 15.0 [9.3, 23.3] & \cellcolor{blue!46} \textcolor{white}{18.0 [11.7, 26.7]} & \cellcolor{blue!18} 7.0 [3.4, 13.7] & \cellcolor{blue!23} 9.0 [4.8, 16.2] & \cellcolor{blue!33} 12.8 [7.7, 20.6] \\
\midrule
\textbf{Discourse} & \textbf{Appraisal} & \cellcolor{blue!0} 0.0 [0.0, 0.0] & \cellcolor{blue!0} 0.0 [0.0, 0.0] & \cellcolor{blue!3} 1.0 [0.2, 5.4] & \cellcolor{blue!3} 1.0 [0.2, 5.4] & \cellcolor{blue!3} 1.0 [0.2, 5.4] & \cellcolor{blue!2} 0.6 [0.1, 3.3] \\
\midrule
\textbf{Varieties} & \textbf{Style} & \cellcolor{blue!10} 4.0 [1.6, 9.8] & \cellcolor{blue!8} 3.0 [1.0, 8.5] & \cellcolor{blue!15} 6.0 [2.8, 12.5] & \cellcolor{blue!10} 4.0 [1.6, 9.8] & \cellcolor{blue!8} 3.0 [1.0, 8.5] & \cellcolor{blue!10} 4.0 [1.6, 9.8] \\
  & \textbf{Dialect} & \cellcolor{blue!5} 2.0 [0.6, 7.0] & \cellcolor{blue!10} 4.0 [1.6, 9.8] & \cellcolor{blue!15} 6.0 [2.8, 12.5] & \cellcolor{blue!3} 1.0 [0.2, 5.4] & \cellcolor{blue!5} 2.0 [0.6, 7.0] & \cellcolor{blue!8} 3.0 [1.1, 8.4] \\
\midrule
\textbf{Syntactic} & \textbf{Conjunction} & \cellcolor{blue!0} 0.0 [0.0, 0.0] & \cellcolor{blue!3} 1.0 [0.2, 5.4] & \cellcolor{blue!10} 4.0 [1.6, 9.8] & \cellcolor{blue!3} 1.0 [0.2, 5.4] & \cellcolor{blue!3} 1.0 [0.2, 5.4] & \cellcolor{blue!4} 1.4 [0.4, 5.2] \\
  & \textbf{Voice} & \cellcolor{blue!5} 2.0 [0.6, 7.0] & \cellcolor{blue!5} 2.0 [0.6, 7.0] & \cellcolor{blue!15} 6.0 [2.8, 12.5] & \cellcolor{blue!3} 1.0 [0.2, 5.4] & \cellcolor{blue!5} 2.0 [0.6, 7.0] & \cellcolor{blue!7} 2.6 [0.9, 7.8] \\
\midrule
\textbf{Average} &  & \cellcolor{blue!8} 2.9 [1.3, 7.1] & \cellcolor{blue!7} 2.7 [1.2, 6.5] & \cellcolor{blue!14} 5.5 [2.6, 11.6] & \cellcolor{blue!4} 1.6 [0.5, 5.7] & \cellcolor{blue!5} 2.0 [0.7, 6.6] & \cellcolor{blue!8} 2.9 [1.3, 7.5] \\
\bottomrule
\end{tabular}}
\caption{GSM: Unrobustness (U, \%) by model and modification}
\label{tab:gsm_unrob_ci}

\end{table*}

\begin{table*}[ht]
\centering
\resizebox{\linewidth}{!}{
\begin{tabular}{llrrrrrr}
\toprule
Category & Modification & \textbf{GPT-4o} & \textbf{Claude-3.5} & \textbf{Llama 3.1} & \textbf{GPT-5} & \textbf{DS R1} & \textbf{Avg} \\
\midrule
\textbf{Bias} & \textbf{Temporal} & \cellcolor{blue!29} 11.1 [6.3, 18.8] & \cellcolor{blue!8} 3.0 [1.0, 8.5] & \cellcolor{blue!29} 11.1 [6.3, 18.8] & \cellcolor{blue!13} 5.1 [2.2, 11.3] & \cellcolor{blue!31} 12.1 [7.1, 20.0] & \cellcolor{blue!22} 8.5 [4.6, 15.5] \\
  & \textbf{Geographical} & \cellcolor{blue!23} 9.0 [4.8, 16.2] & \cellcolor{blue!31} 12.0 [7.0, 19.8] & \cellcolor{blue!36} 14.0 [8.5, 22.1] & \cellcolor{blue!23} 9.0 [4.8, 16.2] & \cellcolor{blue!26} 10.0 [5.5, 17.4] & \cellcolor{blue!28} 10.8 [6.1, 18.4] \\
  & \textbf{Length} & \cellcolor{blue!28} 11.0 [6.3, 18.6] & \cellcolor{blue!26} 10.0 [5.5, 17.4] & \cellcolor{blue!44} 17.0 [10.9, 25.5] & \cellcolor{blue!15} 6.0 [2.8, 12.5] & \cellcolor{blue!41} 16.0 [10.1, 24.4] & \cellcolor{blue!31} 12.0 [7.1, 19.7] \\
\midrule
\textbf{Orthography} & \textbf{Capitalization} & \cellcolor{blue!21} 8.1 [4.2, 15.1] & \cellcolor{blue!13} 5.1 [2.2, 11.3] & \cellcolor{blue!26} 10.1 [5.6, 17.6] & \cellcolor{blue!23} 9.1 [4.9, 16.4] & \cellcolor{blue!36} 14.1 [8.6, 22.3] & \cellcolor{blue!24} 9.3 [5.1, 16.6] \\
  & \textbf{Punctuation} & \cellcolor{blue!18} 7.1 [3.5, 13.9] & \cellcolor{blue!8} 3.0 [1.0, 8.5] & \cellcolor{blue!34} 13.1 [7.8, 21.2] & \cellcolor{blue!21} 8.1 [4.2, 15.1] & \cellcolor{blue!31} 12.1 [7.1, 20.0] & \cellcolor{blue!22} 8.7 [4.7, 15.7] \\
  & \textbf{Spelling} & \cellcolor{blue!11} 4.1 [1.6, 10.1] & \cellcolor{blue!11} 4.1 [1.6, 10.1] & \cellcolor{blue!19} 7.2 [3.5, 14.2] & \cellcolor{blue!16} 6.2 [2.9, 12.8] & \cellcolor{blue!21} 8.2 [4.2, 15.4] & \cellcolor{blue!15} 6.0 [2.8, 12.5] \\
\midrule
\textbf{Syntax} & \textbf{Conjunction} & \cellcolor{blue!28} 11.0 [6.3, 18.6] & \cellcolor{blue!21} 8.0 [4.1, 15.0] & \cellcolor{blue!15} 6.0 [2.8, 12.5] & \cellcolor{blue!21} 8.0 [4.1, 15.0] & \cellcolor{blue!28} 11.0 [6.3, 18.6] & \cellcolor{blue!23} 8.8 [4.7, 15.9] \\
  & \textbf{Voice} & \cellcolor{blue!23} 9.0 [4.8, 16.2] & \cellcolor{blue!23} 9.0 [4.8, 16.2] & \cellcolor{blue!28} 11.0 [6.3, 18.6] & \cellcolor{blue!15} 6.0 [2.8, 12.5] & \cellcolor{blue!39} 15.0 [9.3, 23.3] & \cellcolor{blue!26} 10.0 [5.6, 17.4] \\
\midrule
\textbf{Semantics} & \textbf{Concept} & \cellcolor{blue!23} 9.1 [4.9, 16.4] & \cellcolor{blue!18} 7.1 [3.5, 13.9] & \cellcolor{blue!29} 11.1 [6.3, 18.8] & \cellcolor{blue!18} 7.1 [3.5, 13.9] & \cellcolor{blue!21} 8.1 [4.2, 15.1] & \cellcolor{blue!22} 8.5 [4.5, 15.6] \\
  & \textbf{Negation} & \cellcolor{blue!65} \textcolor{white}{25.0 [17.5, 34.3]} & \cellcolor{blue!57} \textcolor{white}{22.0 [15.0, 31.1]} & \cellcolor{blue!62} \textcolor{white}{24.0 [16.7, 33.2]} & \cellcolor{blue!59} \textcolor{white}{23.0 [15.8, 32.2]} & \cellcolor{blue!62} \textcolor{white}{24.0 [16.7, 33.2]} & \cellcolor{blue!61} \textcolor{white}{23.6 [16.4, 32.8]} \\
\midrule
\textbf{Discourse} & \textbf{Appraisal} & \cellcolor{blue!18} 7.1 [3.5, 14.0] & \cellcolor{blue!11} 4.1 [1.6, 10.0] & \cellcolor{blue!32} 12.2 [7.1, 20.2] & \cellcolor{blue!18} 7.1 [3.5, 14.0] & \cellcolor{blue!16} 6.1 [2.8, 12.7] & \cellcolor{blue!19} 7.3 [3.7, 14.2] \\
\midrule
\textbf{Varieties} & \textbf{Style} & \cellcolor{blue!46} \textcolor{white}{18.0 [11.7, 26.7]} & \cellcolor{blue!18} 7.0 [3.4, 13.7] & \cellcolor{blue!31} 12.0 [7.0, 19.8] & \cellcolor{blue!18} 7.0 [3.4, 13.7] & \cellcolor{blue!28} 11.0 [6.3, 18.6] & \cellcolor{blue!28} 11.0 [6.4, 18.5] \\
  & \textbf{Dialect} & \cellcolor{blue!31} 12.0 [7.0, 19.8] & \cellcolor{blue!34} 13.0 [7.8, 21.0] & \cellcolor{blue!31} 12.0 [7.0, 19.8] & \cellcolor{blue!28} 11.0 [6.3, 18.6] & \cellcolor{blue!39} 15.0 [9.3, 23.3] & \cellcolor{blue!33} 12.6 [7.5, 20.5] \\
\midrule
\textbf{Average} & &  \cellcolor{blue!28} 10.9 [6.3, 18.4] & \cellcolor{blue!21} 8.3 [4.5, 15.1] & \cellcolor{blue!32} 12.4 [7.4, 20.2] & \cellcolor{blue!22} 8.7 [4.7, 15.7] & \cellcolor{blue!32} 12.5 [7.5, 20.4] & \cellcolor{blue!27} 10.5 [6.1, 17.9] \\
\bottomrule
\end{tabular}}
\caption{IFEval: Unrobustness (U, \%) by model and modification}
\label{tab:ifeval_unrob_ci}
\end{table*}

\subsection{Scaling analysis (extended results)}

We present the scaling analysis results with confidence intereval in \Cref{tab:gsm_scaling_unrob_ci}.

\begin{table*}[ht]
\centering
\resizebox{\linewidth}{!}{
\begin{tabular}{llrrrr}
\toprule
Category & Modification & \textbf{Llama-8b} & \textbf{Llama-70b} & \textbf{Llama-405b} & \textbf{Avg} \\
\midrule
\textbf{Bias} & \textbf{Temporal} & \cellcolor{blue!21} 8.0 [4.1, 15.0] & \cellcolor{blue!13} 5.0 [2.2, 11.2] & \cellcolor{blue!10} 4.0 [1.6, 9.8] & \cellcolor{blue!15} 5.7 [2.6, 12.0] \\
  & \textbf{Geographical} & \cellcolor{blue!21} 8.0 [4.1, 15.0] & \cellcolor{blue!8} 3.0 [1.0, 8.5] & \cellcolor{blue!18} 7.0 [3.4, 13.7] & \cellcolor{blue!15} 6.0 [2.9, 12.4] \\
  & \textbf{Length} & \cellcolor{blue!21} 8.0 [4.1, 15.0] & \cellcolor{blue!5} 2.0 [0.6, 7.0] & \cellcolor{blue!5} 2.0 [0.6, 7.0] & \cellcolor{blue!10} 4.0 [1.7, 9.7] \\
\textbf{Orthographic} & \textbf{Spelling} & \cellcolor{blue!26} 10.0 [5.5, 17.4] & \cellcolor{blue!5} 2.0 [0.6, 7.0] & \cellcolor{blue!5} 2.0 [0.6, 7.0] & \cellcolor{blue!12} 4.7 [2.2, 10.5] \\
  & \textbf{Capitalization} & \cellcolor{blue!28} 11.0 [6.3, 18.6] & \cellcolor{blue!5} 2.0 [0.6, 7.0] & \cellcolor{blue!13} 5.0 [2.2, 11.2] & \cellcolor{blue!15} 6.0 [3.0, 12.3] \\
  & \textbf{Punctuation} & \cellcolor{blue!13} 5.0 [2.2, 11.2] & \cellcolor{blue!3} 1.0 [0.2, 5.4] & \cellcolor{blue!13} 5.0 [2.2, 11.2] & \cellcolor{blue!9} 3.7 [1.5, 9.3] \\
\textbf{Semantic} & \textbf{Concept} & \cellcolor{blue!31} 12.0 [7.0, 19.8] & \cellcolor{blue!3} 1.0 [0.2, 5.4] & \cellcolor{blue!13} 5.0 [2.2, 11.2] & \cellcolor{blue!15} 6.0 [3.1, 12.1] \\
  & \textbf{Negation} & \cellcolor{blue!57} \textcolor{white}{22.0 [15.0, 31.1]} & \cellcolor{blue!44} 17.0 [10.9, 25.5] & \cellcolor{blue!46} \textcolor{white}{18.0 [11.7, 26.7]} & \cellcolor{blue!49} \textcolor{white}{19.0 [12.5, 27.8]} \\
\textbf{Discourse} & \textbf{Appraisal} & \cellcolor{blue!18} 7.0 [3.4, 13.7] & \cellcolor{blue!13} 5.0 [2.2, 11.2] & \cellcolor{blue!3} 1.0 [0.2, 5.4] & \cellcolor{blue!11} 4.3 [1.9, 10.1] \\
\textbf{Varieties} & \textbf{Style} & \cellcolor{blue!41} 16.0 [10.1, 24.4] & \cellcolor{blue!15} 6.0 [2.8, 12.5] & \cellcolor{blue!15} 6.0 [2.8, 12.5] & \cellcolor{blue!24} 9.3 [5.2, 16.5] \\
  & \textbf{Dialect} & \cellcolor{blue!26} 10.0 [5.5, 17.4] & \cellcolor{blue!15} 6.0 [2.8, 12.5] & \cellcolor{blue!15} 6.0 [2.8, 12.5] & \cellcolor{blue!19} 7.3 [3.7, 14.1] \\
\textbf{Syntactic} & \textbf{Conjunction} & \cellcolor{blue!31} 12.0 [7.0, 19.8] & \cellcolor{blue!5} 2.0 [0.6, 7.0] & \cellcolor{blue!10} 4.0 [1.6, 9.8] & \cellcolor{blue!15} 6.0 [3.0, 12.2] \\
  & \textbf{Voice} & \cellcolor{blue!28} 11.0 [6.3, 18.6] & \cellcolor{blue!5} 2.0 [0.6, 7.0] & \cellcolor{blue!15} 6.0 [2.8, 12.5] & \cellcolor{blue!16} 6.3 [3.2, 12.7] \\
\midrule
\textbf{Average} &  & \cellcolor{blue!28} 10.8 [6.2, 18.2] & \cellcolor{blue!11} 4.2 [1.9, 9.8] & \cellcolor{blue!14} 5.5 [2.6, 11.6] & \cellcolor{blue!18} 6.8 [3.6, 13.2] \\
\bottomrule
\end{tabular}}
\caption{GSM Scaling: Unrobustness (U, \%) by Llama model size and modification}
\label{tab:gsm_scaling_unrob_ci}
\end{table*}

\subsection{Unrobustness results for each model averaged across tasks}

\Cref{tab:all_tasks_unrob} reports the Unrobustness scores for each model averaged across all tasks. Note that IFEval and GSM tasks only contain LLMs results. Overall, PLMs are more unrobust compared to LLMs. Among LLMs, the newer reasoning models are more robust compared to the base versions. Modification-wise, Geographical Bias, Negation, Style, and Dialect are the most prominent ones in creating instability.

\begin{table*}[ht]
\centering
\resizebox{\linewidth}{!}{
\begin{tabular}{llrrrrrrrrr}
\toprule
Category & Modification & \textbf{BERT} & \textbf{GPT-2} & \textbf{T5} & \textbf{GPT-4o} & \textbf{Claude-3.5} & \textbf{Llama 3.1} & \textbf{GPT-5} & \textbf{DS R1} & \textbf{Avg} \\
\midrule
\textbf{Bias} & \textbf{Temporal} & \cellcolor{blue!17} 6.4 [3.1, 12.3] & \cellcolor{blue!12} 4.5 [1.8, 10.0] & \cellcolor{blue!11} 4.3 [1.9, 9.3] & \cellcolor{blue!12} 4.7 [2.3, 9.5] & \cellcolor{blue!10} 3.9 [1.5, 9.2] & \cellcolor{blue!11} 4.3 [2.0, 9.5] & \cellcolor{blue!9} 3.6 [1.6, 8.0] & \cellcolor{blue!16} 6.1 [3.2, 11.9] & \cellcolor{blue!12} 4.7 [2.2, 10.0] \\ 
  & \textbf{Geographical} & \cellcolor{blue!29} 11.2 [7.2, 17.6] & \cellcolor{blue!35} 13.6 [9.1, 20.5] & \cellcolor{blue!34} 13.2 [9.1, 19.9] & \cellcolor{blue!29} 11.4 [7.1, 18.3] & \cellcolor{blue!30} 11.7 [7.3, 18.8] & \cellcolor{blue!31} 12.0 [7.6, 18.9] & \cellcolor{blue!26} 10.1 [6.3, 16.5] & \cellcolor{blue!30} 11.8 [7.7, 18.5] & \cellcolor{blue!31} 11.9 [7.7, 18.6] \\ 
  & \textbf{Length} & \cellcolor{blue!28} 11.0 [6.4, 17.9] & \cellcolor{blue!23} 8.8 [5.0, 15.2] & \cellcolor{blue!23} 8.8 [5.1, 15.1] & \cellcolor{blue!21} 8.3 [4.6, 14.8] & \cellcolor{blue!27} 10.5 [6.1, 17.6] & \cellcolor{blue!20} 7.6 [4.4, 13.5] & \cellcolor{blue!13} 5.1 [2.5, 10.1] & \cellcolor{blue!22} 8.6 [5.0, 15.0] & \cellcolor{blue!22} 8.6 [4.9, 14.9] \\ 
\midrule
\textbf{Orthographic} & \textbf{Spelling} & \cellcolor{blue!13} 5.0 [2.2, 11.2] & \cellcolor{blue!13} 5.0 [2.2, 11.2] & \cellcolor{blue!5} 2.0 [0.8, 4.9] & \cellcolor{blue!8} 3.0 [1.4, 6.8] & \cellcolor{blue!8} 3.0 [1.3, 7.0] & \cellcolor{blue!6} 2.3 [0.7, 7.5] & \cellcolor{blue!2} 0.7 [0.1, 3.6] & \cellcolor{blue!5} 2.0 [0.9, 4.2] & \cellcolor{blue!7} 2.9 [1.2, 7.0] \\ 
  & \textbf{Capitalization} & \cellcolor{blue!0} 0.0 [0.0, 0.0] & \cellcolor{blue!12} 4.6 [1.9, 10.8] & \cellcolor{blue!12} 4.6 [1.9, 10.8] & \cellcolor{blue!6} 2.4 [0.8, 7.6] & \cellcolor{blue!9} 3.4 [1.4, 8.9] & \cellcolor{blue!9} 3.4 [1.3, 9.0] & \cellcolor{blue!4} 1.7 [0.6, 5.2] & \cellcolor{blue!4} 1.7 [0.5, 6.5] & \cellcolor{blue!7} 2.7 [1.1, 7.3] \\ 
  & \textbf{Punctuation} & \cellcolor{blue!6} 2.5 [0.9, 7.6] & \cellcolor{blue!5} 2.0 [0.6, 7.0] & \cellcolor{blue!10} 4.0 [2.0, 7.5] & \cellcolor{blue!3} 1.3 [0.4, 4.6] & \cellcolor{blue!5} 2.0 [0.7, 5.7] & \cellcolor{blue!8} 3.0 [1.1, 8.4] & \cellcolor{blue!6} 2.3 [0.9, 7.3] & \cellcolor{blue!3} 1.3 [0.3, 5.9] & \cellcolor{blue!6} 2.3 [0.9, 6.8] \\ 
\midrule
\textbf{Semantic} & \textbf{Concept} & \cellcolor{blue!17} 6.5 [3.1, 13.1] & \cellcolor{blue!14} 5.5 [2.5, 11.8] & \cellcolor{blue!12} 4.5 [1.9, 10.5] & \cellcolor{blue!11} 4.3 [2.0, 10.1] & \cellcolor{blue!9} 3.7 [1.7, 7.8] & \cellcolor{blue!8} 3.0 [1.1, 8.4] & \cellcolor{blue!11} 4.3 [2.1, 9.9] & \cellcolor{blue!10} 4.0 [1.7, 9.7] & \cellcolor{blue!12} 4.5 [2.0, 10.2] \\ 
  & \textbf{Negation} & \cellcolor{blue!67} \textcolor{white}{25.9 [18.3, 35.4]} & \cellcolor{blue!77} \textcolor{white}{29.9 [21.9, 39.4]} & \cellcolor{blue!72} \textcolor{white}{28.0 [20.1, 37.5]} & \cellcolor{blue!54} \textcolor{white}{20.9 [14.2, 29.8]} & \cellcolor{blue!54} \textcolor{white}{20.9 [14.2, 29.8]} & \cellcolor{blue!55} \textcolor{white}{21.2 [14.4, 30.2]} & \cellcolor{blue!42} 16.2 [10.5, 24.5] & \cellcolor{blue!44} 17.2 [11.2, 25.7] & \cellcolor{blue!58} \textcolor{white}{22.5 [15.6, 31.5]} \\ 
\midrule
\textbf{Discourse} & \textbf{Appraisal} & \cellcolor{blue!14} 5.2 [2.5, 9.9] & \cellcolor{blue!14} 5.3 [2.4, 10.0] & \cellcolor{blue!11} 4.4 [2.0, 8.8] & \cellcolor{blue!12} 4.6 [2.2, 8.6] & \cellcolor{blue!9} 3.5 [1.5, 7.4] & \cellcolor{blue!16} 6.3 [3.2, 11.8] & \cellcolor{blue!13} 5.0 [2.5, 10.2] & \cellcolor{blue!18} 7.0 [3.9, 12.8] & \cellcolor{blue!13} 5.2 [2.6, 9.9] \\ 
\midrule
\textbf{Varieties} & \textbf{Style} & \cellcolor{blue!36} 13.9 [8.3, 21.0] & \cellcolor{blue!40} 15.6 [9.9, 22.8] & \cellcolor{blue!40} 15.6 [10.1, 22.5] & \cellcolor{blue!29} 11.2 [6.9, 17.5] & \cellcolor{blue!19} 7.2 [3.8, 13.1] & \cellcolor{blue!25} 9.6 [5.3, 16.2] & \cellcolor{blue!21} 8.1 [4.4, 14.2] & \cellcolor{blue!24} 9.2 [5.3, 15.6] & \cellcolor{blue!29} 11.3 [6.8, 17.9] \\ 
  & \textbf{Dialect} & \cellcolor{blue!21} 8.3 [4.4, 15.1] & \cellcolor{blue!69} \textcolor{white}{26.9 [19.4, 35.9]} & \cellcolor{blue!57} \textcolor{white}{22.2 [15.4, 30.9]} & \cellcolor{blue!33} 12.8 [8.3, 20.2] & \cellcolor{blue!25} 9.6 [5.4, 16.7] & \cellcolor{blue!30} 11.6 [6.8, 19.1] & \cellcolor{blue!13} 5.0 [2.5, 10.8] & \cellcolor{blue!26} 9.9 [5.8, 17.0] & \cellcolor{blue!34} 13.3 [8.5, 20.7] \\ 
\midrule
\textbf{Syntactic} & \textbf{Conjunction} & \cellcolor{blue!10} 4.0 [1.6, 9.8] & \cellcolor{blue!13} 5.0 [2.2, 11.2] & \cellcolor{blue!10} 4.0 [1.8, 9.6] & \cellcolor{blue!6} 2.3 [1.0, 6.0] & \cellcolor{blue!4} 1.7 [0.5, 6.4] & \cellcolor{blue!8} 3.0 [1.1, 8.4] & \cellcolor{blue!4} 1.7 [0.6, 5.1] & \cellcolor{blue!4} 1.7 [0.6, 5.1] & \cellcolor{blue!8} 2.9 [1.2, 7.7] \\ 
  & \textbf{Voice} & \cellcolor{blue!23} 9.0 [4.8, 16.2] & \cellcolor{blue!27} 10.5 [5.9, 18.0] & \cellcolor{blue!21} 8.0 [4.1, 14.9] & \cellcolor{blue!11} 4.3 [1.9, 10.2] & \cellcolor{blue!12} 4.7 [2.2, 10.5] & \cellcolor{blue!10} 4.0 [1.6, 9.8] & \cellcolor{blue!13} 5.0 [2.4, 10.9] & \cellcolor{blue!9} 3.7 [1.5, 9.3] & \cellcolor{blue!16} 6.1 [3.0, 12.5] \\ 
\midrule
\textbf{Average} &  \cellcolor{blue!22} 8.4 [4.8, 14.4] & \cellcolor{blue!27} 10.6 [6.5, 17.2] & \cellcolor{blue!25} 9.5 [5.9, 15.5] & \cellcolor{blue!18} 7.0 [4.1, 12.6] & \cellcolor{blue!17} 6.6 [3.7, 12.2] & \cellcolor{blue!18} 7.0 [3.9, 13.1] & \cellcolor{blue!14} 5.3 [2.9, 10.5] & \cellcolor{blue!17} 6.5 [3.7, 12.1] & \cellcolor{blue!20} 7.6 [4.4, 13.5] \\ 
\bottomrule
\end{tabular}}
\caption{All Tasks: Unrobustness (U, \%) by model and modification (averaged across tasks)}
\label{tab:all_tasks_unrob}
\end{table*}

\subsection{AI usage disclosure}

Claude-4-Sonnet (through Claude Code) was used to help debug issues in the code and help generating the tables in LaTex.

\end{document}